\documentclass{article}

\usepackage{microtype}
\usepackage{graphicx}
\usepackage{booktabs} 

\usepackage{hyperref}

\usepackage[accepted]{icml2024}

\usepackage{amsmath}
\usepackage{amssymb}
\usepackage{mathtools}
\usepackage{amsthm}
\usepackage{subfig}
\usepackage{adjustbox}

\usepackage[capitalize,noabbrev]{cleveref}

\theoremstyle{plain}

\theoremstyle{definition}

\theoremstyle{remark}

\usepackage[textsize=tiny]{todonotes}

\icmltitlerunning{PIPER}

\begin{document}

\twocolumn[
\icmltitle{PIPER: Primitive-Informed Preference-based \\ Hierarchical Reinforcement Learning via Hindsight Relabeling}




\begin{icmlauthorlist}
\icmlauthor{Utsav Singh}{aaa}
\icmlauthor{Wesley A. Suttle}{bbb}
\icmlauthor{Brian M. Sadler}{ut}
\icmlauthor{Vinay P. Namboodiri}{ddd}
\icmlauthor{Amrit Singh Bedi}{ccc}
\end{icmlauthorlist}

\icmlaffiliation{aaa}{CSE dept., IIT Kanpur, Kanpur, India.}
\icmlaffiliation{bbb}{U.S. Army Research Laboratory, Adelphi, MD, USA.}
\icmlaffiliation{ut}{University of Texas, Austin, Texas, USA.}
\icmlaffiliation{ccc}{CS dept., University of Central Florida, Orlando, Florida, USA.}
\icmlaffiliation{ddd}{Department of Computer Science, University of Bath, Bath, UK.}
%

\icmlcorrespondingauthor{Utsav Singh}{utsavz@cse.iitk.ac.in}

\icmlkeywords{Machine Learning, ICML}

\vskip 0.3in
]



\printAffiliationsAndNotice{} 

\begin{abstract}
\label{sec:abstract}

In this work, we introduce PIPER: Primitive-Informed Preference-based Hierarchical reinforcement learning via Hindsight Relabeling, a novel approach that leverages preference-based learning to learn a reward model, and subsequently uses this reward model to relabel higher-level replay buffers. Since this reward is unaffected by lower primitive behavior, our relabeling-based approach is able to mitigate non-stationarity, which is common in existing hierarchical approaches, and demonstrates impressive performance across a range of challenging sparse-reward tasks. Since obtaining human feedback is typically impractical, we propose to replace the human-in-the-loop approach with our primitive-in-the-loop approach, which generates feedback using sparse rewards provided by the environment. Moreover, in order to prevent infeasible subgoal prediction and avoid degenerate solutions, we propose primitive-informed regularization that conditions higher-level policies to generate feasible subgoals for lower-level policies. We perform extensive experiments to show that PIPER mitigates non-stationarity in hierarchical reinforcement learning and achieves greater than 50$\%$ success rates in challenging, sparse-reward robotic environments, where most other baselines fail to achieve any significant progress. 
\end{abstract}
\section{Introduction}
\label{sec:introduction}

\begin{figure*}[t]
\centering
\captionsetup{font=small,labelfont=small,textfont=small}
\includegraphics[scale=0.38]{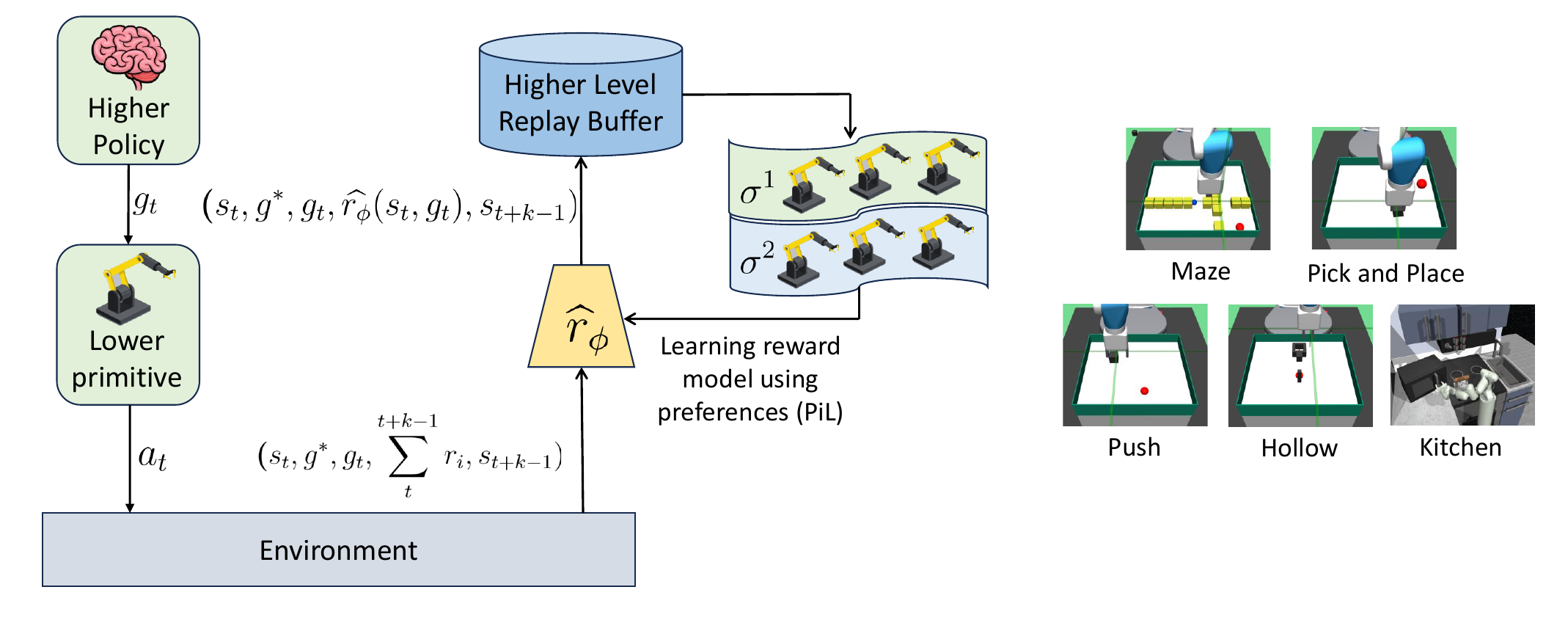}
\caption{\textbf{PIPER Overview} This figure shows the overview of PIPER (left). The higher level policy predicts subgoals $g_t$ for the lower primitive, which executes actions $a_t$ on the environment. We propose to learn a preference-based reward model $\widehat{r}_{\phi}$ using our PiL feedback on higher level trajectories sampled from higher level replay buffer, and subsequently use $\widehat{r}_{\phi}$ to relabel the replay buffer transitions, thereby mitigating non-stationarity in HRL. On the right, we depict the training environments: $(i)$ maze navigation environment, $(ii)$ pick and place environment, $(iii)$ push environment, $(iv)$ hollow environment, and $(v)$ franka kitchen environment.}
\label{fig:explain method}
\end{figure*}

Deep reinforcement learning (RL) has propelled significant advances in sequential decision-making tasks, where an agent learns complex behaviors through trial and error. Examples of such tasks include playing Atari games~\citep{DBLP:journals/corr/MnihKSGAWR13}, mastering the game of Go~\citep{silver2016mastering}, and performing complicated robotic manipulation tasks~\citep{DBLP:journals/corr/abs-1709-10087, DBLP:journals/corr/abs-1806-10293, DBLP:journals/corr/GuHLL16, DBLP:journals/corr/LevineFDA15}. However, the success of RL-based approaches is impeded by issues like ineffective exploration and long-term credit assignment, especially in sparse-reward scenarios~\citep{nair2018overcoming}. 

Off-policy hierarchical reinforcement learning (HRL)~\citep{sutton1999between,dayan1992feudal,vezhnevets2017feudal,klissarov2017learnings,harb2018waiting} approaches hold the promise of improved sample efficiency due to temporal abstraction and improved performance due to more effective exploration~\citep{nachum2019does}. 
In goal-conditioned feudal architectures~\citep{dayan1992feudal, vezhnevets2017feudal}, a widely used subset of HRL methods, higher-level policies predict subgoals for the lower-level policies, while the lower-level policies try to achieve those subgoals by executing primitive actions directly in the environment.
Despite their advantages, serious challenges remain for such approaches. In this paper, we focus on two important issues: the destabilizing effect of lower-level reward non-stationarity on off-policy HRL \citep{levy2018learning,nachum2018data}, and performance degradation due to infeasible subgoal generation by higher-level policies \cite{chane2021goal}. \footnote{The implementation code and data is provided \href{https://github.com/Utsavz/piper.git}{here}.}
%
%
%


Recent approaches in preference-based learning~\cite{christiano2017deep,DBLP:journals/corr/abs-1811-06521,DBLP:journals/corr/abs-2106-05091} use human feedback to learn a reward model and subsequently apply RL to solve the task using the learned reward model. Impressively, such methods have demonstrated performance comparable to RL policies trained with access to ground truth rewards. This raises the following question: \textit{can advances in preference-based learning be used to simultaneously address the twin issues of reward non-stationarity and infeasible subgoal generation in HRL?}

In this work, we provide an affirmative answer to the above question by proposing PIPER: \textbf{P}rimitive \textbf{I}nformed \textbf{P}r\textbf{E}ference-based hierarchical reinforcement learning via hindsight \textbf{R}elabeling. The key realization underlying PIPER is that a suitably designed preference-based RL approach can be used to learn a high-level reward model that is simultaneously decoupled from non-stationary, lower-level rewards and carefully tailored to generate feasible subgoals. To achieve this, several innovations are necessary. First, to overcome the problem of acquiring the human preference feedback needed to learn the higher-level reward model, we propose a goal-conditioned, sparse reward-based approach to replace human trajectory preferences. Second, to address the sample-inefficiency arising from the use of sparse rewards, we apply hindsight experience relabeling \citep{andrychowicz2017hindsight} to learn a denser, more informative higher-level reward model. Finally, to encourage our higher-level policies to predict subgoals achievable by lower-level policies, we propose a novel value-function regularization scheme that calibrates subgoal selection to the current lower-level policy's abilities. Taken together, these techniques form the core of PIPER (see Figure \ref{fig:explain method}). We perform extensive experiments in complex, sparse-reward environments that empirically show that PIPER demonstrates impressive performance and consistently outperforms the baselines, as well as ablation studies illustrating the importance of each of PIPER's component techniques.

We summarize the main contributions of PIPER:
\begin{itemize}
    \item We demonstrate that PIPER is able to mitigate non-stationarity in off-policy HRL, by employing preference-based feedback.
    \item Since collecting human preference feedback is impractical, we propose an alternative primitive-in-the-loop (PiL) scheme to determine preferences between trajectories without human feedback.
    \item Using primitive informed regularization, PIPER generates efficient subgoals, thus improving performance.
    \item We employ hindsight relabeling to improve sample efficiency in preference-based learning and deal with sparsity in sparse-reward scenarios.
    \item PIPER uses soft target updates to mitigate instability due to non-stationary reward model learned using preference-based feedback.
    \item PIPER achieves greater than 50$\%$ success rates in challenging, sparse-reward robotic environments, where most other hierarchical and non-hierarchical baselines fail to achieve any significant progress.
\end{itemize}

\section{Related Works}
\label{sec:related_work}


\textbf{Hierarchical Reinforcement Learning.} HRL approaches~\citep{sutton1999between,Barto03recentadvances, NIPS1997_5ca3e9b1, DBLP:journals/corr/cs-LG-9905014} promise the intuitive benefits of improved exploration and temporal abstraction~\citep{nachum2019does}. In goal-conditioned feudal hierarchical learning-based approaches~\citep{dayan1992feudal,vezhnevets2017feudal}, the higher-level policy predicts subgoals for the lower-level primitive, which in turn tries to achieve them by executing primitive actions on the environment. However, such off-policy HRL approaches face non-stationarity due to dynamically changing lower-level primitive behavior. Prior approaches~\citep{nachum2018data, levy2018learning} partially address this non-stationarity by relabeling the goals in the replay buffer. In contrast, we propose a novel approach that first uses preference-based learning~\citep{christiano2017deep,DBLP:journals/corr/abs-2106-05091} to learn a reward model, and subsequently uses the reward model to relabel the non-stationary rewards in the replay buffer. 

The option learning framework~\citep{sutton1999between,klissarov2017learnings} provides the benefits of temporal abstraction by learning extended macro actions. Unfortunately, such approaches can lead to degenerate solutions and thus require additional regularization approaches to ensure feasible subgoal prediction. Notably, such degenerate solutions result in unbalanced task-split which nullifies the advantages of hierarchical learning. In our approach, we perform primitive-informed regularization to regularize the higher-level policy to predict achievable subgoals for lower primitive. Another line of work uses previously hand-designed action primitives~\cite{DBLP:journals/corr/abs-2110-03655,dalal2021accelerating} to accelerate hierarchical learning. Such approaches, however, depend on the quality of hand-designed primitives, which is difficult to design in complex environments.

\textbf{Preference-based Learning.} Several prior works have proposed approaches that perform reinforcement learning on human rankings~\citep{knox2009interactively,pilarski2011online,wilson2012bayesian,daniel2015active}. ~\citep{warnell2018deep} extend the TAMER framework by replacing the reward function by feedback signal. ~\citep{christiano2017deep} used deep RL advancements to learn a reward model using neural networks, based on human preference feedback. Although some works employed on-policy RL~\citep{christiano2017deep} for solving tasks, a more sample efficient approach~\citep{DBLP:journals/corr/abs-2106-05091} used off-policy RL~\citep{DBLP:journals/corr/abs-1801-01290} to learn a policy using preference feedback, thereby improving sample efficiency. In this work, we propose an off-policy hierarchical RL method that uses preference-based feedback to mitigate non-stationarity in HRL, and uses primitive informed regularization to generate achievable subgoals for the lower primitive.

\section{Problem Formulation}


We consider an MDP $(\mathcal{S}, \mathcal{A}, p, r, \gamma)$, where $\mathcal{S}$ is the state space, $\mathcal{A}$ is the action space, $p : \mathcal{S} \times \mathcal{A} \rightarrow \Delta(\mathcal{S})$ is the transition probability function mapping state-action pairs to probability distributions over the state space, $r:\mathcal{S} \times \mathcal{A} \rightarrow \mathbb{R}$ is the reward function, and $\gamma \in (0, 1)$ is the discount factor. At a given timestep $t$, the agent is in state $s_t$, takes action $a_t \sim \pi(\cdot | s_t)$ according to some policy $\pi : \mathcal{S} \rightarrow \Delta(\mathcal{A})$ which maps states to probability distributions over the action space, receives reward $r_t = r(s_t,a_t)$, and the system transitions to a new state $s_{t+1} \sim p(\cdot | s_t, a_t)$. The standard reinforcement learning  (RL) objective is given by
\begin{align}\label{policy_optim}
    \pi^*:=\arg\max_{\pi} J(\pi) = \mathbb{E}_{\pi} \left[ \sum_{t=0}^{\infty} \gamma^t r_t \right],
\end{align}
which is also called policy optimization in literature. The state action value function $Q(s,g,a_i)$ computes the expected cumulative reward when the start state is $s$, $g$ is the final goal, and the next primitive action is $a_i$. In hierarchical RL (HRL) problem,  the higher-level policy aims to achieve an end goal by issuing subgoals to the lower-level policy, while the lower-level policy chooses primitive actions oriented towards achieving the specified subgoals.
%
In {HRL}, the higher-level policy $\pi^H : \mathcal{S} \rightarrow \Delta(\mathcal{G})$ specifies a subgoal $g_t \in \mathcal{G}$, where $\mathcal{G} \subset \mathcal{S}$ is the set of possible goals. During execution at each time step $t$, the subgoal $g_t \sim \pi^H(\cdot | s_t)$ after every $k$ timesteps and $g_t = g_{k \cdot \lceil t / k \rceil }$, otherwise. The effect of this is that the higher-level policy issues new subgoals every $k$ timesteps and keeps subgoals fixed in between. 

Furthermore, at each $t$, the lower-level policy $\pi^L : \mathcal{S} \times \mathcal{G} \rightarrow \Delta(\mathcal{A})$ selects primitive actions $a_t \sim \pi^L(\cdot | s_t, g_t)$ according to the current state and subgoal specified by $\pi^H$, and the state transitions to $s_{t+1} \sim p(\cdot | s_t, a_t)$. Finally, at each timestep $t$, the higher level of the hierarchy provides the lower level with reward $r^L_t = r^L(s_t, g_t, a_t) = -\mathbf{1}_{\{ \| s_t - g_t \|_2 > \varepsilon \}}$, where $\mathbf{1}_{B}$ is the indicator function on a given set $B$, while the higher level receives reward $r^H_t = r^H(s_t, g^*, g_t)$, where $g^* \in \mathcal{G}$ is the end goal and $r^H : \mathcal{S} \times \mathcal{G} \times \mathcal{G} \rightarrow \mathbb{R}$ is a high-level reward function that we have yet to specify. The lower level populates its replay buffer with samples $(s_t, g_t, a_t, r^L_t, s_{t+1})$, while, at each $t$ such that after every $k$ timesteps, the higher level populates its buffer with samples of the form $(s_t, g^*, g_t, \sum_{i=t}^{t+k-1} r^H_i, s_{t+k})$. Next, we highlight key limitations of existing HRL methods.

\subsection{Limitations of Existing Approaches to HRL} \label{subsec:HRL_limitations}

While HRL promises significant advantages over non-hierarchical RL, including improved sample efficiency due to temporal abstraction and improved exploration \cite{nachum2018data, nachum2019does}, serious challenges remain. In this work, we focus on two outstanding issues: 

C1: the destabilizing effect of lower-level reward non-stationarity on off-policy HRL, 

C2: and performance degradation due to infeasible subgoal generation by higher-level policies. 

As discussed in \citet{nachum2018data} and \citet{levy2018learning}, off-policy HRL suffers from non-stationarity due to changing lower primitive behavior, since, for a given transition in the higher-level replay buffer $(s_t, g^*,g_t, \sum_{i=t}^{t+k-1} r^H_i, s_{t+k})$, the rewards and transition may have been generated by an outdated lower-level policy. In addition, it was observed in \citet{chane2021goal} that, without taking care to ensure valid subgoal selection at the higher level, HRL methods may issue unachievable subgoals to the lower-level policy, stalling learning at the lower level and significantly impacting performance. The primary motivation of this work is the development of a novel, preference-based learning-inspired technique for learning the high-level reward function, $r^H$, that addresses these key limitations of existing HRL methods.

\section{Proposed Approach}


In this section, we introduce our approach, PIPER: \textbf{P}rimitive \textbf{I}nformed \textbf{P}r\textbf{E}ference-based hierarchical reinforcement learning via hindsight \textbf{R}elabeling for solving complex sparse-reward tasks. The motivating idea behind our approach is that preference-based RL, where human preferences are used to learn a dense, informative reward function, can be used to learn a high-level reward function that simultaneously mitigates the issues of reward non-stationarity and infeasible subgoal generation described in Section \ref{subsec:HRL_limitations}. Before presenting the main idea, let us briefly discuss preference-based learning. 
\subsection{Preference-based Learning (PBL)}  \label{subsec:pbl}
In traditional RL, agents learn to maximize the accumulated rewards $r(s_t,a_t)$ obtained from the environment, where the reward $r : \mathcal{S} \times \mathcal{A} \rightarrow \mathbb{R}$ is assumed to be known. In many real-world scenarios, however, suitable environment rewards are notoriously difficult to construct and typically require domain-specific knowledge \cite{ng1999policy}. To learn the high-level reward necessary for the HRL setting considered in this paper, we use a preference-based RL setup, where the agent learns to perform the high-level task using preferences over agent behaviors~\citep{NIPS2012_16c222aa, christiano2017deep, DBLP:journals/corr/abs-2106-05091, DBLP:journals/corr/abs-1811-06521}.

In the preference-based setting, agent behavior over a $k$-length trajectory can be represented by a sequence, $\delta$, of state observations and actions: $\delta=((s_t, a_t),(s_{t+1}, a_{t+1})...(s_{t+k-1}, a_{t+k-1}))$. The goal is to learn a reward function, $\widehat{r}_{\phi} : \mathcal{S} \times \mathcal{A} \rightarrow \mathbb{R}$, with neural network parameters $\phi$, such that preferences between any two trajectories $\delta^1, \delta^2$ can be modeled using the Bradley-Terry model
%
%
\citep{bradley_terry}:
\begin{align}
\label{eqn:bradley_terry}
    P_\phi\left[\delta^1 \succ \delta^2\right]=\frac{\exp \sum_t \widehat{r}_\phi\left(s_t^1, a_t^1\right)}{\sum_{i \in\{1,2\}} \exp \sum_t \widehat{r}_\phi\left(s_t^i, a_t^i\right)},
\end{align}
where $\delta^1 \succ \delta^2$ denotes the event that $\delta^1$ is preferred over $\delta^2$. To learn parameters $\phi$ such that \eqref{eqn:bradley_terry} matches the true preferences, behavior and preference data are recorded in a dataset $\mathcal{D}$ with entries of the form $(\delta^1, \delta^2, y)$, where $y=(1,0)$ when $\delta^1$ is preferred over $\delta^2$, $y=(0,1)$ when $\delta^2$ is preferred over $\delta^1$, and $y=(0.5,0.5)$ when there is no preference. The standard  approach in the preference-based literature (see \cite{christiano2017deep,DBLP:journals/corr/abs-2106-05091}), which we adopt in PIPER, is to learn the reward function $\widehat{r}_{\phi}$ using a cross-entropy loss:
%
%
%
%
\begin{equation} \label{eqn:reward_cross_entropy}
    \operatorname{L}(\phi) = -\!\sum_{\mathcal{D}} \left( y_1 \!\log P_\phi\left[\delta^1 \!\succ\! \delta^2 \right] +
    y_2 \log P_\phi\left[ \delta^2 \!\succ\! \delta^1 \right] \right),
\end{equation}
where $(\delta^1, \delta^2, y) \in \mathcal{D}$ and $y_1$ and $y_2$ denote the first and second entries of $y$. Given the preference model and learning objective in \eqref{eqn:bradley_terry} and \eqref{eqn:reward_cross_entropy}, we now turn to the problem of acquiring preference data in our hierarchical setting.

\textbf{Challenges of directly applying PBL in HRL:}

\begin{itemize}
    \item First, collecting the quantity of human feedback needed to enable preference-based learning of the high-level reward function is impractical. To address this, we propose a replacement PiL scheme to determine preferences between trajectories without human feedback. 

    \item Second, standard goal-conditioned rewards are too sparse for sample-efficient learning. We therefore incorporate hindsight relabeling to reduce sparsity and increase the informativeness of observed trajectories. 

    \item Third, even with hindsight relabeling the higher-level policy may choose subgoals that are infeasible for the lower-level policy, so we regularize the PiL reward with the value function of the lower-level policy to encourage feasible subgoal selection. 

    \item Finally, we incorporate soft target updates \cite{lillicrap2015continuous} to mitigate the training instability while learning the high-level reward function. Taken together, this combination of techniques constitutes PIPER, the primary contribution of this paper. The rest of this section provides details on each aspect of our approach. Pseudo-code for PIPER is provided in Algorithm \ref{alg:algo_piper}.
\end{itemize}

\subsection{PiL: Primitive-in-the-Loop Feedback for HRL} \label{section:pil}
We next introduce our Primitive-in-the-Loop (PiL) approach for generating higher-level preference feedback. This technique provides an effective, easily computable replacement for human feedback, avoiding the impracticality of obtaining human preference data for trajectories observed during training in the HRL setting. The key idea is to replace human preferences over higher-level trajectories with preferences generated by using a simple, $g^*$-conditioned sparse reward to perform pairwise comparisons. We note that the sparsity of the proposed reward presents additional issues, which we address in Section \ref{sec:hr} below.

\textbf{Implicit reward functions:} Let $s$ be the current state, $g$ be the subgoal predicted by higher level, and $g^*$ be the final goal.
%
%
In our goal-conditioned HRL setting, we represent $k$-length trajectories of lower-level primitive behavior as sequences $\Delta$ given by $\Delta=((s_t, a_t),(s_{t+1}, a_{t+1}), \ldots, (s_{t+k-1}, a_{t+k-1}))$, and $n$-length trajectories of higher-level behavior as sequences $\sigma$ of states and subgoal predictions, i.e., $\sigma = ((s_t, g_t), (s_{t+k}, g_{t+k}), \ldots, (s_{t+(n-1)k}, g_{t+(n-1)k}))$, where the higher-level policy is executed for $n$ timesteps and the lower-level policy is executed for $k$ timesteps between each pair of consecutive subgoal predictions.
In preference-based learning, the preferences are assumed to correspond to an implicit reward function $r:\mathcal{S}\times\mathcal{G}\times\mathcal{A}\rightarrow\mathbb{R}$ implied by the true preferences. Let $g^*$ be the final goal and $\sigma^{1}$ and $\sigma^{2}$ be two higher-level trajectories. Whenever $\sigma^1$ is preferred to $\sigma^2$, denoted by $\sigma^1 \succ \sigma^2$, then the assumption that $r$ encodes the true preferences implies:

\begin{algorithm}[ht]
\caption{PIPER}
\label{alg:algo_piper}
\begin{algorithmic}[1]
    \STATE Initialize preference dataset $D = \{\}$
    \STATE Initialize higher level replay buffer $\mathcal{R}^{H} = \{\}$ and lower level replay buffer $\mathcal{R}^{L} = \{\}$
    \STATE Initialize reward model parameters $\phi$
    \STATE Initialize target reward model parameters of $\phi' \leftarrow \phi$
    \FOR{$i = 1 \ldots N $}
        \STATE // Collect experience using $\pi^{H}$ and $\pi^{L}$ and store transitions in $\mathcal{R}^{H}$ and $\mathcal{R}^{L}$
        \FOR{each timestep $t$}
            \STATE $d^{H} \leftarrow d^{H} \cup \{(s_t, g^*, g_t, \sum_{i=t}^{t+k-1}r_i, s_{t+k-1})\}$
            \STATE $d^{L} \leftarrow d^{L} \cup \{(s_t, g_t, a_t, r_t, s_{t+1})\}$
        \ENDFOR
        \STATE $\mathcal{R}^{H} \leftarrow \mathcal{R}^{H} \cup d^{H}$
        \STATE $\mathcal{R}^{L} \leftarrow \mathcal{R}^{L} \cup d^{L}$
        \STATE // Sample higher level behavior trajectories and goal
        \STATE $(\sigma^{1}, \sigma^{2}) \sim \mathcal{R^{H}}$ and $g \sim \mathcal{G}$
        \STATE Sample a set of additional goals for relabeling $\mathcal{G^{'}}$
        \FOR{$g^{'} \in \mathcal{G^{'}}$}
            \STATE Relabel $g$ by $g^{'}$ and generate label $y$ using Equation~\eqref{eqn:hr_with_reg}
            \STATE Store preference $\mathcal{D} \leftarrow \mathcal{D} \cup \{(\sigma^{1}, \sigma^{2}, y)\}$
        \ENDFOR
        \STATE // Reward Model Learning
        \FOR{each gradient step}
            \STATE Optimize model reward $\widehat{r}_{\phi}$ using Equation~\eqref{eqn:reward_cross_entropy}
            \STATE // Soft update target reward model parameters
            \STATE $\phi' \gets \tau \phi + (1 - \tau) \phi'$
        \ENDFOR
        \STATE // Policy Learning
        \FOR{each gradient step}
            \STATE Sample $\{(\sigma_j)\}_{j=1}^{m}$ from $\mathcal{R}^{H}$
            \STATE Sample $\{(\delta_j)\}_{j=1}^{m}$ from $\mathcal{R}^{L}$
            \STATE Relabel rewards in $\{(\sigma_j)\}_{j=1}^{m}$ using target reward model $\widehat{r}_{\phi^{'}}$
            \STATE Optimize higher policy $\pi^{H}$ using SAC
            \STATE Optimize lower policy $\pi^{L}$ using SAC
        \ENDFOR
    \ENDFOR
\end{algorithmic}
\end{algorithm}

\begin{equation} \label{eqn:PiL}
    \begin{aligned}
        \small
        \sum_{i=0}^{n-1} & r(s^1_{t + ik}, g^*, g^1_{t + ik}) > \sum_{i=0}^{n-1} r(s^2_{t + ik}, g^*, g^2_{t + ik}).
    \end{aligned}
\end{equation}

\textbf{Replacing human feedback:} In the standard preference-learning framework, preferences are elicited from human feedback~\cite{christiano2017deep} and are subsequently used to learn the reward model $\widehat{r}_{\phi}$.
In this work, we replace this Human-in-the-Loop (HiL) feedback with Primitive-in-the-Loop (PiL) feedback by using implicit sparse rewards, $r^s (s_t, g^*, g_{t})$, defined presently, to determine preferences $y$ between behavior sequences $\sigma^{1}$ and $\sigma^{2}$. We call this feedback \textit{Primitive-in-the-Loop} since we generate this feedback using primitive sparse rewards.
%
%
We obtain these primitive rewards as follows. Suppose the higher level policy $\pi^{H}$ predicts subgoal $g_t \sim \pi^{H}( \cdot | s_t, g^*)$ for state $s_t$ and goal $g^*$. The lower-level primitive executes primitive actions according to its policy $\pi^{L}$ for $k$ timesteps and ends up in state $s_{t+k-1}$. We use the sparse reward  provided by the environment at state $s_{t+k-1}$ as the implicit sparse reward, i.e., $r^s(s_t, g^*, g_{t}) = -\mathbf{1}_{\{ \| s_{t+k-1} - g^* \|_2 > \epsilon \}}$.
%
%
Note that this reward is directly available from the environment. We replace the implicit reward $r$ in Equation \eqref{eqn:PiL} with $r^{s}$, and thus use $r^{s}$ to obtain preferences between higher level behavior sequences. Concretely, for the goal $g^*$, $\sigma^1 \succ \sigma^2$ implies
%
%
\begin{equation} \label{eqn:reward_s}
    \begin{aligned}
        \small
        \sum_{i=0}^{n-1} & r^s(s^1_{t + ik}, g^*, g^1_{t + ik}) > \sum_{i=0}^{n-1} r^s(s^2_{t + ik}, g^*, g^2_{t + ik}).
    \end{aligned}
\end{equation}
The preferences elicited using $r^{s}$ are subsequently used to learn the preference reward function $\widehat{r}_{\phi}$.
%

\textbf{Effect on non-stationarity:} Off-policy HRL approaches suffer from non-stationarity due to outdated transitions in the higher-level replay buffer caused by the changing lower-level policy. In our approach, the rewards in the higher-level replay buffer are relabeled using $\widehat{r}_{\phi}$. Hence, the transitions are updated from $(s_t, g^*, g_t, \sum_{i=t}^{t+k-1} r_i, s_{t+k-1})$ to $(s_t, g^*, g_t, \widehat{r}_{\phi}(s_t,g_t), s_{t+k-1})$. Since $\widehat{r}_{\phi}$ does not depend on changing lower primitive behavior, this reward relabeling eliminates the non-stationarity typically encountered in off-policy HRL. 

\subsection{PiL with Goal-conditioned Hindsight Relabeling} \label{sec:hr}

Despite the intuitive appeal of the sparse PiL feedback reward $r^{s}$ proposed in Section~\ref{section:pil} for providing preferences between higher-level trajectories, due to its sparsity it mostly fails to generate a meaningful reward signal. As a result, learning an appropriate reward function $\widehat{r}_{\phi}$ using $r^s$ as our PiL primitive is unreliable.
%
%
%
To address this issue, we employ hindsight relabeling \citep{andrychowicz2017hindsight} when comparing trajectories $\sigma^{1}$ and $\sigma^{2}$.
Specifically, we first randomly sample a new goal $\widehat{g}$ from the set $\{ s^1_{t+ik}, s^2_{t+ik} \}_{i = 1}^{n-1}$
%
%
%
of states encountered during trajectories $\sigma^1, \sigma^2$, then apply Equation~\eqref{eqn:reward_s} with $g^*$ replaced by $\widehat{g}$:
%
%
%
\begin{equation} \label{eqn:hindsight_relabeling}
    \begin{aligned}
        \small
        \sum_{i=0}^{n-1} & r^s(s^1_{t + ik}, \widehat{g}, g^1_{t + ik}) > \sum_{i=0}^{n-1} r^s(s^2_{t + ik}, \widehat{g}, g^2_{t + ik}).
    \end{aligned}
\end{equation}
Using hindsight relabeling, we are able to generate significantly better preference feedback, resulting in improved sample efficiency and performance during training. We show in Section~\ref{sec:experiments}, Figure \ref{fig:ablation_no_hr} that this simple hindsight relabeling approach significantly boosts performance in complex sparse-reward tasks.

\subsection{Primitive-informed Regularization} \label{primitive_informed_regularization}

As discussed above, the preference reward learned using PiL with hindsight relabeling motivates the higher-level policy to reach the goal $g$, while mitigating non-stationarity. However, the higher-level subgoal predictions $g_t$ may be too difficult for the current lower-level policy, which may stall learning at the lower level (see Figure \ref{fig:success_rate_comparison} for a comparison of PIPER with and without regularization).
%
%
Ideally, the higher-level policy should produce subgoals at an appropriate level of difficulty, according to the current capabilities of the lower primitive. Properly balancing the task split between hierarchical levels is a recurring challenge in HRL.

For a given lower-level policy $\pi^L : \mathcal{S} \times \mathcal{G} \rightarrow \mathcal{A}$, denote the corresponding state value function by $V_{\pi^L}(s,g)$. To encourage appropriate subgoal selection by the higher-level policy, we propose using the lower-level state value function $V_{\pi^{L}}$ to regularize the higher-level policy to predict feasible subgoals for the lower-level policy. Intuitively, $V_{\pi^{L}}(s_t,g_t)$ provides an estimate of the achievability of subgoal $g_t$ from current state $s_t$, since a high value of $V_{\pi^{L}}(s_t,g_t)$ implies that the lower level expects to achieve high reward for subgoal $g_t$. Let $r^s(s, g^*, g)$ denote the parameterized reward model corresponding to the preference data, as defined in Section~\ref{section:pil}. For a trajectory $\tau$ of length $T$, consider the following KL-regularized formulation of preference-based learning, where $\pi_{reg}$ is the regularizing policy for the higher-level policy $\pi^{H}$:
\begin{equation}
\label{eqn:dpo_input}
    \max_{\pi^{H}} \mathbb{E}_{\pi^{H}}[\sum_{t=0}^{T}(r^s(s, g^*, g)-\beta \mathbb{D}_{\mathrm{KL}}[\pi^{H}(\cdot|s_t) \| \pi_{reg}(\cdot|s_t)])],
\end{equation}
where $\beta \geq 0$ is a scalar hyperparameter controlling the deviation from the regularization policy $\pi_{reg}$. We propose the following formulation of the regularization policy:
\begin{equation}
\label{eqn:eqn_reg}
    \pi_{{reg}}(g_t \mid s_t)=\frac{\exp (m(V_{\pi^{L}}(s_t, g_t)))}{\hat{Z}(s_t)},
\end{equation}
where $\hat{Z}(s_t) = \sum_{g_t} \exp (m(V_{\pi^{L}}(s_t, g_t))$, and $m=\frac{\alpha}{\beta}$.
The policy $\pi_{reg}(\cdot | g_t)$ is simply the softmax distribution generating a given subgoal $g_t$ with probability proportional to its value $V_{\pi^{L}}(s_t,g_t)$.
%
%
We substitute~\eqref{eqn:eqn_reg} in~\eqref{eqn:dpo_input} to get
\begin{equation}
\label{eqn:dpo_with_reg_policy}
\begin{split}
     \max_{\pi^{H}}  \mathbb{E}_{\pi^{H}} \left[ \sum_{t=0}^{T}(r^s(s, g^*, g) + \alpha (V_{\pi^{L}}(s_t,g_t)) + \hat{m}(s_t) ) \right],   
\end{split}
\end{equation}
where $\hat{m}(s_t) = \beta \mathcal{H}(s_t) - \beta \log Z(s_t)$, and  $\mathcal{H}(s_t) = - \log \pi^{H}(g_t|s_t)$ is the entropy term for $\pi^{H}$. Following prior work~\cite{levine2018reinforcement,ziebart2008maximum}, we get the following optimal solution for the higher-level policy:
\begin{equation}
\label{eqn:eqn_dpo_optimal_policy}
    \pi^{H}(g_t|s_t) = \frac{1}{Z(s_t)} \exp (\frac{1}{\beta} (r^s(s, g^*, g) + \alpha (V_{\pi^{L}}(s_t,g_t)))),
\end{equation}
where $Z(s_t) = \sum_{g_t} \exp (\frac{1}{\beta} (r^s(s, g^*, g) + \alpha (V_{\pi^{L}}(s_t,g_t))))$ is the partition function and $\alpha$ is the primitive regularization weight hyperparameter. Appendix \ref{appendix:pi_star_derivation} contains the complete derivation. Notice that this optimal policy $\pi^{H}(g_t|s_t)$ assigns high probability to subgoals $g_t$ which maximize the regularized reward $r^{total}(s, g^*, g)=r^s(s, g^*, g) + \alpha (V_{\pi^{L}}(s_t,g_t))$.
%
%
If we use $r^{total}$ to generate preferences between trajectories instead of the standard preferences outlined in Section~\ref{subsec:pbl}, we end up with the optimal policy under this primitive-regularized, preference-based learning scheme.
Hence, we substitute $r^{total}(s, g^*, g)$ into inequality \eqref{eqn:hindsight_relabeling} to yield our final preference condition for determining if $\delta^1 \succ \delta^2$:
%
%
    \begin{align}\label{eqn:hr_with_reg}
        \sum_{i=0}^{n-1} r^{total}(s^1_{t + ik}, \widehat{g}, g^1_{t + ik}) > \sum_{i=0}^{n-1} r^{total}(s^2_{t + ik}, \widehat{g}, g^2_{t + ik}).
    \end{align}
%
%
%
PIPER uses the preferences elicited by \eqref{eqn:hr_with_reg} to learn the reward function $\widehat{r}_{\phi}$, which is in turn used to perform reward relabeling of the higher-level replay buffer. As illustrated in Figure \ref{fig:success_rate_comparison}, value regularization can lead to significantly improved performance in certain tasks, while leading to minimal performance degradation in others.
%

\subsection{PIPER Implementation}

\textbf{Reward stabilization using target networks:}
To mitigate potential training instability when the reward model $\widehat{r}_{\phi}$ is learned using preference-based learning, we utilize target networks with soft target updates \citep{lillicrap2015continuous}. In practice, we found this to greatly stabilize learning.

\textbf{Pseudo-code details:} We explain our approach in detail in Algorithm~\ref{alg:algo_piper}. The higher- and lower-level policies are both trained using Soft Actor Critic (SAC)~\citep{DBLP:journals/corr/abs-1801-01290}. The rewards in higher level transitions are relabeled using the reward model $\widehat{r}_{\phi}$. Instead of relabeling all replay buffer transitions, we relabel the higher-level replay buffer transitions as they are sampled while training.

\section{Experiments}
\label{sec:experiments}

\begin{figure*}
\centering
\captionsetup{font=footnotesize,labelfont=scriptsize,textfont=scriptsize}
\subfloat[][Maze navigation]{\includegraphics[scale=0.185]{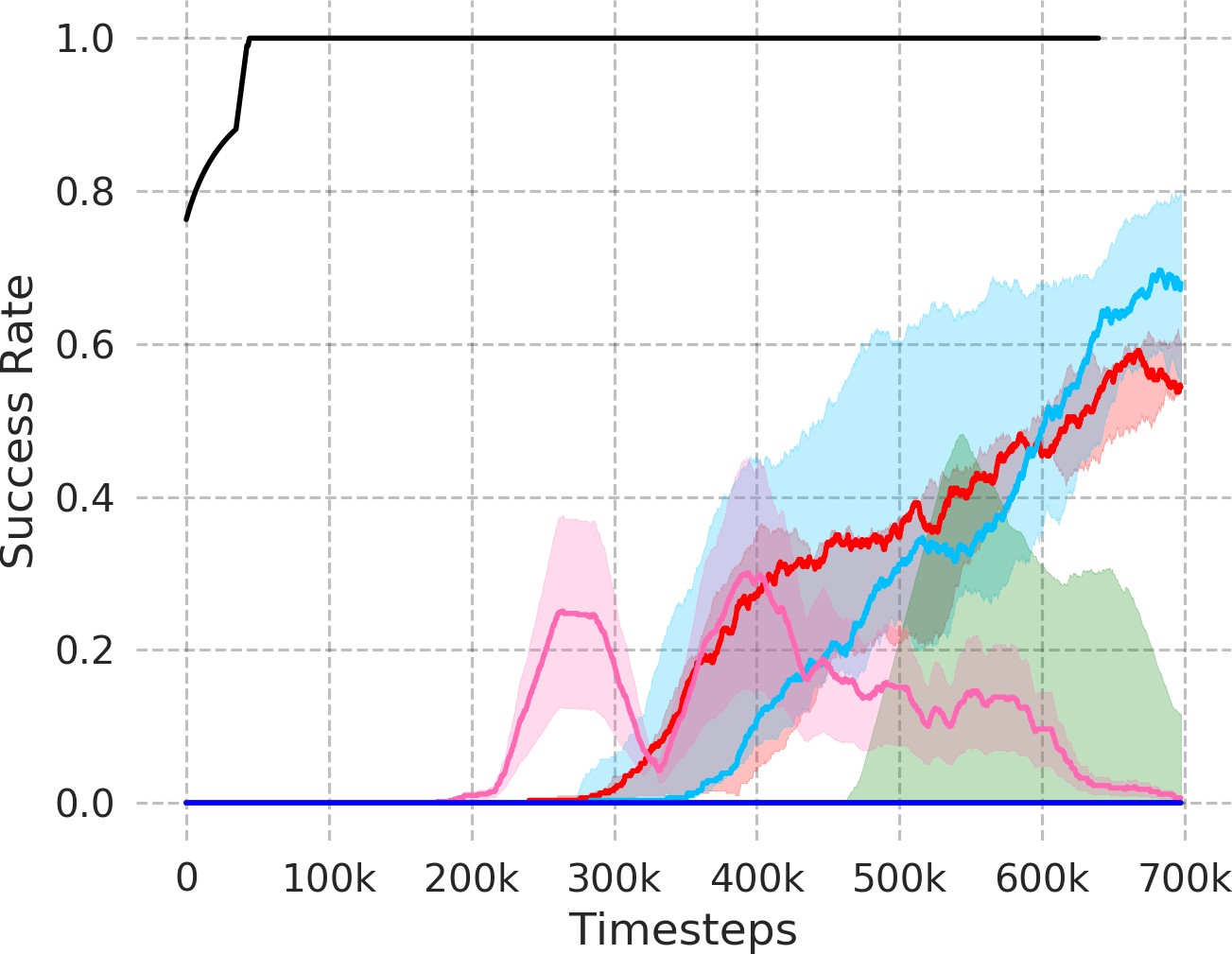}}
\subfloat[][Pick and place]{\includegraphics[scale=0.185]{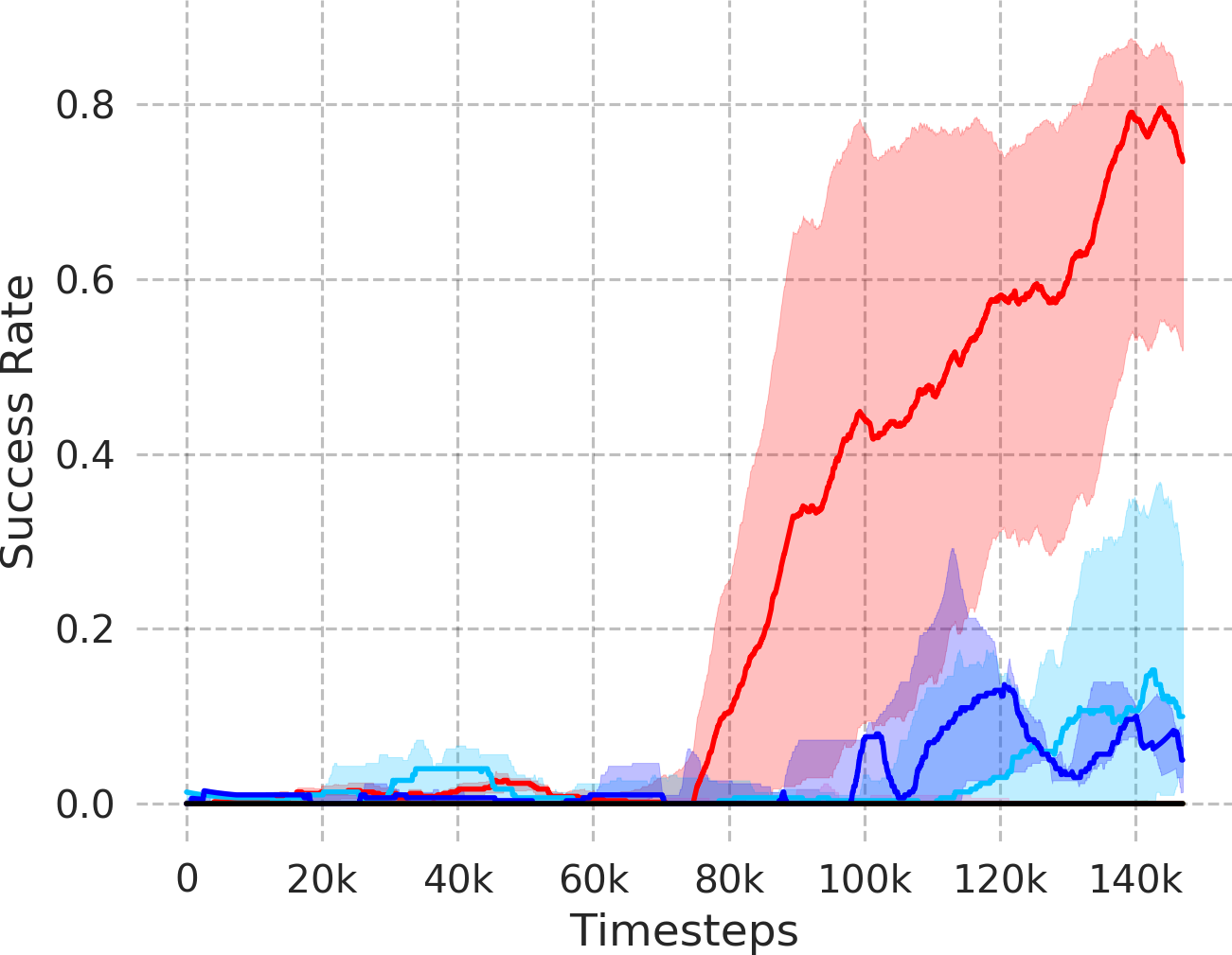}}
\subfloat[][Push]{\includegraphics[scale=0.185]{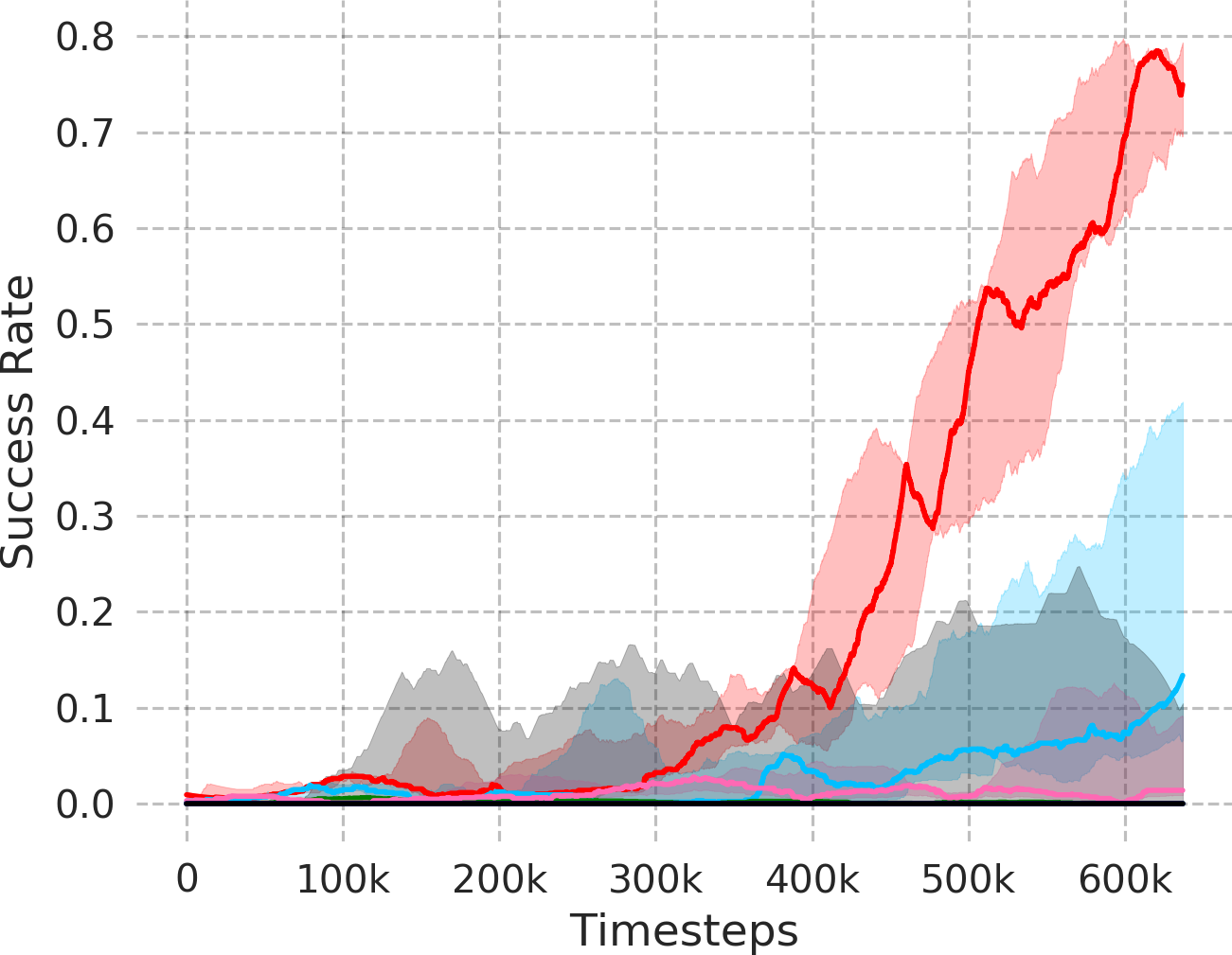}}
\subfloat[][Hollow]{\includegraphics[scale=0.185]{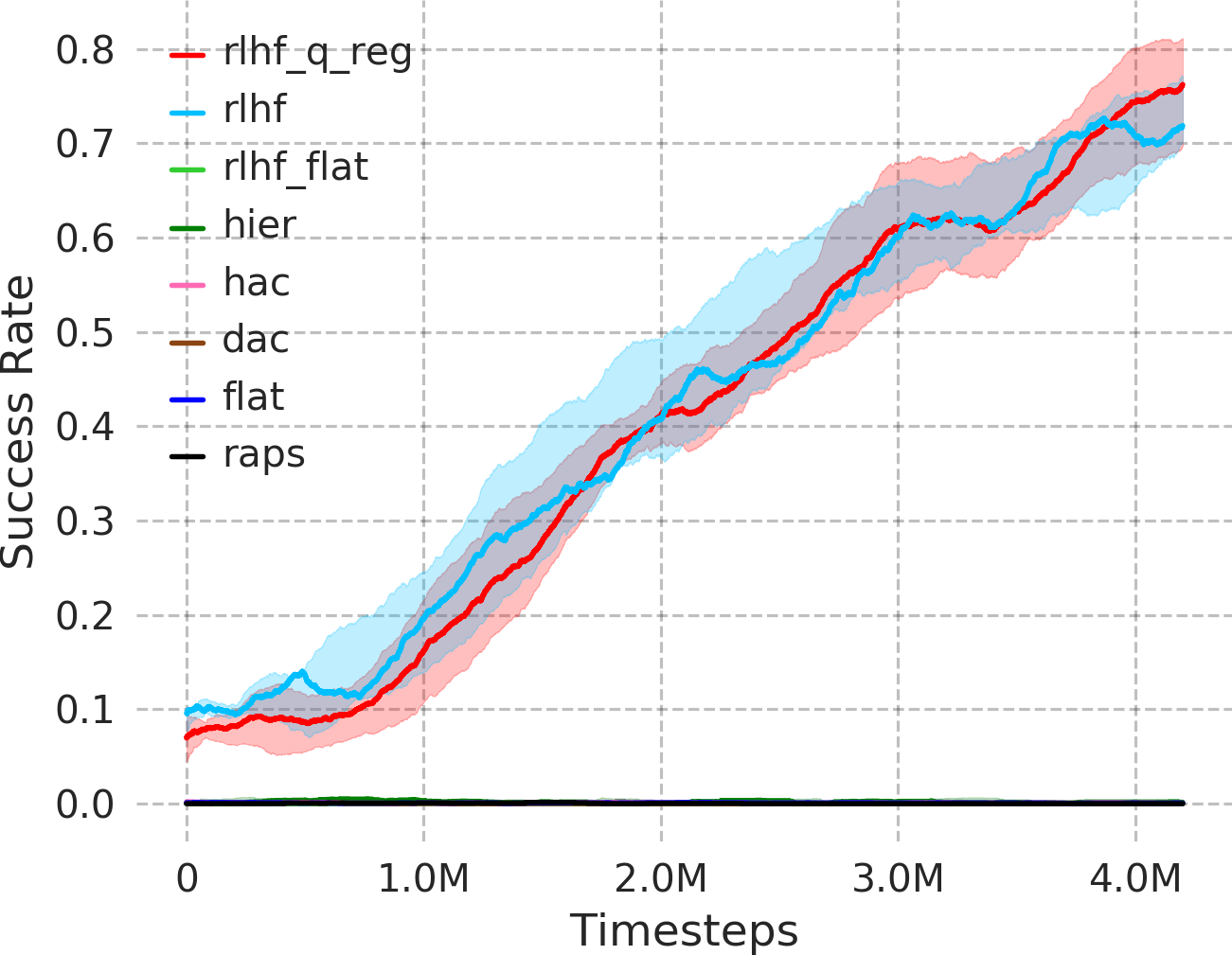}}
\subfloat[][Kitchen]{\includegraphics[scale=0.185]{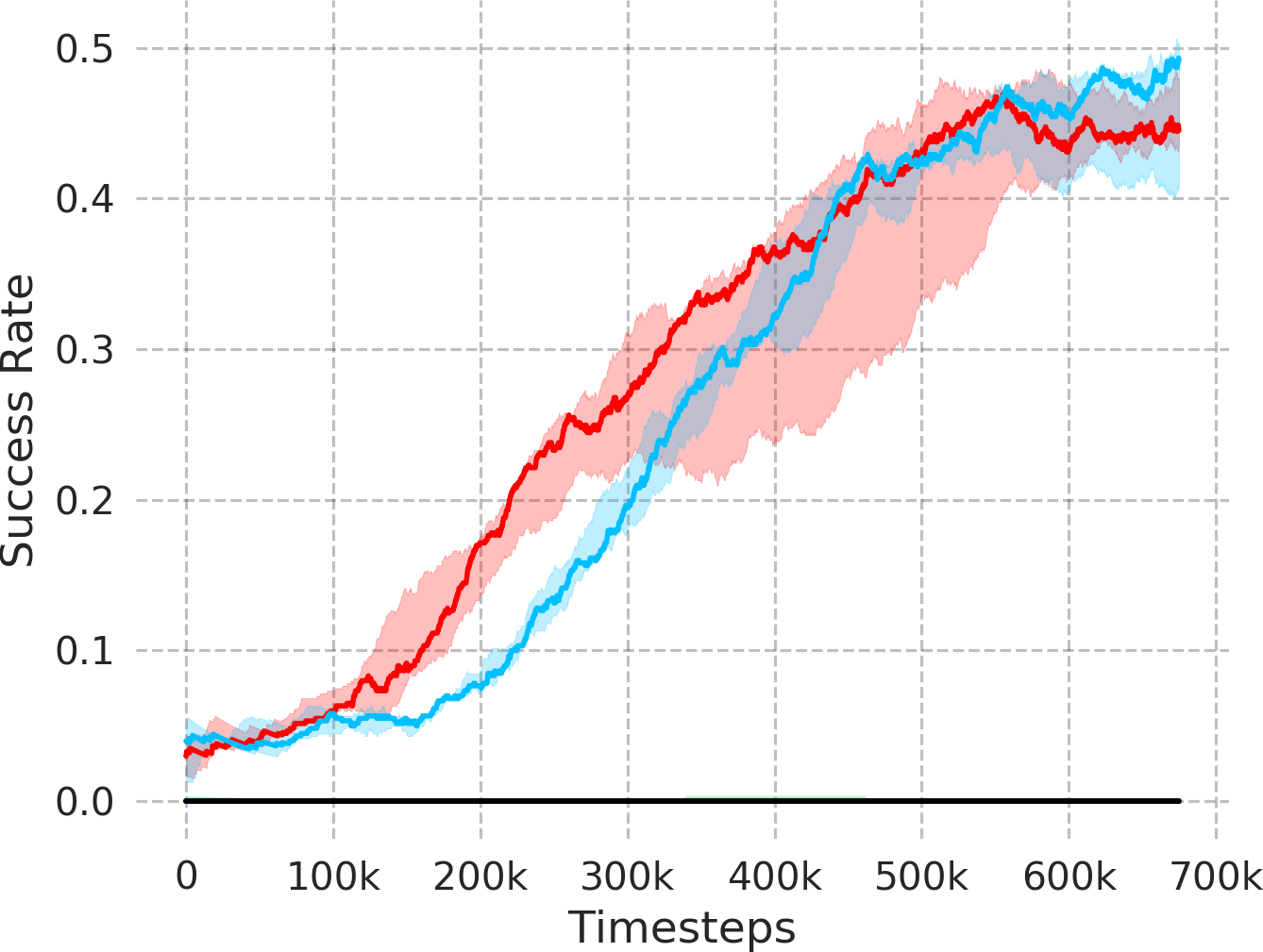}}
\\
{\includegraphics[scale=0.5]{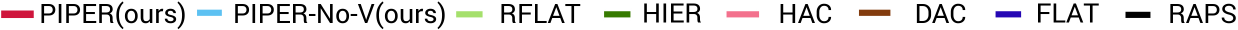}}
\caption{\textbf{Success rate comparison} This figure compares the success rate performances on four sparse maze navigation and robotic manipulation environments. The solid line and shaded regions represent the mean and standard deviation, across $5$ seeds. We compare our approach PIPER against multiple baselines. As can be seen, PIPER shows impressive performance and significantly outperforms the baselines.}
\label{fig:success_rate_comparison}
\end{figure*}
\begin{figure*}
\centering
\captionsetup{font=footnotesize,labelfont=scriptsize,textfont=scriptsize}
\subfloat[][Maze navigation]{\includegraphics[scale=0.185]{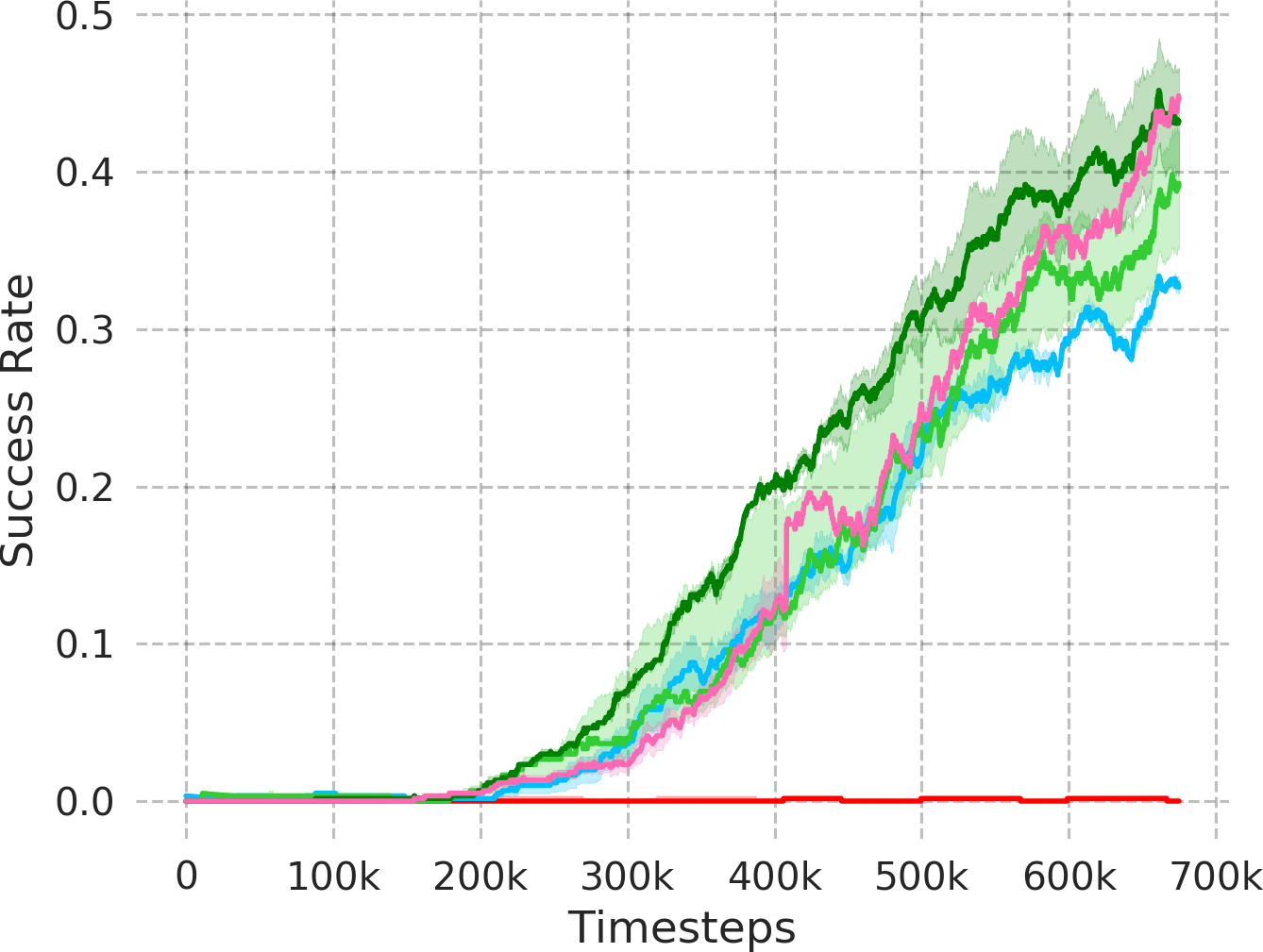}}
\subfloat[][Pick and place]{\includegraphics[scale=0.185]{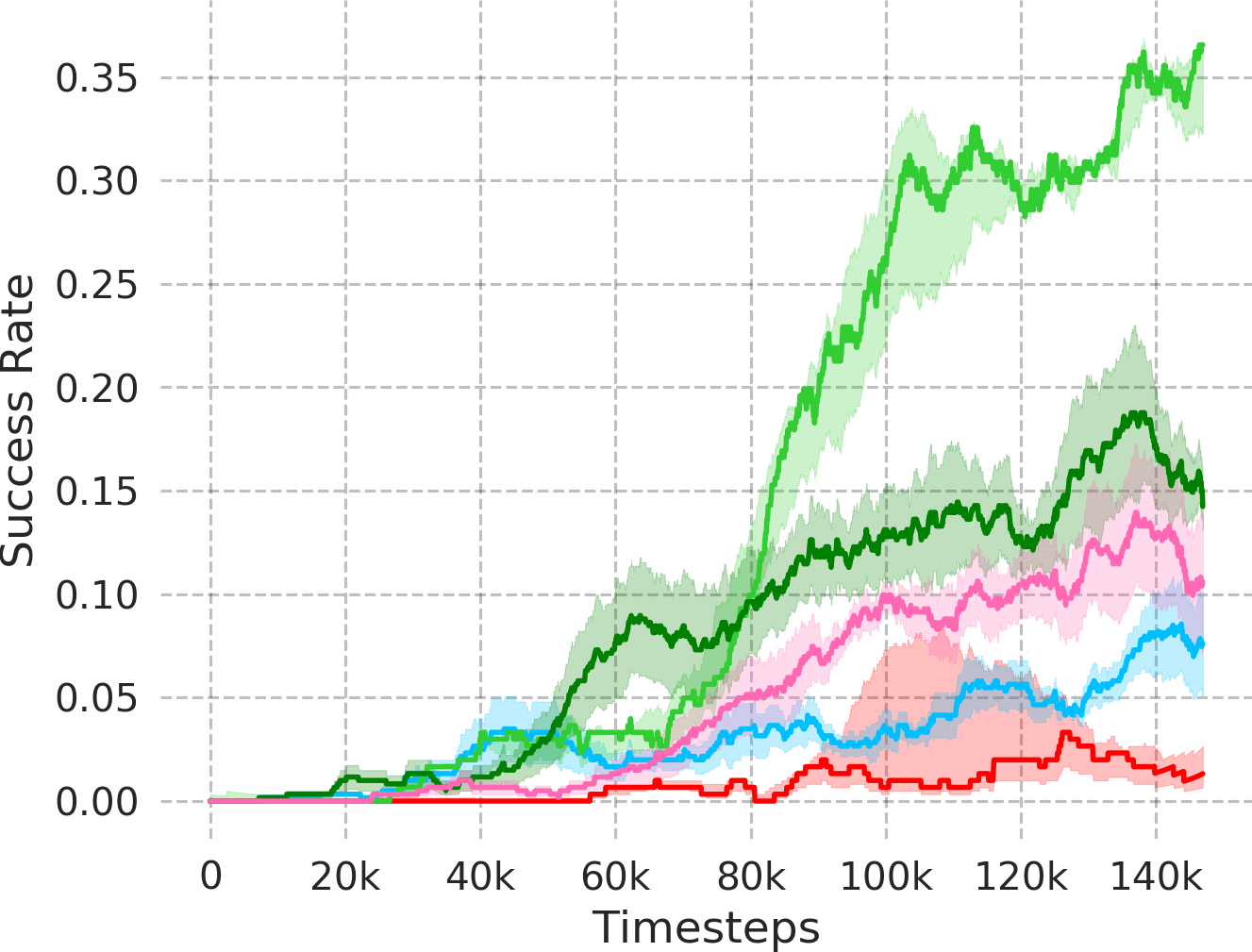}}
\subfloat[][Push]{\includegraphics[scale=0.185]{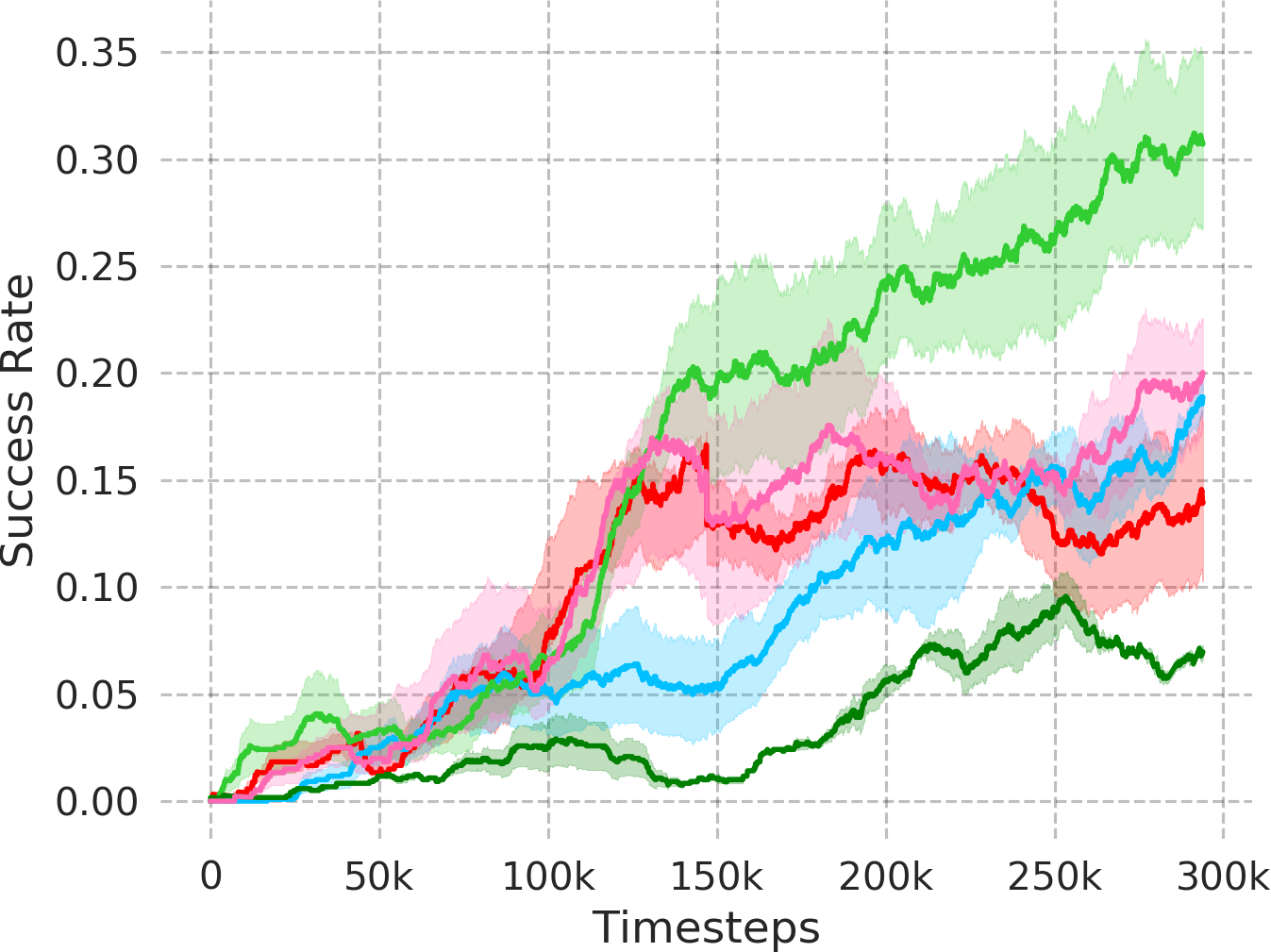}}
\subfloat[][Hollow]{\includegraphics[scale=0.185]{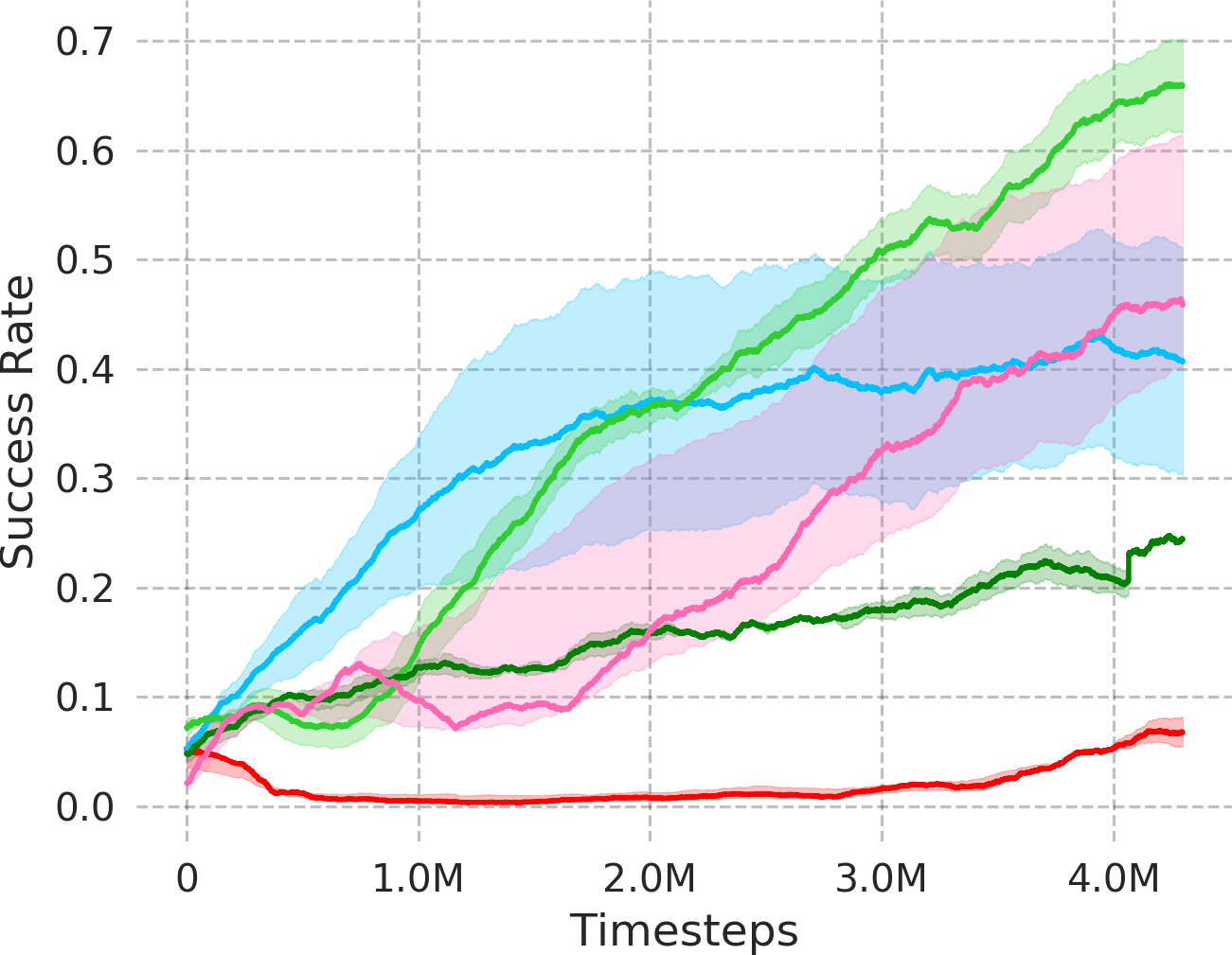}}
\subfloat[][Kitchen]{\includegraphics[scale=0.185]{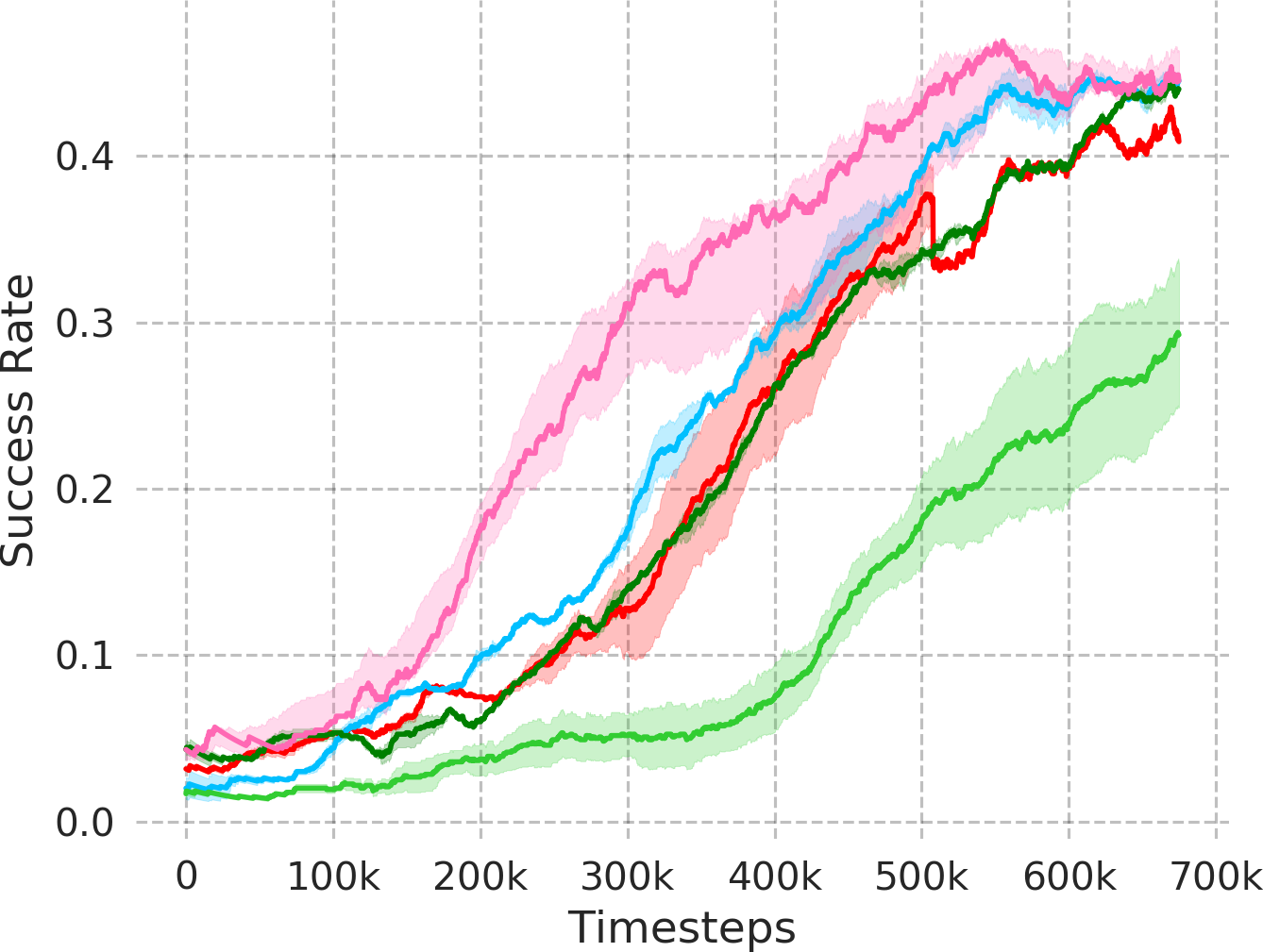}}
\\
{\includegraphics[scale=0.5]{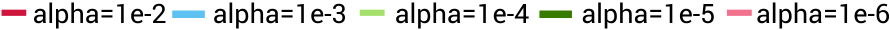}}
\caption{\textbf{Learning rate $\alpha$ ablation} This figure compares the success rate performances for various values of primitive informed regularization weight $\alpha$ hyper-parameter. If $\alpha$ is too small, we loose the advantages of primitive informed regularization, leading to degrading performance. In contrast, if $\alpha$ is too large, it may lead to degenerate solutions. Thus, these success rate performance plots demonstrate that proper primitive subgoal regularization is crucial for appropriate subgoal prediction, and improving overall performance.}
\label{fig:ablation_alpha}
\end{figure*}
\begin{figure*}
\centering
\captionsetup{font=footnotesize,labelfont=scriptsize,textfont=scriptsize}
\subfloat[][Maze navigation]{\includegraphics[scale=0.185]{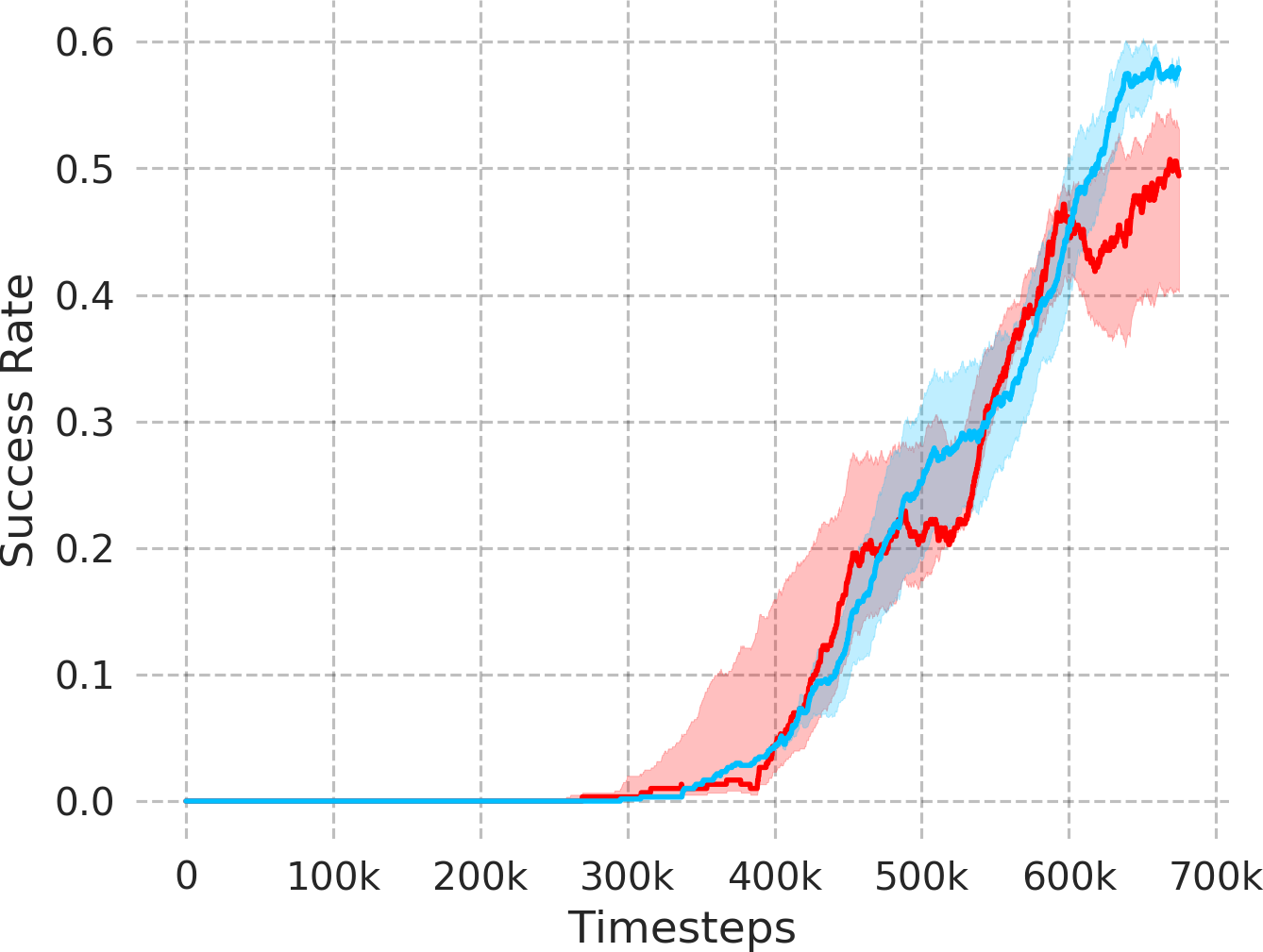}}
\subfloat[][Pick and place]{\includegraphics[scale=0.185]{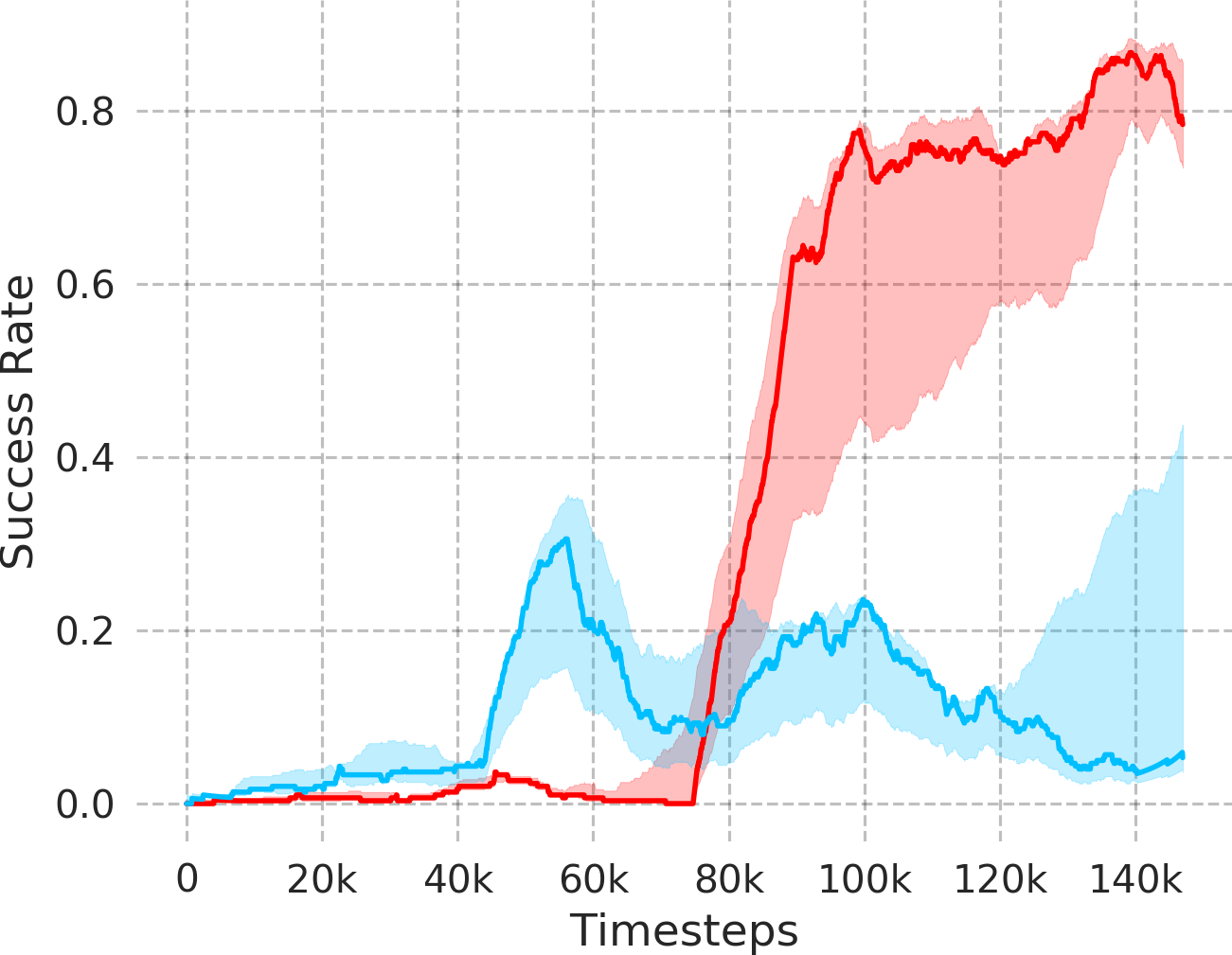}}
\subfloat[][Push]{\includegraphics[scale=0.185]{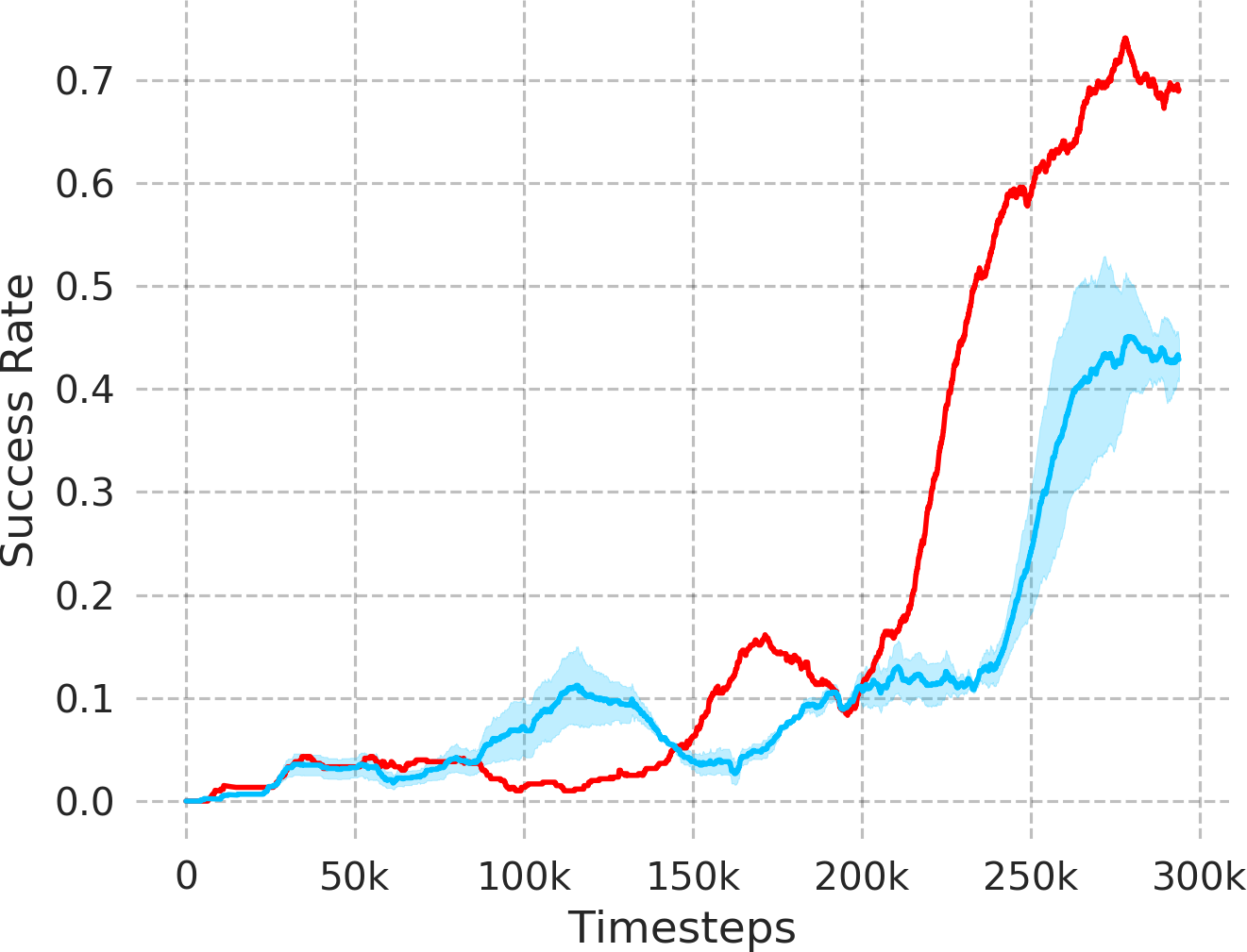}}
\subfloat[][Hollow]{\includegraphics[scale=0.185]{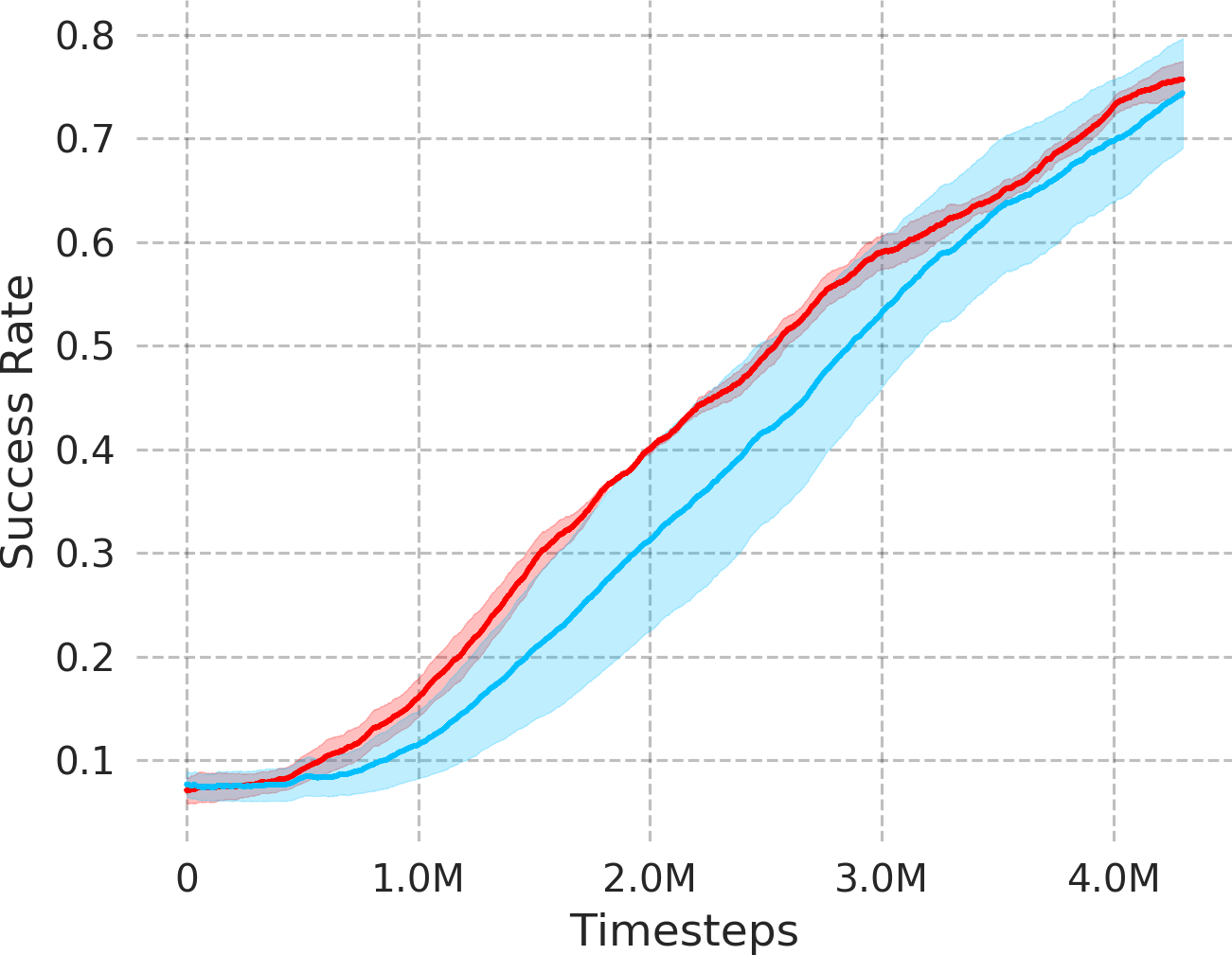}}
\subfloat[][Kitchen]{\includegraphics[scale=0.185]{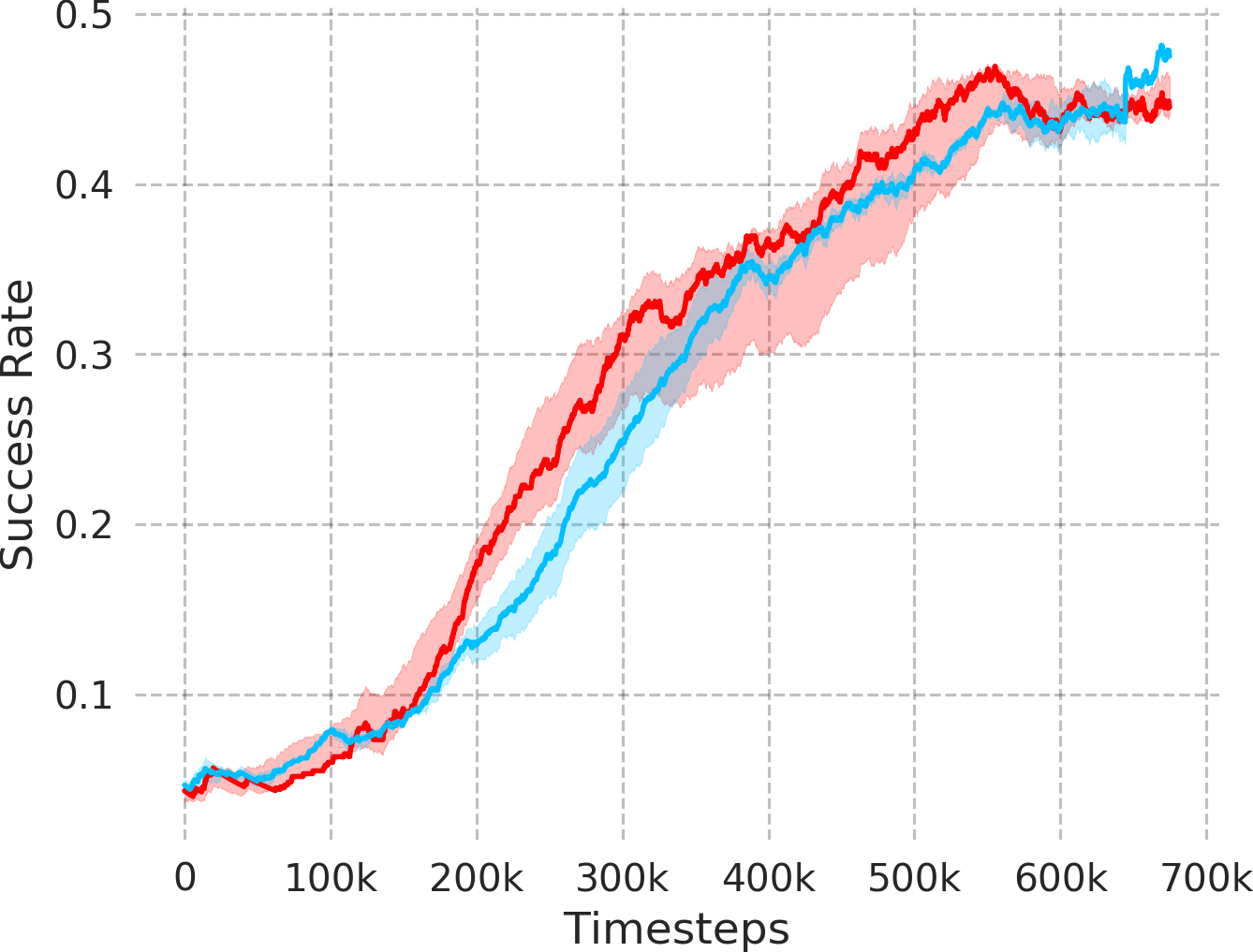}}
\\
{\includegraphics[scale=0.5]{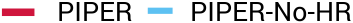}}
\caption{\textbf{Hindsight Relabeling ablation} This figure compares the performance of our PIPER approach with PIPER-No-HR ablation, which is effectively PIPER without hindsight relabeling~\citep{andrychowicz2017hindsight} (as explained in Section~\ref{sec:hr}). The plots showcase that although hindsight relabeling demonstrates minor performance improvement in sparse maze and kitchen tasks, it provides significant training speedup in sparse pick and place and push environments.}
\label{fig:ablation_no_hr}
\end{figure*}
\begin{figure*}
\centering
\captionsetup{font=footnotesize,labelfont=scriptsize,textfont=scriptsize}
\subfloat[][Maze navigation]{\includegraphics[scale=0.185]{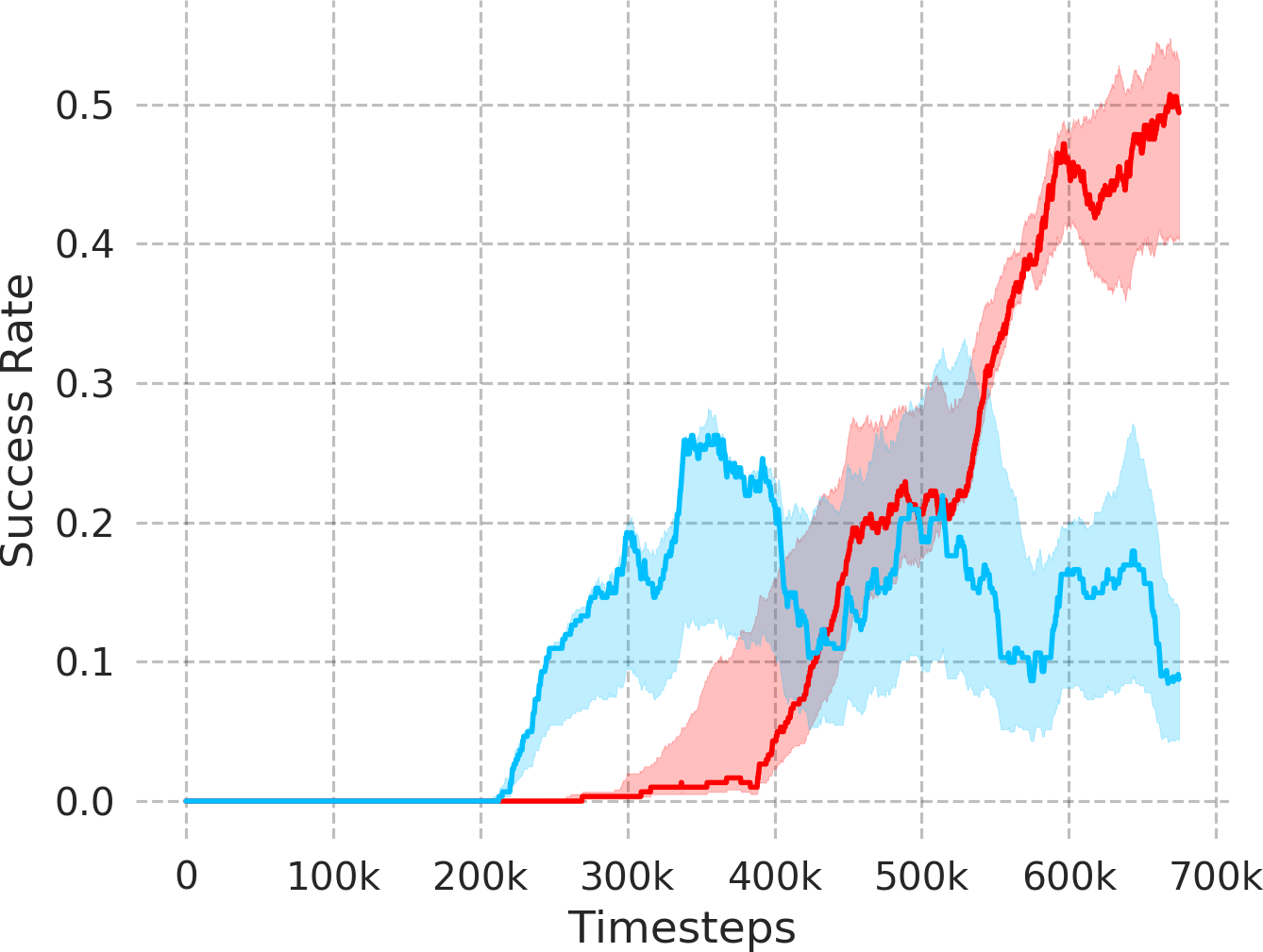}}
\subfloat[][Pick and place]{\includegraphics[scale=0.185]{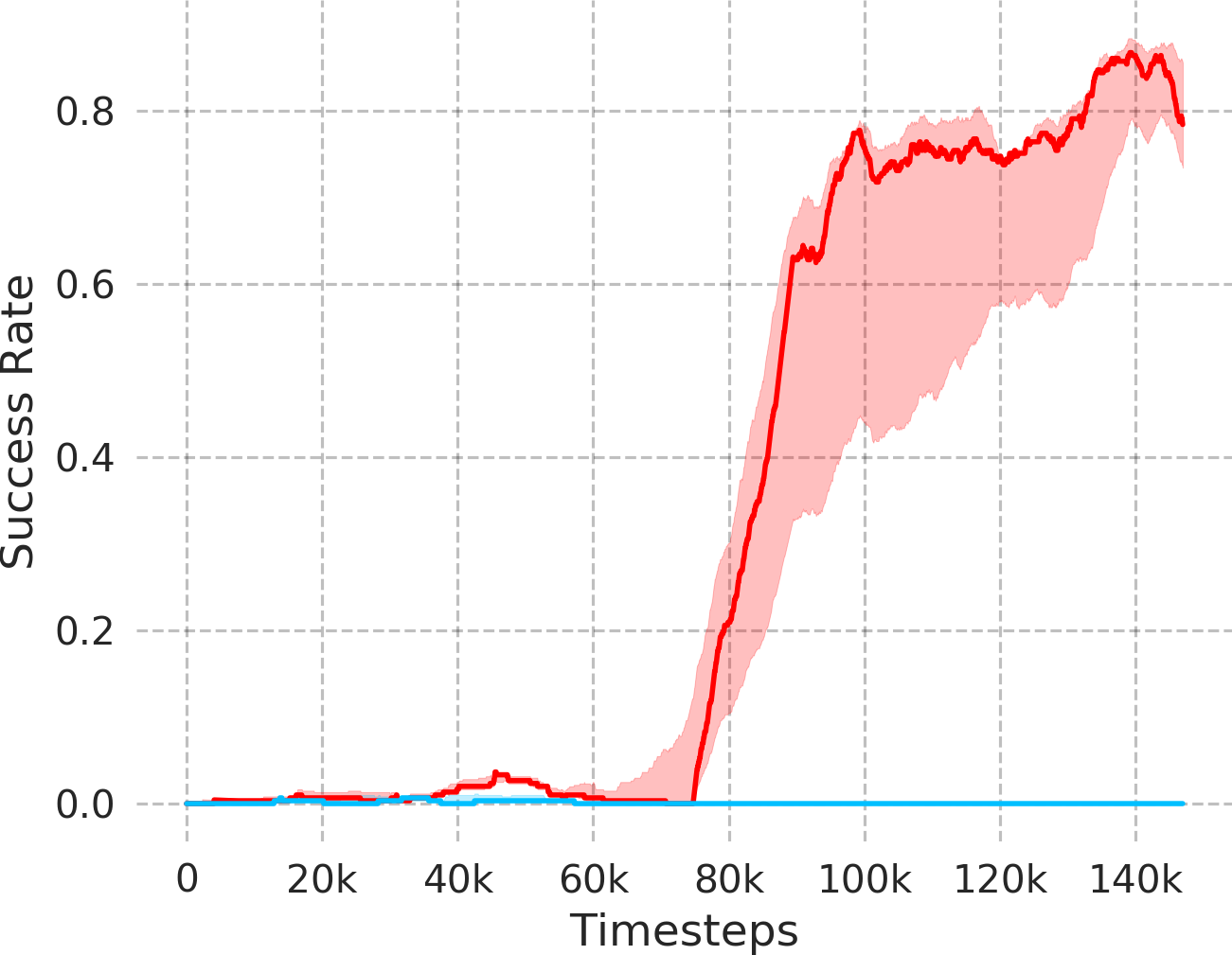}}
\subfloat[][Push]{\includegraphics[scale=0.185]{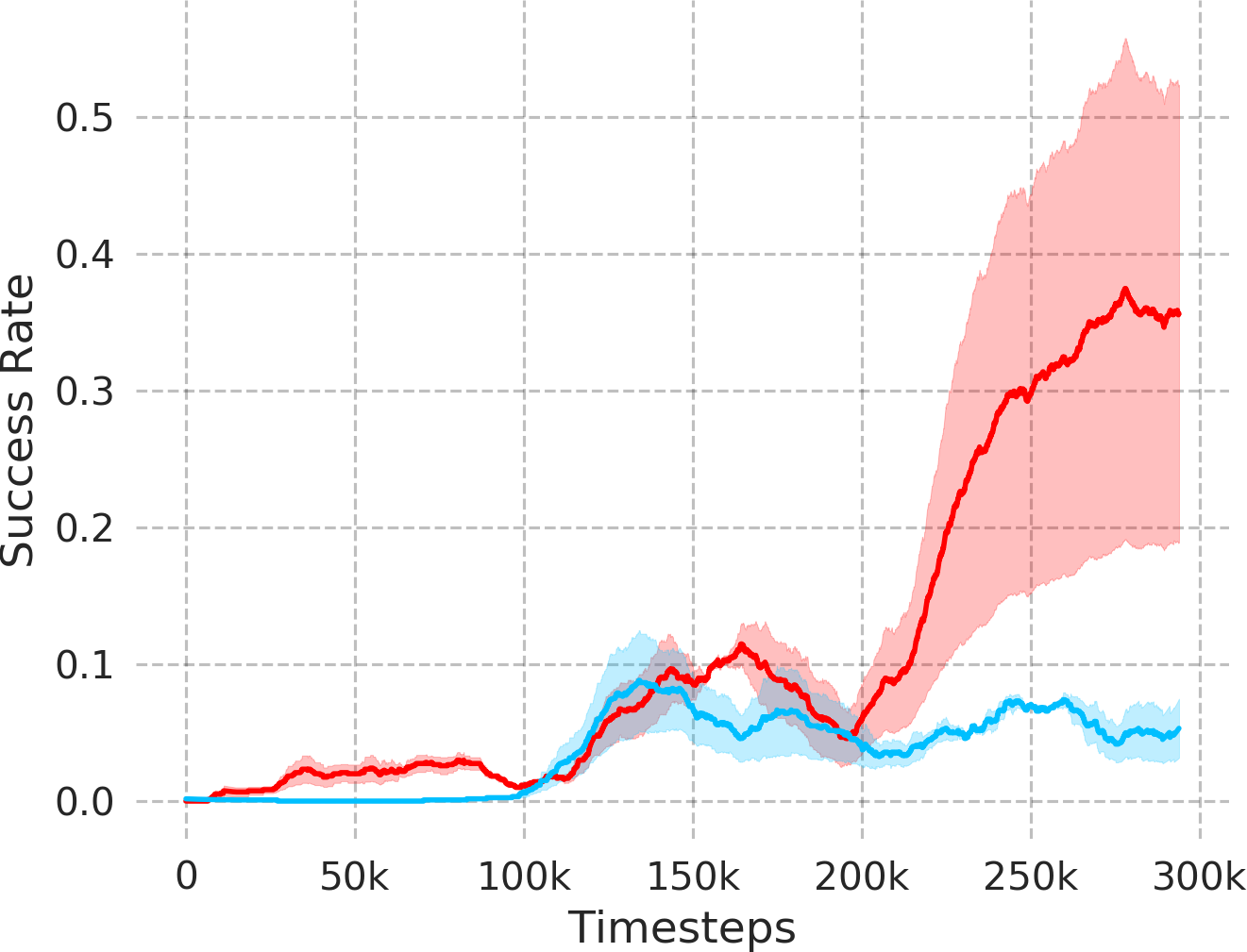}}
\subfloat[][Hollow]{\includegraphics[scale=0.185]{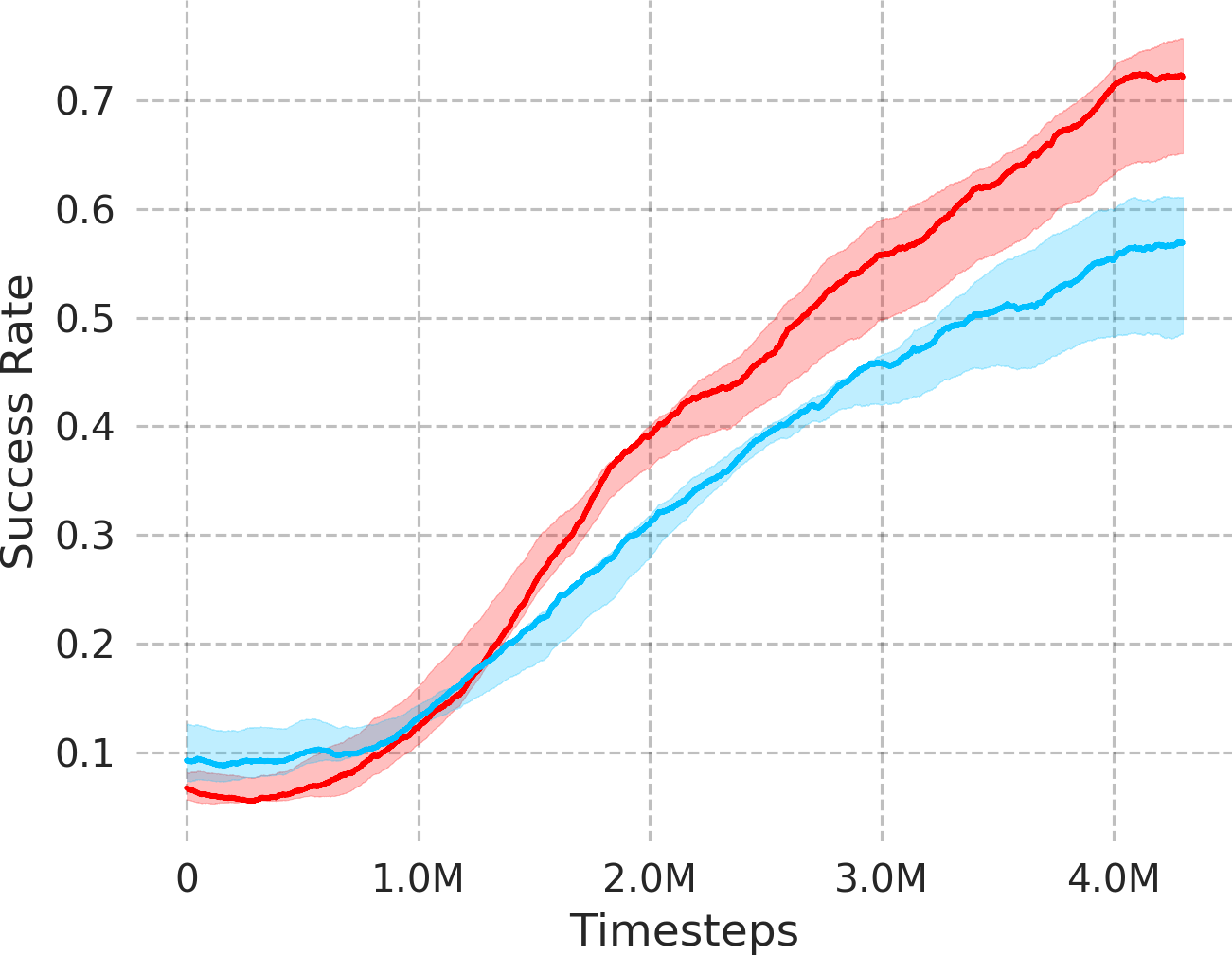}}
\subfloat[][Kitchen]{\includegraphics[scale=0.185]{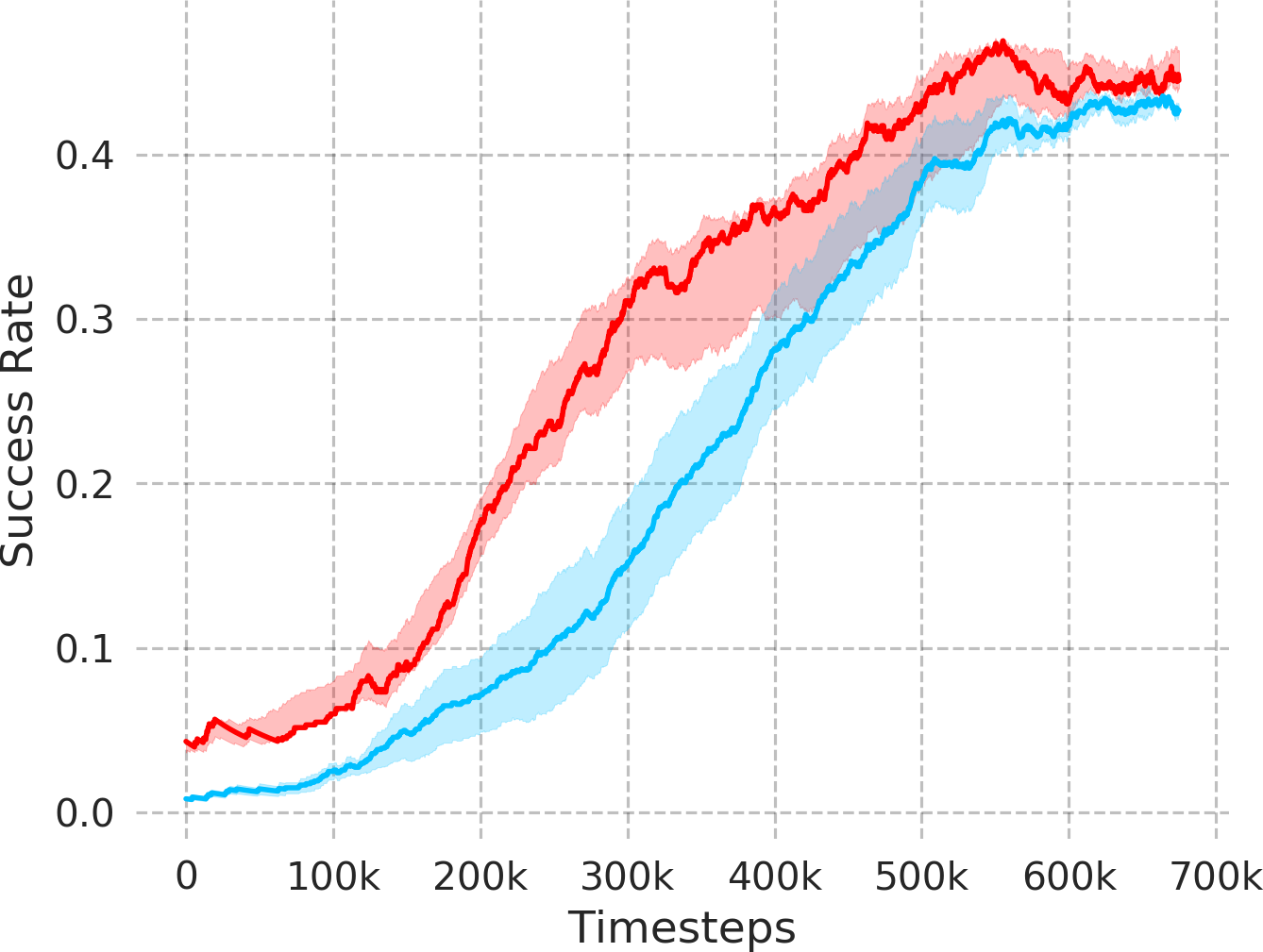}}
\\
{\includegraphics[scale=0.5]{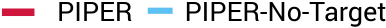}}
\caption{\textbf{Target networks ablation} This figure compares the performance of our PIPER approach with PIPER-No-Target ablation, which is effectively PIPER without target networks implementation~\citep{lillicrap2015continuous}. The plots showcase that using target networks significantly improves performance and indeed reduces training instability caused by non-stationary reward models $\widehat{r}_{\phi}$, learnt using preference based learning.}
\label{fig:ablation_target}
\end{figure*}
\begin{figure*}
\centering
\captionsetup{font=footnotesize,labelfont=scriptsize,textfont=scriptsize}
\subfloat[][Maze navigation]{\includegraphics[scale=0.185]{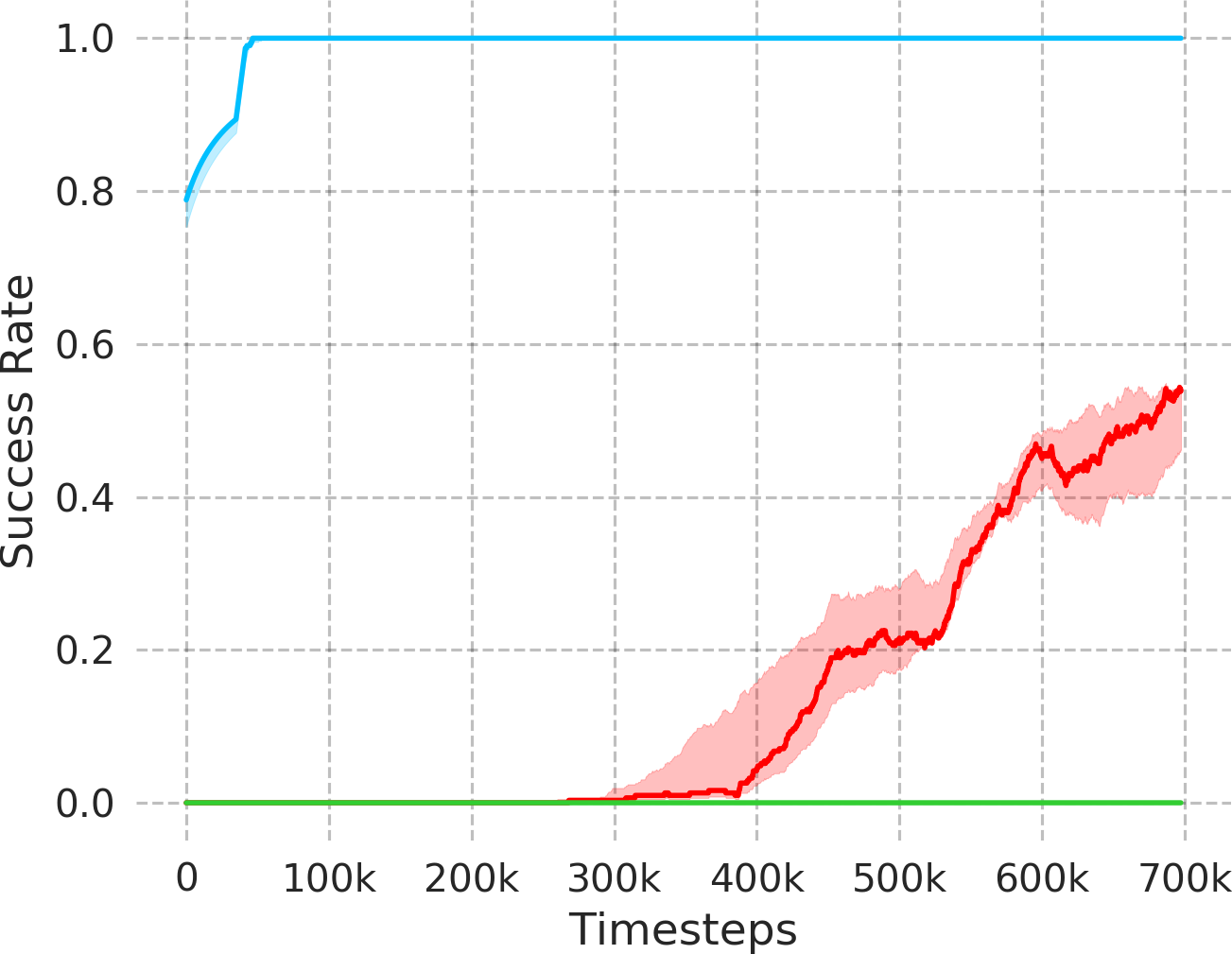}}
\subfloat[][Pick and place]{\includegraphics[scale=0.185]{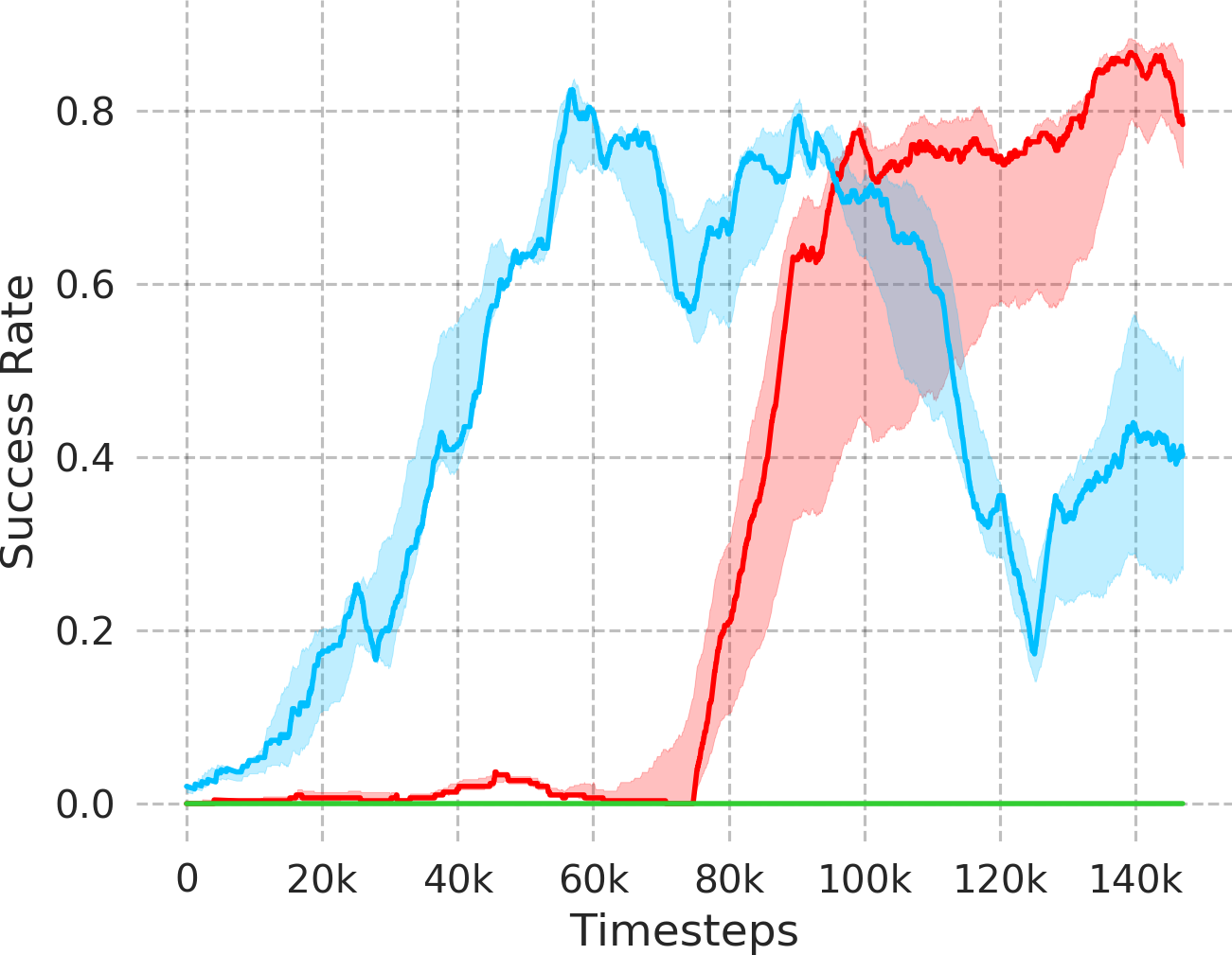}}
\subfloat[][Push]{\includegraphics[scale=0.185]{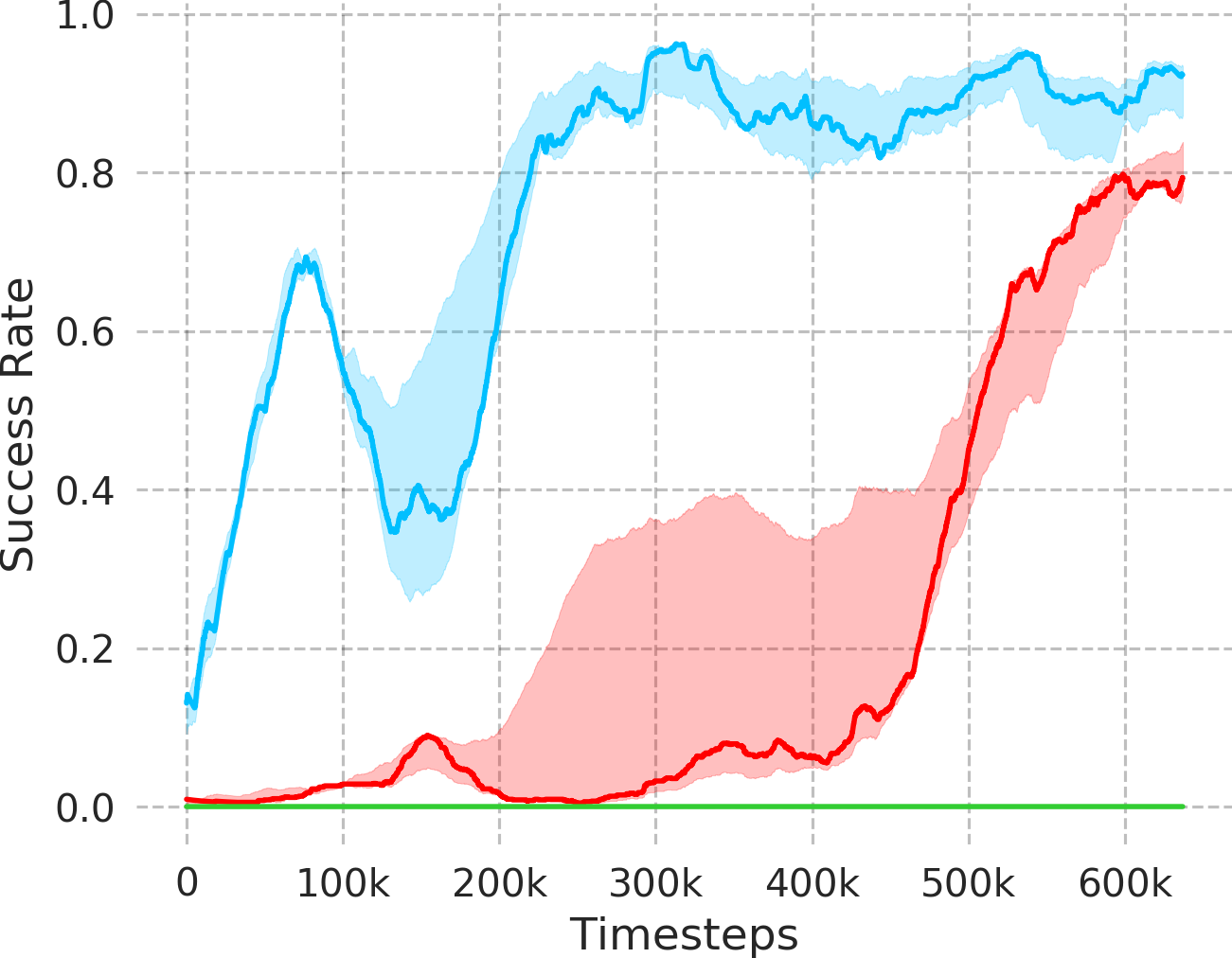}}
\subfloat[][Hollow]{\includegraphics[scale=0.185]{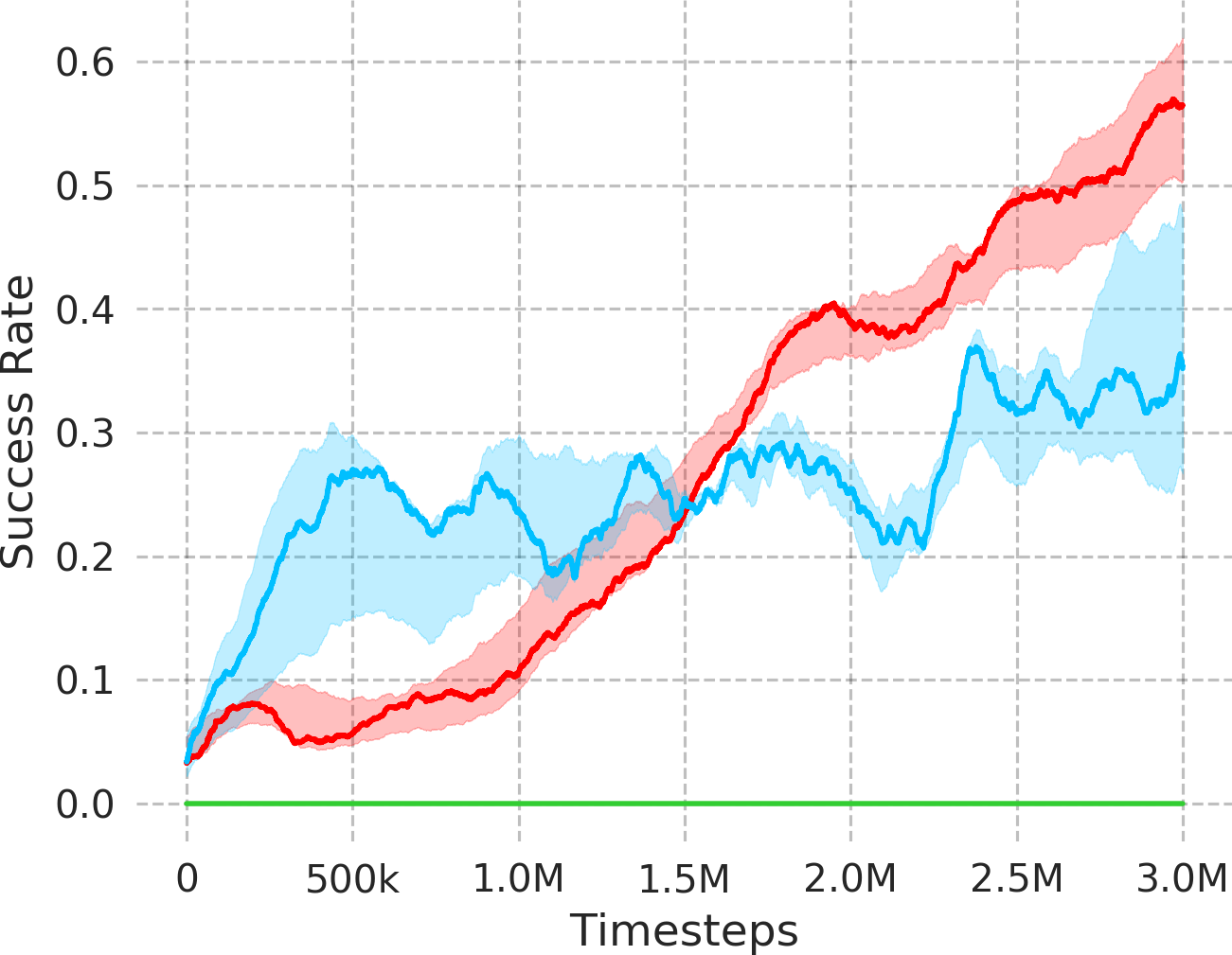}}
\subfloat[][Kitchen]{\includegraphics[scale=0.185]{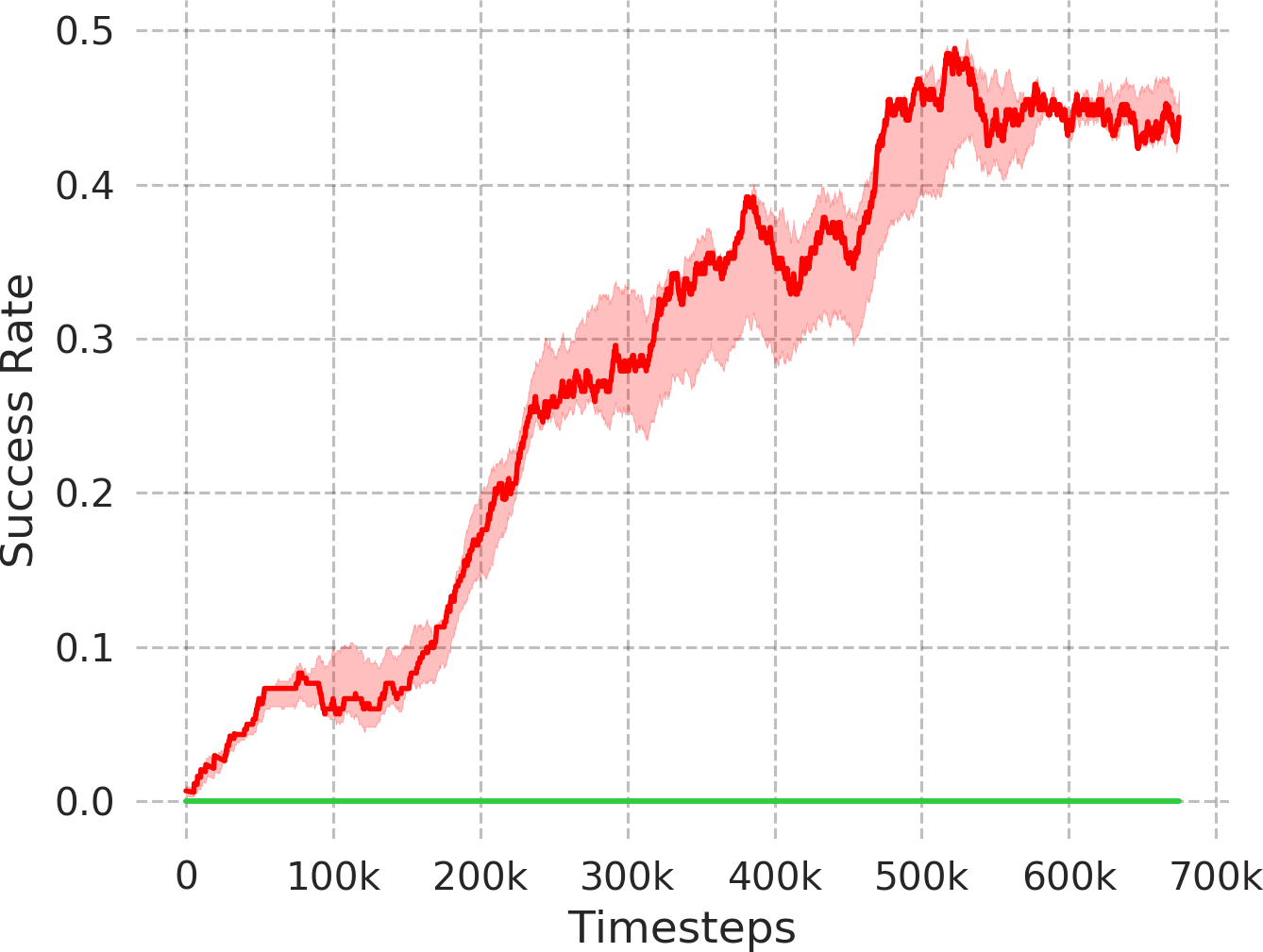}}
\\
{\includegraphics[scale=0.5]{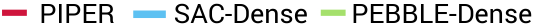}}
\caption{\textbf{Dense rewards ablation} This figure compares the success rate performances between PIPER and two baselines based on dense rewards: $(i)$ SAC-Dense, and $(ii)$ PEBBLE-Dense. Although SAC-Dense outperforms PIPER on easier tasks, PIPER is able to outperform SAC-Dense in harder tasks. Notably, PEBBLE-Dense is unable to solve any of the tasks. Thus, PIPER shows impressive performance, and is a viable approach in complex sparse reward scenarios.}
\label{fig:ablation_dense_reward}
\end{figure*}

We perform experiments to empirically investigate the following questions:
%
%
\textbf{(1)} How well does PIPER perform in sparse maze navigation and robotic manipulation tasks?
\textbf{(2)} Is PIPER able to mitigate the recurring issue of non-stationarity in HRL?
\textbf{(3)} Does PIPER outperform flat preference-based learning?
\textbf{(4)} Does PIPER enhance sample efficiency and training stability?
\textbf{(5)} What is the contribution of each of the design choices in PIPER?

\subsection{Setup} 
We evaluate PIPER on four robotic navigation and manipulation tasks: $(i)$ maze navigation, $(ii)$ pick and place~\citep{andrychowicz2017hindsight}, $(iii)$ push, $(iv)$ hollow, and $(v)$ franka kitchen~\citep{gupta2019relay}. Since we focus on sparse reward scenarios, we re-implement these environments as sparse reward environments. Notably, since the pick and place, push, hollow and kitchen task environments are difficult, in order to speedup training, we assume access to a single human demonstration, and use an additional imitation learning objective at the lower level. We do not assume access to any demonstration in the maze navigation task. We keep this assumption consistent among all baselines to ascertain fair comparisons. We provide additional implementation details in Appendix~\ref{sec:appendix_env_details}. 
\par For environments $(i)-(v)$ ($(i)$ maze navigation, $(ii)$ pick and place~\citep{andrychowicz2017hindsight}, $(iii)$ push, $(iv)$ hollow, and $(v)$ franka kitchen~\citep{gupta2019relay}), the maximum task horizon $\mathcal{T}$ is set to $225$, $50$, $50$, $100$, $225$ timesteps, respectively, and the lower primitive is allowed to execute for $k = 15, 7, 7, 10$ and $15$ timesteps, respectively. For environments $(i)-(v)$, the primitive regularization weight hyper-parameter $\alpha$ is set as $10^{-5}$, $10^{-4}$, $10^{-4}$, $10^{-4}$, $10^{-6}$ in environments $(i)-(v)$, respectively (see Figure~\ref{fig:ablation_alpha}). In our experiments, we use off-policy Soft Actor Critic (SAC)~\citep{DBLP:journals/corr/abs-1801-01290} for optimizing RL objective, using the Adam~\citep{kingma2014adam} optimizer. The actor and critic networks are formulated as three-layer, fully connected neural networks with $512$ neurons in each layer. The experiments are run for 6.75e5, 1.5e5, 6.5E5, and 6.75e5 timesteps in environments $(i)-(iv)$, respectively. 
In our experiments, we use off-policy Soft Actor Critic (SAC)~\citep{DBLP:journals/corr/abs-1801-01290} for optimizing RL objective, using the Adam~\citep{kingma2014adam} optimizer. The actor and critic networks are formulated as three-layer, fully connected neural networks with $512$ neurons in each layer. The experiments are run for 6.75e5, 1.5e5, 6.5E5, and 6.75e5 timesteps in environments $(i)-(iv)$, respectively.
\par In the maze navigation task, the closed gripper (fixed at table height) of a $7$-degree-of-freedom (7-DoF) robotic arm has to navigate across a four room maze to the goal position. In pick and place task, a $7$-DoF robotic arm gripper has to navigate to the square block, pick it up and bring it to the goal position. In the push task, a $7$-DoF robotic arm gripper has to push the square block towards the goal position. In hollow task, the $7$-DOF robotic arm gripper has to pick a square hollow block and place it such that a fixed vertical structure on the table goes through the hollow block. In the kitchen task, a $9$-DoF franka robot has to navigate and open the microwave door. To ensure fair comparisons, we keep parameters including neural network layer width, number of layers, choice of optimizer, SAC implementation parameters, etc., consistent across all baselines. For implementing the RAPS baseline, we use the following lower-level behaviors: in maze navigation, we design a single primitive, \textit{reach}, where the lower-level primitive travels in a straight line towards the subgoal predicted by higher level. In the pick and place, push and hollow tasks, we design three primitives: \textit{gripper-reach}, where the gripper goes to given position $(x_i, y_i, z_i)$; \textit{gripper-open}, which opens the gripper; and \textit{gripper-close}, which closes the gripper. In the kitchen environment, we use the action primitives implemented in RAPS~\citep{dalal2021accelerating}.

\subsection{Evaluation and Results}
\label{sec:results}
In order to analyse our design choices, we consider multiple baselines: PIPER-No-V (PIPER without primitive informed regularization, Section~\ref{primitive_informed_regularization}), RFLAT (Single-level PEBBLE~\citep{christiano2017deep} implementation with target networks), HIER (vanilla hierarchical SAC implementation), HAC (Hindsight Actor Critic~\citep{levy2018learning}), DAC (Discriminator Actor Critic~\citep{kostrikov2018discriminator}), FLAT (Single-level SAC), and RAPS~\cite{dalal2021accelerating}. In Figure~\ref{fig:success_rate_comparison}, we compare the success rate performances of PIPER with hierarchical and non-hierarchical baselines. To illustrate the advantage of primitive informed regularization, we implemented PIPER-No-V, which is PIPER without primitive-informed regularization. As seen in Figure~\ref{fig:success_rate_comparison}, though PIPER significantly outperforms PIPER-No-V baseline in pick and place and push tasks, it only slightly outperforms PIPER-No-V in the maze, hollow and kitchen environments. This illustrates that primitive-informed regularization successfully encourages the higher-level policy to generate subgoals achievable by lower-level policies.

\par We implemented two other baselines: HAC (Hierarchical Actor Critic)~\citep{andrychowicz2017hindsight}, a hierarchical approach that deals with the non-stationarity issue by relabeling transitions, while assuming an optimal lower primitive; and HIER, which is a vanilla HRL baseline implemented using SAC. As seen in Figure ~\ref{fig:success_rate_comparison}, PIPER is able to significantly outperform both these baselines, illustrating that PIPER is indeed able to mitigate non-stationarity in HRL. We also compare PIPER with RFLAT, which is a re-implementation of PEBBLE~\citep{christiano2017deep} with target networks, and where the preference feedback is generated using PiL approach. Since the environments are sparse, we also augmented PEBBLE with hindsight relabeling in our implementation of RFLAT. As seen in Figure ~\ref{fig:success_rate_comparison}, PIPER is able to significantly outperform RFLAT. This empirically illustrates that our hierarchical preference-based learning approach is able to outperform flat preference-based learning approach.
\par We finally compare against RAPS~\citep{dalal2021accelerating}, a hierarchical approach where the higher-level policy picks from among hand-designed action primitives at the lower level. Importantly, the performance of RAPS depends on the quality of action primitives. We found that RAPS was able to outperform PIPER and all other baselines in the maze navigation environment. We hypothesize that this is because the hand-designed lower primitive used in the maze task makes the task significantly easier for the higher level. However, RAPS did not show progress in the more difficult pick and place, push, hollow and kitchen tasks. We also implemented Discriminator Actor-Critic (DAC)~\citep{kostrikov2018discriminator} with access to a single demonstration in the pick and place, push, hollow and kitchen environments, to analyse how PIPER fares against single-level baselines trained with privileged information. Additionally, we compared PIPER with a single-level RL baseline implemented using SAC. Both baselines failed to show significant progress, while PIPER clearly outperforms the baselines.

\subsection{Ablation Analysis}
\label{sec:ablations}

Here, we empirically demonstrate the importance of our each various component technique, by performing corresponding ablation study. We analyze the effect of varying the weight parameter $\alpha$ in Figure~\ref{fig:ablation_alpha}, and found that setting appropriate value of hyperparameter $\alpha$ is crucial for improved performance. If $\alpha$ is too small, we lose the advantages of primitive-informed regularization, leading to poor performance. In contrast, if $\alpha$ is too large, it leads to degenerate solutions. We also analyze the effect of removing the hindsight relabeling (PIPER-No-HR ablation). In Figure~\ref{fig:ablation_no_hr}, the empirical results show that hindsight relabeling indeed leads to improved performance. Further, we remove the target networks from PIPER (PIPER-No-Target), to analyze whether using target networks stabilizes training. As seen in Figure~\ref{fig:ablation_target}, target networks significantly improve training stability. Finally, in Figure~\ref{fig:ablation_dense_reward}, we compare PIPER against two baselines with well-behaved, hand designed dense reward functions ($(i)$ flat SAC with dense reward function (SAC-Dense), and $(ii)$ PEBBLE with actual human preferences (PEBBLE-Dense)). This comparison aims to aid the practitioners for choosing between hand-designing dense reward functions, or implementing PIPER. Although SAC-Dense outperforms PIPER in easier tasks, PIPER is able to outperform SAC-Dense in harder environments. Notably, SAC-Dense completely fails to solve the kitchen task. PEBBLE-Dense is unable to show good performance in any task. We provide further discussion in Appendix~\ref{sec:sec:appendix_dense_rewards}.

\section{Discussion}
\label{sec:discussion}
\textbf{Limitations}: PIPER uses L2 distance between states as the informative metric, which might be hard to compute in scenarios where the subgoals and goals are high-dimensional (e.g. images). A possible way to deal with this is to find a near optimal latent representation~\cite{nachum2018near}. Further, although PIPER replaces human-in-the-loop preferences via primitive-in-the-loop (PiL) approach to generate trajectory preferences, human preferences may contain additional information that PiL may be unable to capture (say, humans may prefer trajectories that are \textit{safe} to traverse, which do not involve any wall collisions). A possible way to deal with this issue is to employ additional rewards in Eqn~\ref{eqn:reward_s}, which prefer a safe trajectory. Similarly, the proposed primitive regularization approach may not generate such \textit{safe} subgoals in its naive form. To this end, we can additionally learn a new \textit{safe} value function $V^s(s,g)$, which regularizes the higher agents to produce \textit{safe} subgoals. Finally, the proposed target networks require additional memory and training overhead, and require setting the hyper-parameter $\tau$. However, we found in practice that the memory and training overheads are minimal, and setting $\tau$ does not cause a significant overhead. Although not explored in this work, we would like to overcome these limitations in the future.

\textbf{Conclusion and future work}: In this work, we propose PIPER, a primitive informed hierarchical RL algorithm that leverages preference based learning to mitigate the non-stationarity issue in HRL. Using primitive informed regularization, PIPER is able to generate efficient subgoals, according to the goal achieving capability of the lower primitive. We demonstrate that by incorporating primitive informed regularization, hindsight relabeling, and target networks, PIPER is able to perform complex robotic tasks in sparse scenarios, and significantly outperform the baselines. Additionally, PIPER is able to outperform flat preference based learning. We therefore believe that hierarchical preference based learning is a promising step towards building practical robotic systems that solve complex real-world tasks.

\section*{Software and Data}

The implementation code and data is provided \href{https://github.com/Utsavz/piper.git}{here}.



\section*{Impact Statement}

Regarding the impact of our proposed approach and algorithm, we don't see any immediate impact in terms of technological advancements. Instead, our contributions are principally conceptual, aiming to address foundational aspects of Hierarchical Reinforcement Learning (HRL). By advocating for a preference-based methodology, we propose a novel framework that we believe holds significant potential for enriching HRL research and its associated domains. This conceptual groundwork lays the foundation for future explorations and could catalyze advancements in HRL and related areas.


\bibliography{example_paper}

\begin{thebibliography}{42}
\providecommand{\natexlab}[1]{#1}
\providecommand{\url}[1]{\texttt{#1}}
\expandafter\ifx\csname urlstyle\endcsname\relax
  \providecommand{\doi}[1]{doi: #1}\else
  \providecommand{\doi}{doi: \begingroup \urlstyle{rm}\Url}\fi

\bibitem[Andrychowicz et~al.(2017)Andrychowicz, Wolski, Ray, Schneider, Fong, Welinder, McGrew, Tobin, Abbeel, and Zaremba]{andrychowicz2017hindsight}
Andrychowicz, M., Wolski, F., Ray, A., Schneider, J., Fong, R., Welinder, P., McGrew, B., Tobin, J., Abbeel, P., and Zaremba, W.
\newblock Hindsight experience replay.
\newblock \emph{CoRR}, abs/1707.01495, 2017.
\newblock URL \url{http://arxiv.org/abs/1707.01495}.

\bibitem[Barto \& Mahadevan(2003)Barto and Mahadevan]{Barto03recentadvances}
Barto, A.~G. and Mahadevan, S.
\newblock Recent advances in hierarchical reinforcement learning.
\newblock \emph{Discrete Event Dynamic Systems}, 13:\penalty0 341--379, 2003.

\bibitem[Bradley \& Terry(1952)Bradley and Terry]{bradley_terry}
Bradley, R.~A. and Terry, M.~E.
\newblock Rank analysis of incomplete block designs: I. the method of paired comparisons.
\newblock \emph{Biometrika}, 39:\penalty0 324, 1952.
\newblock URL \url{https://api.semanticscholar.org/CorpusID:125209808}.

\bibitem[Cao et~al.(2020)Cao, Wong, and Lin]{cao2020human}
Cao, Z., Wong, K., and Lin, C.-T.
\newblock Human preference scaling with demonstrations for deep reinforcement learning.
\newblock \emph{arXiv preprint arXiv:2007.12904}, 2020.

\bibitem[Chane-Sane et~al.(2021)Chane-Sane, Schmid, and Laptev]{chane2021goal}
Chane-Sane, E., Schmid, C., and Laptev, I.
\newblock Goal-conditioned reinforcement learning with imagined subgoals.
\newblock In \emph{International Conference on Machine Learning}, pp.\  1430--1440. PMLR, 2021.

\bibitem[Christiano et~al.(2017)Christiano, Leike, Brown, Martic, Legg, and Amodei]{christiano2017deep}
Christiano, P.~F., Leike, J., Brown, T., Martic, M., Legg, S., and Amodei, D.
\newblock Deep reinforcement learning from human preferences.
\newblock \emph{Advances in neural information processing systems}, 30, 2017.

\bibitem[Dalal et~al.(2021)Dalal, Pathak, and Salakhutdinov]{dalal2021accelerating}
Dalal, M., Pathak, D., and Salakhutdinov, R.~R.
\newblock Accelerating robotic reinforcement learning via parameterized action primitives.
\newblock \emph{Advances in Neural Information Processing Systems}, 34:\penalty0 21847--21859, 2021.

\bibitem[Daniel et~al.(2015)Daniel, Kroemer, Viering, Metz, and Peters]{daniel2015active}
Daniel, C., Kroemer, O., Viering, M., Metz, J., and Peters, J.
\newblock Active reward learning with a novel acquisition function.
\newblock \emph{Autonomous Robots}, 39:\penalty0 389--405, 2015.

\bibitem[Dayan \& Hinton(1992)Dayan and Hinton]{dayan1992feudal}
Dayan, P. and Hinton, G.~E.
\newblock Feudal reinforcement learning.
\newblock \emph{Advances in neural information processing systems}, 5, 1992.

\bibitem[Dietterich(1999)]{DBLP:journals/corr/cs-LG-9905014}
Dietterich, T.~G.
\newblock Hierarchical reinforcement learning with the {MAXQ} value function decomposition.
\newblock \emph{CoRR}, cs.LG/9905014, 1999.
\newblock URL \url{https://arxiv.org/abs/cs/9905014}.

\bibitem[Gu et~al.(2016)Gu, Holly, Lillicrap, and Levine]{DBLP:journals/corr/GuHLL16}
Gu, S., Holly, E., Lillicrap, T.~P., and Levine, S.
\newblock Deep reinforcement learning for robotic manipulation.
\newblock \emph{CoRR}, abs/1610.00633, 2016.
\newblock URL \url{http://arxiv.org/abs/1610.00633}.

\bibitem[Gupta et~al.(2019)Gupta, Kumar, Lynch, Levine, and Hausman]{gupta2019relay}
Gupta, A., Kumar, V., Lynch, C., Levine, S., and Hausman, K.
\newblock Relay policy learning: Solving long-horizon tasks via imitation and reinforcement learning.
\newblock \emph{arXiv preprint arXiv:1910.11956}, 2019.

\bibitem[Haarnoja et~al.(2018)Haarnoja, Zhou, Abbeel, and Levine]{DBLP:journals/corr/abs-1801-01290}
Haarnoja, T., Zhou, A., Abbeel, P., and Levine, S.
\newblock Soft actor-critic: Off-policy maximum entropy deep reinforcement learning with a stochastic actor.
\newblock \emph{CoRR}, abs/1801.01290, 2018.
\newblock URL \url{http://arxiv.org/abs/1801.01290}.

\bibitem[Harb et~al.(2018)Harb, Bacon, Klissarov, and Precup]{harb2018waiting}
Harb, J., Bacon, P.-L., Klissarov, M., and Precup, D.
\newblock When waiting is not an option: Learning options with a deliberation cost.
\newblock In \emph{Proceedings of the AAAI Conference on Artificial Intelligence}, volume~32, 2018.

\bibitem[Ibarz et~al.(2018)Ibarz, Leike, Pohlen, Irving, Legg, and Amodei]{DBLP:journals/corr/abs-1811-06521}
Ibarz, B., Leike, J., Pohlen, T., Irving, G., Legg, S., and Amodei, D.
\newblock Reward learning from human preferences and demonstrations in atari, 2018.

\bibitem[Kalashnikov et~al.(2018)Kalashnikov, Irpan, Pastor, Ibarz, Herzog, Jang, Quillen, Holly, Kalakrishnan, Vanhoucke, and Levine]{DBLP:journals/corr/abs-1806-10293}
Kalashnikov, D., Irpan, A., Pastor, P., Ibarz, J., Herzog, A., Jang, E., Quillen, D., Holly, E., Kalakrishnan, M., Vanhoucke, V., and Levine, S.
\newblock Qt-opt: Scalable deep reinforcement learning for vision-based robotic manipulation.
\newblock \emph{CoRR}, abs/1806.10293, 2018.
\newblock URL \url{http://arxiv.org/abs/1806.10293}.

\bibitem[Kingma \& Ba(2014)Kingma and Ba]{kingma2014adam}
Kingma, D.~P. and Ba, J.
\newblock Adam: A method for stochastic optimization.
\newblock \emph{arXiv preprint arXiv:1412.6980}, 2014.

\bibitem[Klissarov et~al.(2017)Klissarov, Bacon, Harb, and Precup]{klissarov2017learnings}
Klissarov, M., Bacon, P.-L., Harb, J., and Precup, D.
\newblock Learnings options end-to-end for continuous action tasks.
\newblock \emph{arXiv preprint arXiv:1712.00004}, 2017.

\bibitem[Knox \& Stone(2009)Knox and Stone]{knox2009interactively}
Knox, W.~B. and Stone, P.
\newblock Interactively shaping agents via human reinforcement: The tamer framework.
\newblock In \emph{Proceedings of the fifth international conference on Knowledge capture}, pp.\  9--16, 2009.

\bibitem[Kostrikov et~al.(2018)Kostrikov, Agrawal, Dwibedi, Levine, and Tompson]{kostrikov2018discriminator}
Kostrikov, I., Agrawal, K.~K., Dwibedi, D., Levine, S., and Tompson, J.
\newblock Discriminator-actor-critic: Addressing sample inefficiency and reward bias in adversarial imitation learning.
\newblock \emph{arXiv preprint arXiv:1809.02925}, 2018.

\bibitem[Lee et~al.(2021)Lee, Smith, and Abbeel]{DBLP:journals/corr/abs-2106-05091}
Lee, K., Smith, L., and Abbeel, P.
\newblock Pebble: Feedback-efficient interactive reinforcement learning via relabeling experience and unsupervised pre-training, 2021.

\bibitem[Levine(2018)]{levine2018reinforcement}
Levine, S.
\newblock Reinforcement learning and control as probabilistic inference: Tutorial and review.
\newblock \emph{arXiv preprint arXiv:1805.00909}, 2018.

\bibitem[Levine et~al.(2015)Levine, Finn, Darrell, and Abbeel]{DBLP:journals/corr/LevineFDA15}
Levine, S., Finn, C., Darrell, T., and Abbeel, P.
\newblock End-to-end training of deep visuomotor policies.
\newblock \emph{CoRR}, abs/1504.00702, 2015.
\newblock URL \url{http://arxiv.org/abs/1504.00702}.

\bibitem[Levy et~al.(2018)Levy, Konidaris, Platt, and Saenko]{levy2018learning}
Levy, A., Konidaris, G., Platt, R., and Saenko, K.
\newblock Learning multi-level hierarchies with hindsight.
\newblock In \emph{International Conference on Learning Representations}, 2018.

\bibitem[Lillicrap et~al.(2015)Lillicrap, Hunt, Pritzel, Heess, Erez, Tassa, Silver, and Wierstra]{lillicrap2015continuous}
Lillicrap, T.~P., Hunt, J.~J., Pritzel, A., Heess, N., Erez, T., Tassa, Y., Silver, D., and Wierstra, D.
\newblock Continuous control with deep reinforcement learning.
\newblock \emph{arXiv preprint arXiv:1509.02971}, 2015.

\bibitem[Mnih et~al.(2013)Mnih, Kavukcuoglu, Silver, Graves, Antonoglou, Wierstra, and Riedmiller]{DBLP:journals/corr/MnihKSGAWR13}
Mnih, V., Kavukcuoglu, K., Silver, D., Graves, A., Antonoglou, I., Wierstra, D., and Riedmiller, M.~A.
\newblock Playing atari with deep reinforcement learning.
\newblock \emph{CoRR}, abs/1312.5602, 2013.
\newblock URL \url{http://arxiv.org/abs/1312.5602}.

\bibitem[Nachum et~al.(2018{\natexlab{a}})Nachum, Gu, Lee, and Levine]{nachum2018near}
Nachum, O., Gu, S., Lee, H., and Levine, S.
\newblock Near-optimal representation learning for hierarchical reinforcement learning.
\newblock \emph{arXiv preprint arXiv:1810.01257}, 2018{\natexlab{a}}.

\bibitem[Nachum et~al.(2018{\natexlab{b}})Nachum, Gu, Lee, and Levine]{nachum2018data}
Nachum, O., Gu, S.~S., Lee, H., and Levine, S.
\newblock Data-efficient hierarchical reinforcement learning.
\newblock \emph{Advances in neural information processing systems}, 31, 2018{\natexlab{b}}.

\bibitem[Nachum et~al.(2019)Nachum, Tang, Lu, Gu, Lee, and Levine]{nachum2019does}
Nachum, O., Tang, H., Lu, X., Gu, S., Lee, H., and Levine, S.
\newblock Why does hierarchy (sometimes) work so well in reinforcement learning?
\newblock \emph{arXiv preprint arXiv:1909.10618}, 2019.

\bibitem[Nair et~al.(2018)Nair, McGrew, Andrychowicz, Zaremba, and Abbeel]{nair2018overcoming}
Nair, A., McGrew, B., Andrychowicz, M., Zaremba, W., and Abbeel, P.
\newblock Overcoming exploration in reinforcement learning with demonstrations.
\newblock In \emph{2018 IEEE international conference on robotics and automation (ICRA)}, pp.\  6292--6299. IEEE, 2018.

\bibitem[Nasiriany et~al.(2021)Nasiriany, Liu, and Zhu]{DBLP:journals/corr/abs-2110-03655}
Nasiriany, S., Liu, H., and Zhu, Y.
\newblock Augmenting reinforcement learning with behavior primitives for diverse manipulation tasks.
\newblock \emph{CoRR}, abs/2110.03655, 2021.
\newblock URL \url{https://arxiv.org/abs/2110.03655}.

\bibitem[Ng et~al.(1999)Ng, Harada, and Russell]{ng1999policy}
Ng, A.~Y., Harada, D., and Russell, S.
\newblock Policy invariance under reward transformations: Theory and application to reward shaping.
\newblock In \emph{Icml}, volume~99, pp.\  278--287. Citeseer, 1999.

\bibitem[Parr \& Russell(1998)Parr and Russell]{NIPS1997_5ca3e9b1}
Parr, R. and Russell, S.
\newblock Reinforcement learning with hierarchies of machines.
\newblock In Jordan, M., Kearns, M., and Solla, S. (eds.), \emph{Advances in Neural Information Processing Systems}, volume~10. MIT Press, 1998.

\bibitem[Pilarski et~al.(2011)Pilarski, Dawson, Degris, Fahimi, Carey, and Sutton]{pilarski2011online}
Pilarski, P.~M., Dawson, M.~R., Degris, T., Fahimi, F., Carey, J.~P., and Sutton, R.~S.
\newblock Online human training of a myoelectric prosthesis controller via actor-critic reinforcement learning.
\newblock In \emph{2011 IEEE international conference on rehabilitation robotics}, pp.\  1--7. IEEE, 2011.

\bibitem[Rajeswaran et~al.(2017)Rajeswaran, Kumar, Gupta, Schulman, Todorov, and Levine]{DBLP:journals/corr/abs-1709-10087}
Rajeswaran, A., Kumar, V., Gupta, A., Schulman, J., Todorov, E., and Levine, S.
\newblock Learning complex dexterous manipulation with deep reinforcement learning and demonstrations.
\newblock \emph{CoRR}, abs/1709.10087, 2017.
\newblock URL \url{http://arxiv.org/abs/1709.10087}.

\bibitem[Silver et~al.(2016)Silver, Huang, Maddison, Guez, Sifre, Van Den~Driessche, Schrittwieser, Antonoglou, Panneershelvam, Lanctot, et~al.]{silver2016mastering}
Silver, D., Huang, A., Maddison, C.~J., Guez, A., Sifre, L., Van Den~Driessche, G., Schrittwieser, J., Antonoglou, I., Panneershelvam, V., Lanctot, M., et~al.
\newblock Mastering the game of go with deep neural networks and tree search.
\newblock \emph{nature}, 529\penalty0 (7587):\penalty0 484--489, 2016.

\bibitem[Sutton et~al.(1999)Sutton, Precup, and Singh]{sutton1999between}
Sutton, R.~S., Precup, D., and Singh, S.
\newblock Between mdps and semi-mdps: A framework for temporal abstraction in reinforcement learning.
\newblock \emph{Artificial intelligence}, 112\penalty0 (1-2):\penalty0 181--211, 1999.

\bibitem[Vezhnevets et~al.(2017)Vezhnevets, Osindero, Schaul, Heess, Jaderberg, Silver, and Kavukcuoglu]{vezhnevets2017feudal}
Vezhnevets, A.~S., Osindero, S., Schaul, T., Heess, N., Jaderberg, M., Silver, D., and Kavukcuoglu, K.
\newblock Feudal networks for hierarchical reinforcement learning.
\newblock In \emph{International Conference on Machine Learning}, pp.\  3540--3549. PMLR, 2017.

\bibitem[Warnell et~al.(2018)Warnell, Waytowich, Lawhern, and Stone]{warnell2018deep}
Warnell, G., Waytowich, N., Lawhern, V., and Stone, P.
\newblock Deep tamer: Interactive agent shaping in high-dimensional state spaces.
\newblock In \emph{Proceedings of the AAAI conference on artificial intelligence}, volume~32, 2018.

\bibitem[Wilson et~al.(2012{\natexlab{a}})Wilson, Fern, and Tadepalli]{NIPS2012_16c222aa}
Wilson, A., Fern, A., and Tadepalli, P.
\newblock A bayesian approach for policy learning from trajectory preference queries.
\newblock In Pereira, F., Burges, C., Bottou, L., and Weinberger, K. (eds.), \emph{Advances in Neural Information Processing Systems}, volume~25. Curran Associates, Inc., 2012{\natexlab{a}}.
\newblock URL \url{https://proceedings.neurips.cc/paper_files/paper/2012/file/16c222aa19898e5058938167c8ab6c57-Paper.pdf}.

\bibitem[Wilson et~al.(2012{\natexlab{b}})Wilson, Fern, and Tadepalli]{wilson2012bayesian}
Wilson, A., Fern, A., and Tadepalli, P.
\newblock A bayesian approach for policy learning from trajectory preference queries.
\newblock \emph{Advances in neural information processing systems}, 25, 2012{\natexlab{b}}.

\bibitem[Ziebart et~al.(2008)Ziebart, Maas, Bagnell, Dey, et~al.]{ziebart2008maximum}
Ziebart, B.~D., Maas, A.~L., Bagnell, J.~A., Dey, A.~K., et~al.
\newblock Maximum entropy inverse reinforcement learning.
\newblock In \emph{Aaai}, volume~8, pp.\  1433--1438. Chicago, IL, USA, 2008.

\end{thebibliography}
\bibliographystyle{icml2024}

\newpage
\appendix
\onecolumn
\section{Appendix}
\label{sec:appendix}


\subsection{Deriving the final optimum of KL-Constrained Reward Maximization Objective}
\label{appendix:pi_star_derivation}

In this appendix, we will derive Eqn~\ref{eqn:eqn_dpo_optimal_policy} from Eqn~\ref{eqn:dpo_input}. Thus, we optimize the following objective:
\begin{equation}
\label{eqn:dpo_input_1}
    \max_{\pi_U}  \mathbb{E}_{\pi_U}[\sum_{t=0}^{T}(r^s(s, g^*, g)-\beta \mathbb{D}_{\mathrm{KL}}[\pi^{H}(g_t|s_t) \| \pi_{reg}(g_t|s_t)])].
\end{equation}

Re-writing the above equation after expanding KL divergence:

\begin{equation}
\label{eqn:dpo_input_2}
    = \max_{\pi_U}  \mathbb{E}_{\pi_U}[\sum_{t=0}^{T}(r^s(s, g^*, g)-\beta \log \frac{\pi^{H}(g_t|s_t)}{\pi_{reg}(g_t|s_t)})]
\end{equation}

\begin{equation}
\label{eqn:dpo_input_3}
    = \max_{\pi_U}  \mathbb{E}_{\pi_U}[\sum_{t=0}^{T}(r^s(s, g^*, g)-\beta \log \pi^{H}(g_t|s_t) + \beta \log \pi_{reg}(g_t|s_t))].
\end{equation}

Here, we substitute $\pi_{reg}$ from Eqn~\ref{eqn:eqn_reg}, and $m=\frac{\alpha}{\beta}$ in Equation \ref{eqn:dpo_input_3},

\begin{equation}
\label{eqn:dpo_input_5}
\begin{split}
     = \max_{\pi_U}  \mathbb{E}_{\pi_U}[\sum_{t=0}^{T}(r^s(s, g^*, g)-\beta \log \pi^{H}(g_t|s_t) + 
     \beta \log \exp (m(V_{\pi^{L}}(s_t,g_t)))  - \beta \log \sum_{g_t}^{} \exp (m(V_{\pi^{L}}(s_t,g_t))))]   
\end{split}
\end{equation}

\begin{equation}
\label{eqn:dpo_input_6}
\begin{split}
     = \max_{\pi_U}  \mathbb{E}_{\pi_U}[\sum_{t=0}^{T}(r^s(s, g^*, g)-\beta \log \pi^{H}(g_t|s_t) + \alpha (V_{\pi^{L}}(s_t,g_t)) 
      - \beta \log \sum_{g_t}^{} \exp (m(V_{\pi^{L}}(s_t,g_t))))]    
\end{split}
\end{equation}

\begin{equation}
\label{eqn:dpo_input_7}
\begin{split}
     = \min_{\pi_U}  \mathbb{E}_{\pi_U}[\sum_{t=0}^{T}(\log \pi^{H}(g_t|s_t) - \frac{1}{\beta} (r^s(s, g^*, g) + \alpha (V_{\pi^{L}}(s_t,g_t)))
      + \log \sum_{g_t}^{} \exp (m(V_{\pi^{L}}(s_t,g_t))))]    
\end{split}
\end{equation}

\begin{equation}
\label{eqn:dpo_input_8}
\begin{split}
    = \min_{\pi_U}  \mathbb{E}_{\pi_U}[\sum_{t=0}^{T}(\log (\frac{\pi^{H}(g_t|s_t)}{\exp(\frac{1}{\beta} (r^s(s, g^*, g) + \alpha (V_{\pi^{L}}(s_t,g_t))))})
      + \log \sum_{g_t}^{} \exp (m(V_{\pi^{L}}(s_t,g_t))))]
\end{split}
\end{equation}

\begin{equation}
\label{eqn:dpo_input_10}
\begin{split}
     = \min_{\pi_U}  \mathbb{E}_{\pi_U}[\sum_{t=0}^{T}(\log (\frac{\pi^{H}(g_t|s_t)}{\frac{1}{Z(s_t)}\exp(\frac{1}{\beta} (r^s(s, g^*, g) + \alpha (V_{\pi^{L}}(s_t,g_t))))})
      + \log \sum_{g_t}^{} \exp (m(V_{\pi^{L}}(s_t,g_t))) - \log Z(s_t))],
\end{split}
\end{equation}

where, $Z(s_t) = \sum_{g_t} \exp (\frac{1}{\beta} (r^s(s, g^*, g) + \alpha (V_{\pi^{L}}(s_t,g_t))))$ is the partition function.

Note that the partition function $Z(s_t)$ and the term $\log \sum_{g_t}^{} \exp (m(V_{\pi^{L}}(s_t,g_t)))$ do not depend on policy $\pi_U$. We define

\begin{equation}
\label{eqn:dpo_input_111}
\begin{split}
    \pi^{H^{*}}(g_t|s_t) = \frac{1}{Z(s_t)} \exp (\frac{1}{\beta} (r^s(s, g^*, g) + \alpha (V_{\pi^L}(s_t,g_t) ))),
\end{split}    
\end{equation}

which is a valid probability distribution, as $\pi^{H^{*}}(g_t|s_t) \geq 0$ and $\sum_{g_t}^{} \pi^{H^{*}}(g_t|s_t)=1$. We can re-organize the equation as

\begin{equation}
\label{eqn:dpo_input_11}
\begin{split}
     = \min_{\pi_U}  \mathbb{E}_{\pi_U}[\sum_{t=0}^{T}(\mathbb{D}_{\mathrm{KL}}[\pi^{H}(g_t|s_t) \| \pi^{H^{*}}(g_t|s_t)]
      - \log \sum_{g_t}^{} \exp (m(V_{\pi^{L}}(s_t,g_t))) - \log Z(s_t))],
\end{split}    
\end{equation}

where $\pi^{H^{*}}(g_t|s_t)$ is minimized when the KL-divergence is $0$, according to Gibbs' inequality. Hence, we get the optimal solution

\begin{equation}
\label{eqn:dpo_input_12}
\begin{split}
    \pi^{H}(g_t|s_t) = \pi^{H^{*}}(g_t|s_t) = \frac{1}{Z(s_t)} \exp (\frac{1}{\beta} (r^s(s, g^*, g) + \alpha (V_{\pi^L}(s_t,g_t) ))).
\end{split}
\end{equation}

\subsection{Additional Implementation details}
\label{sec:implementation_details}
In addition, we compare our approach with Discriminator Actor-Critic \citep{kostrikov2018discriminator}, which is provided a single expert demonstration. Although not explored in this work, combining preference-based learning and learning from demonstrations is an interesting research avenue~\citep{cao2020human}.

\subsubsection{Additional hyper-parameters}
Here, we enlist the additional hyper-parameters used in PIPER:

\texttt{\textbf{activation}: tanh [activation for reward model] \\
\textbf{layers}: 3 [number of layers in the critic/actor networks] \\
\textbf{hidden}: 512 [number of neurons in each hidden layers] \\
\textbf{Q\_lr}: 0.001 [critic learning rate] \\
\textbf{pi\_lr}: 0.001 [actor learning rate] \\
\textbf{buffer\_size}: int(1E7) [for experience replay] \\
\textbf{tau}: 0.8 [polyak averaging coefficient] \\
\textbf{clip\_obs}: 200 [clip observation] \\
\textbf{n\_cycles}: 1 [per epoch] \\
\textbf{n\_batches}: 10 [training batches per cycle] \\
\textbf{batch\_size}: 1024 [batch size hyper-parameter] \\
\textbf{reward\_batch\_size}: 50 [reward batch size for PEBBLE and RFLAT] \\
\textbf{random\_eps}: 0.2 [percentage of time a random action is taken] \\
\textbf{alpha}: 0.05 [weightage parameter for SAC] \\
\textbf{noise\_eps}: 0.05 [std of gaussian noise added to not-completely-random actions] \\
\textbf{norm\_eps}: 0.01 [epsilon used for observation normalization] \\
\textbf{norm\_clip}: 5 [normalized observations are cropped to this values] \\
\textbf{adam\_beta1}: 0.9 [beta 1 for Adam optimizer] \\
\textbf{adam\_beta2}: 0.999 [beta 2 for Adam optimizer] \\
}

\subsection{Environment details}
\label{sec:appendix_env_details}
\subsubsection{Maze navigation task}
In this environment, a $7$-DOF robotic arm gripper navigates across random four room mazes. The gripper arm is kept closed and the positions of walls and gates are randomly generated. The table is discretized into a rectangular $W*H$ grid, and the vertical and horizontal wall positions $W_{P}$ and $H_{P}$ are randomly picked from $(1,W-2)$ and $(1,H-2)$ respectively. In the four room environment thus constructed, the four gate positions are randomly picked from $(1,W_{P}-1)$, $(W_{P}+1,W-2)$, $(1,H_{P}-1)$ and $(H_{P}+1,H-2)$. The height of gripper is kept fixed at table height, and it has to navigate across the maze to the goal position(shown as red sphere). 
\par The following implementation details refer to both the higher and lower level polices, unless otherwise explicitly stated. The state and action spaces in the environment are continuous. The state is represented as the vector $[p,\mathcal{M}]$, where $p$ is current gripper position and $\mathcal{M}$ is the sparse maze array. The higher level policy input is thus a concatenated vector $[p,\mathcal{M},g]$, where $g$ is the target goal position, whereas the lower level policy input is concatenated vector $[p,\mathcal{M},s_g]$, where $s_g$ is the sub-goal provided by the higher level policy. The current position of the gripper is the current achieved goal. The sparse maze array $\mathcal{M}$ is a discrete $2D$ one-hot vector array, where $1$ represents presence of a wall block, and $0$ absence. In our experiments, the size of $p$ and $\mathcal{M}$ are kept to be $3$ and $110$ respectively. The upper level predicts subgoal $s_g$, hence the higher level policy action space dimension is the same as the dimension of goal space of lower primitive. The lower primitive action $a$ which is directly executed on the environment, is a $4$ dimensional vector with every dimension $a_i \in [0,1]$. The first $3$ dimensions provide offsets to be scaled and added to gripper position for moving it to the intended position. The last dimension provides gripper control($0$ implies a fully closed gripper, $0.5$ implies a half closed gripper and $1$ implies a fully open gripper).

\subsubsection{Pick and place, Push and Hollow Environments}
In the pick and place environment, a $7$-DOF robotic arm gripper has to pick a square block and bring/place it to a goal position. We set the goal position slightly higher than table height. In this complex task, the gripper has to navigate to the block, close the gripper to hold the block, and then bring the block to the desired goal position. In the push environment, the $7$-DOF robotic arm gripper has to push a square block towards the goal position. In hollow task, the $7$-DOF robotic arm gripper has to pick a square hollow block and place it such that a fixed vertical structure on the table goes through the hollow block. The state is represented as the vector $[p,o,q,e]$, where $p$ is current gripper position, $o$ is the position of the block object placed on the table, $q$ is the relative position of the block with respect to the gripper, and $e$ consists of linear and angular velocities of the gripper and the block object. The higher level policy input is thus a concatenated vector $[p,o,q,e,g]$, where $g$ is the target goal position. The lower level policy input is concatenated vector $[p,o,q,e,s_g]$, where $s_g$ is the sub-goal provided by the higher level policy. The current position of the block object is the current achieved goal. In our experiments, the sizes of $p$, $o$, $q$, $e$ are kept to be $3$, $3$, $3$ and $11$ respectively. The upper level predicts subgoal $s_g$, hence the higher level policy action space and goal space have the same dimension. The lower primitive action $a$ is a $4$ dimensional vector with every dimension $a_i \in [0,1]$. The first $3$ dimensions provide gripper position offsets, and the last dimension provides gripper control ($0$ means closed gripper and $1$ means open gripper). While training, the position of block object and goal are randomly generated (block is always initialized on the table, and goal is always above the table at a fixed height). 

\subsection{Additional experiments}
In this section, we provide additional experiments, comparing PIPER with additional baselines.
\subsubsection{Experiments with dense rewards baselines}
\label{sec:sec:appendix_dense_rewards}

We compare the performance of these baselines with PIPER in Figure~\ref{fig:ablation_dense_reward}. We find that single level SAC implementation with explicitly hand-designed and fine-tuned dense rewards (which we call SAC-Dense baseline) outperforms PIPER in easier tasks like Maze navigation and Push environments. However, we find that as the task complexity increases, the performance of SAC-Dense degrades and PIPER is able to outperform SAC-Dense in Pick and Place and kitchen environments. In the kitchen environment, SAC-Dense fails to solve the task. We believe this to is due to two reasons:
\begin{itemize}
    \item hand-designing a suitable reward in complex environments is comparatively hard, and may lead to sub-optimal performance, and
    \item the advantages of various design choices in PIPER (exploration and task abstraction due to hierarchical structure, mitigating non-stationarity using preference based RL, reward densification using hindsight relabeling, feasible subgoal generation using primitive regularization, and added training stability due to target networks) out-weigh the presence of dense rewards in SAC-Dense baseline. 
\end{itemize}
We also implemented PEBBLE with actual human preferences (rather than primitives), but find that this baseline is unable to show good performance in any of the tasks. This shows that although PEBBLE shows impressive performance in simple environments, it needs further improvements to make it work in complex robotic manipulation tasks like the ones explored in this work.

\subsection{Environment visualizations}
\label{sec:viz}

Here, we provide some visualizations of the agent successfully performing the task.

\begin{figure}[H]
\vspace{5pt}
\centering

\includegraphics[scale=0.11]{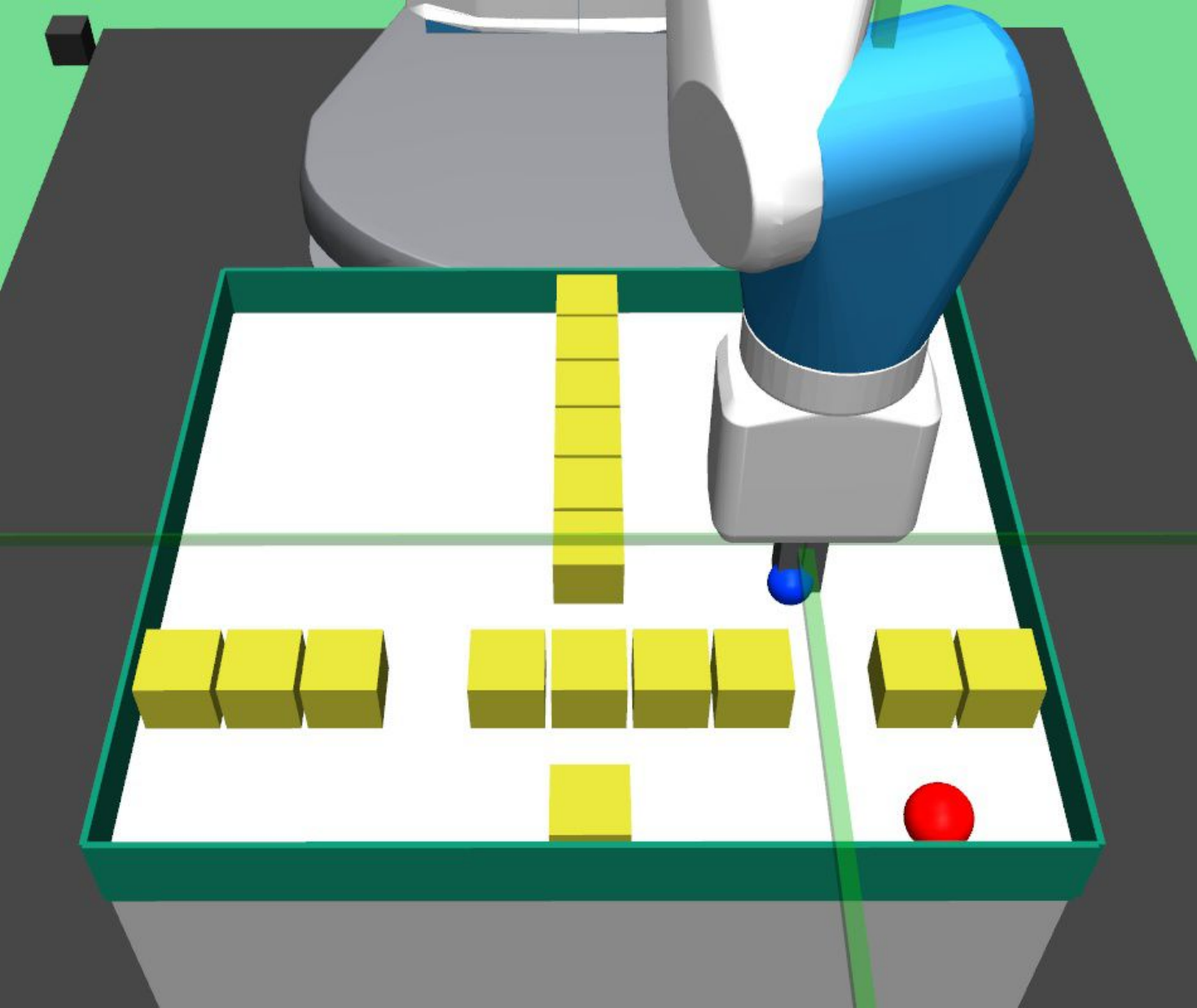}
\includegraphics[scale=0.11]{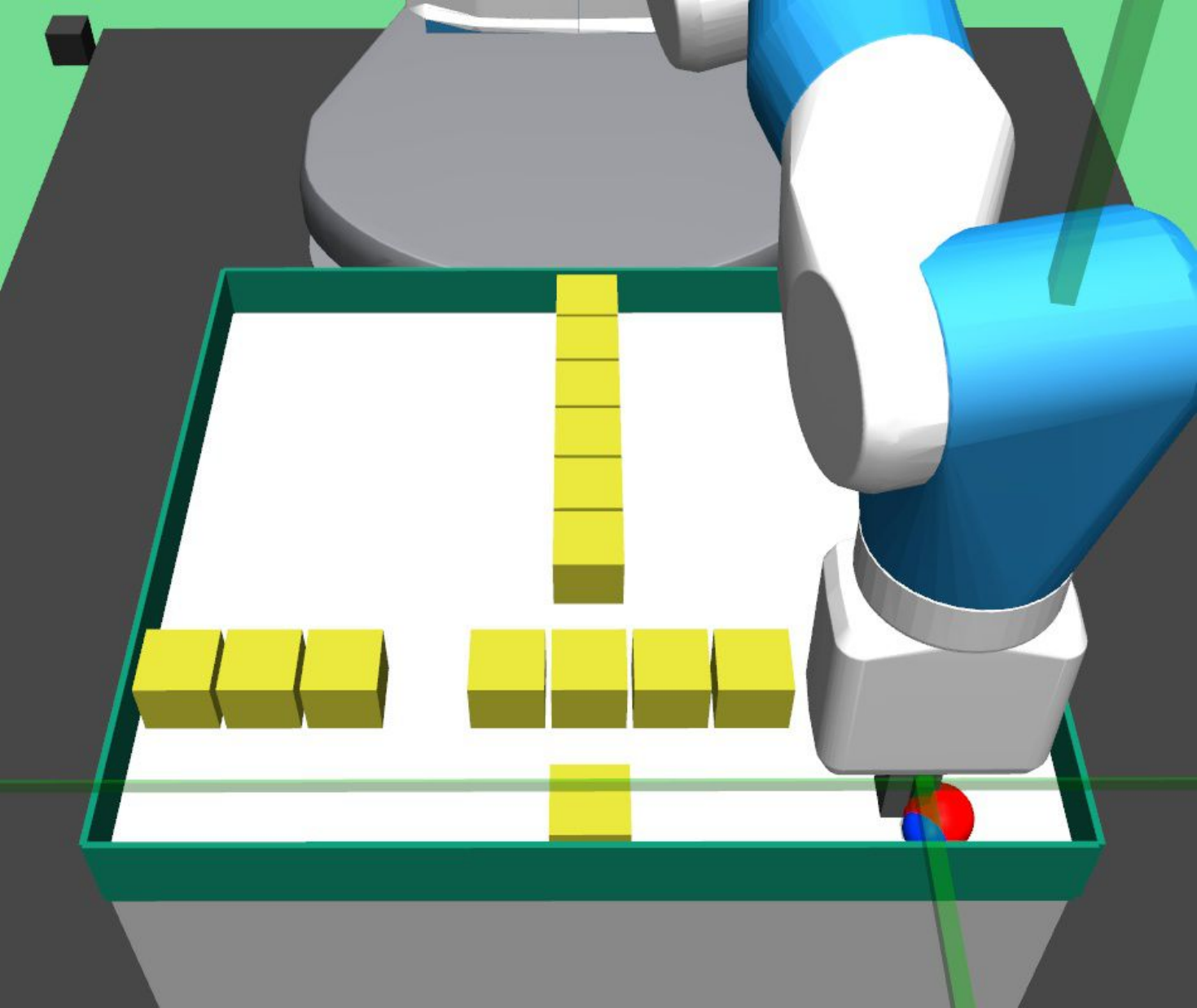}
\includegraphics[scale=0.11]{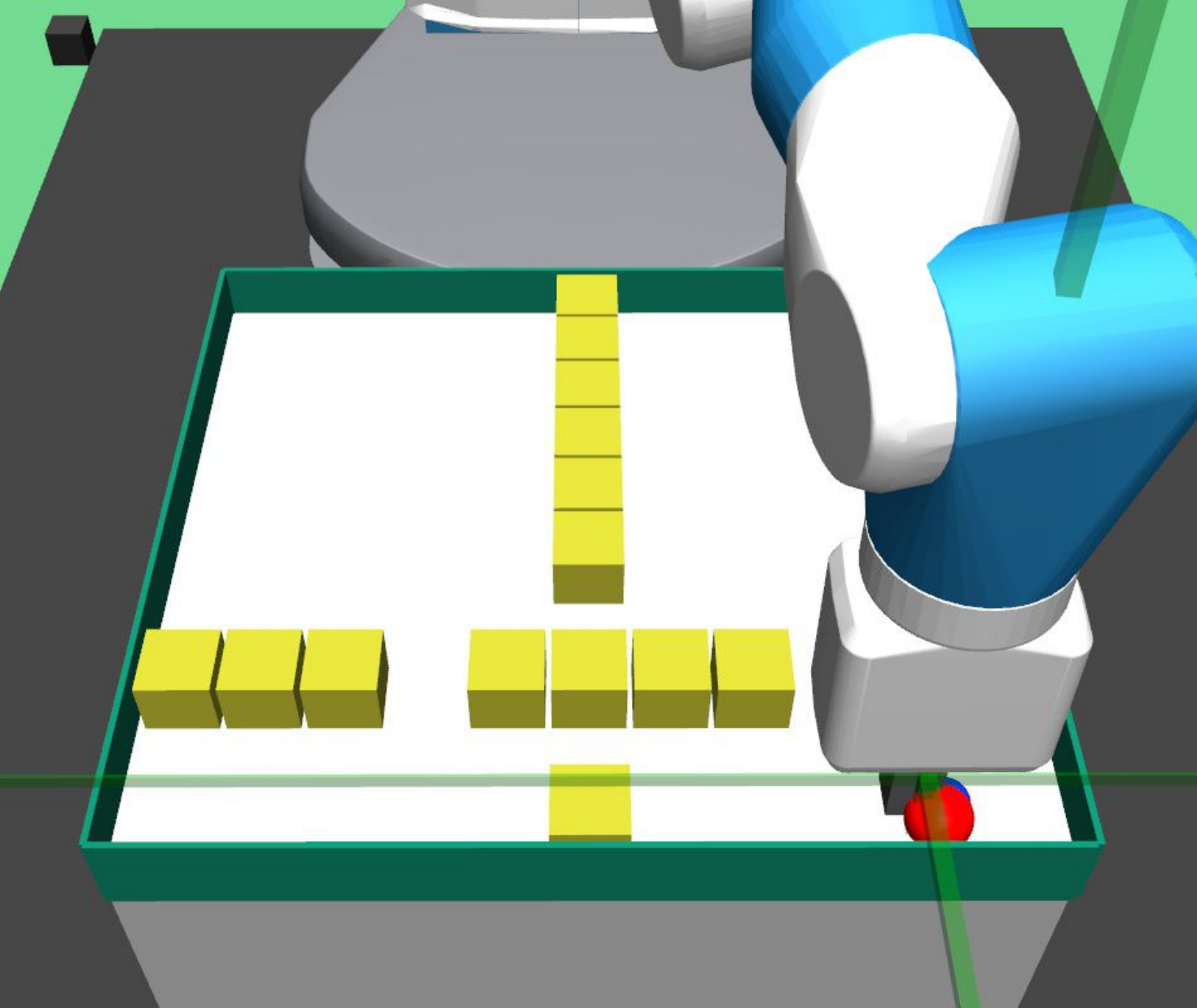}
\includegraphics[scale=0.11]{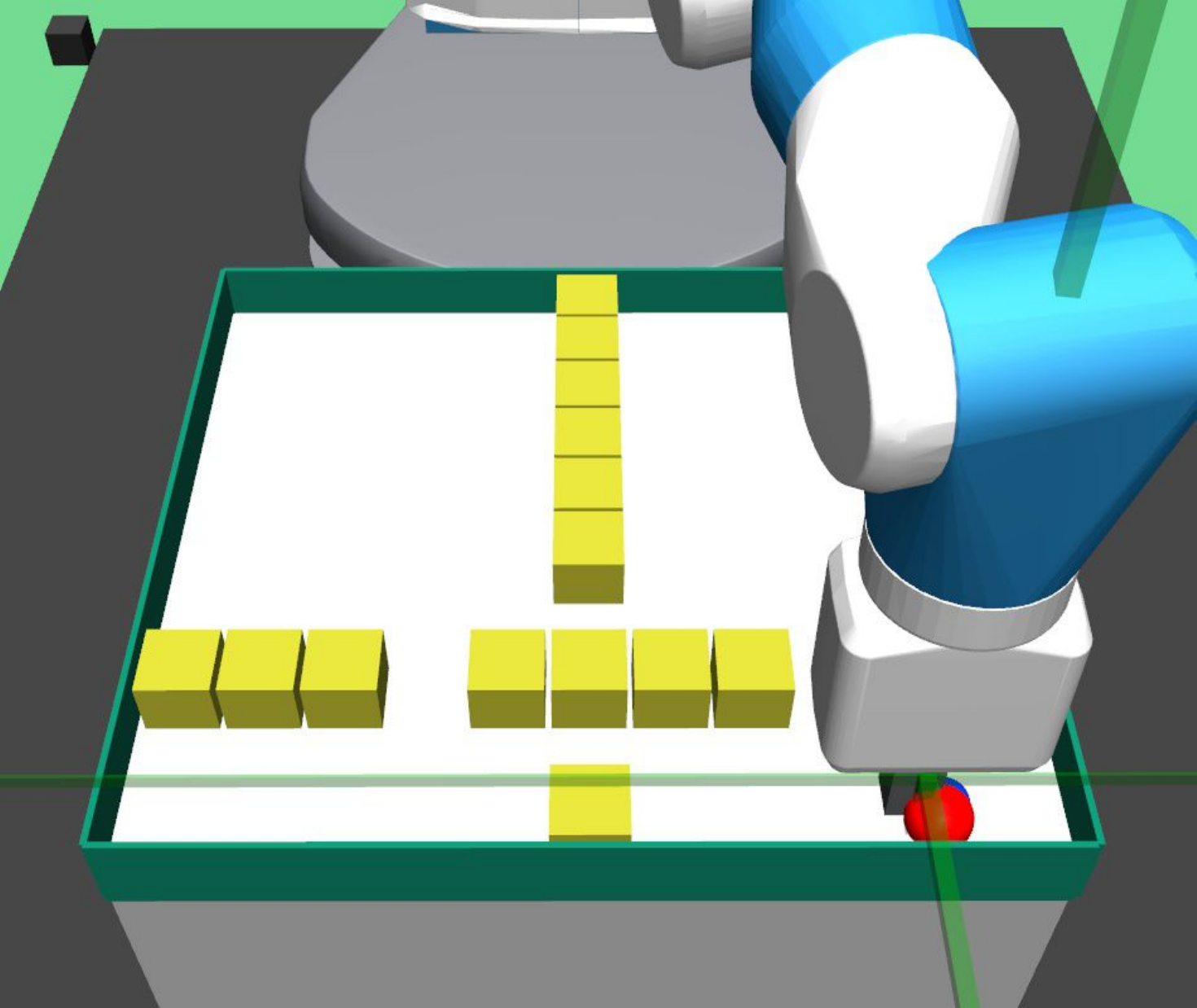}
\includegraphics[scale=0.11]{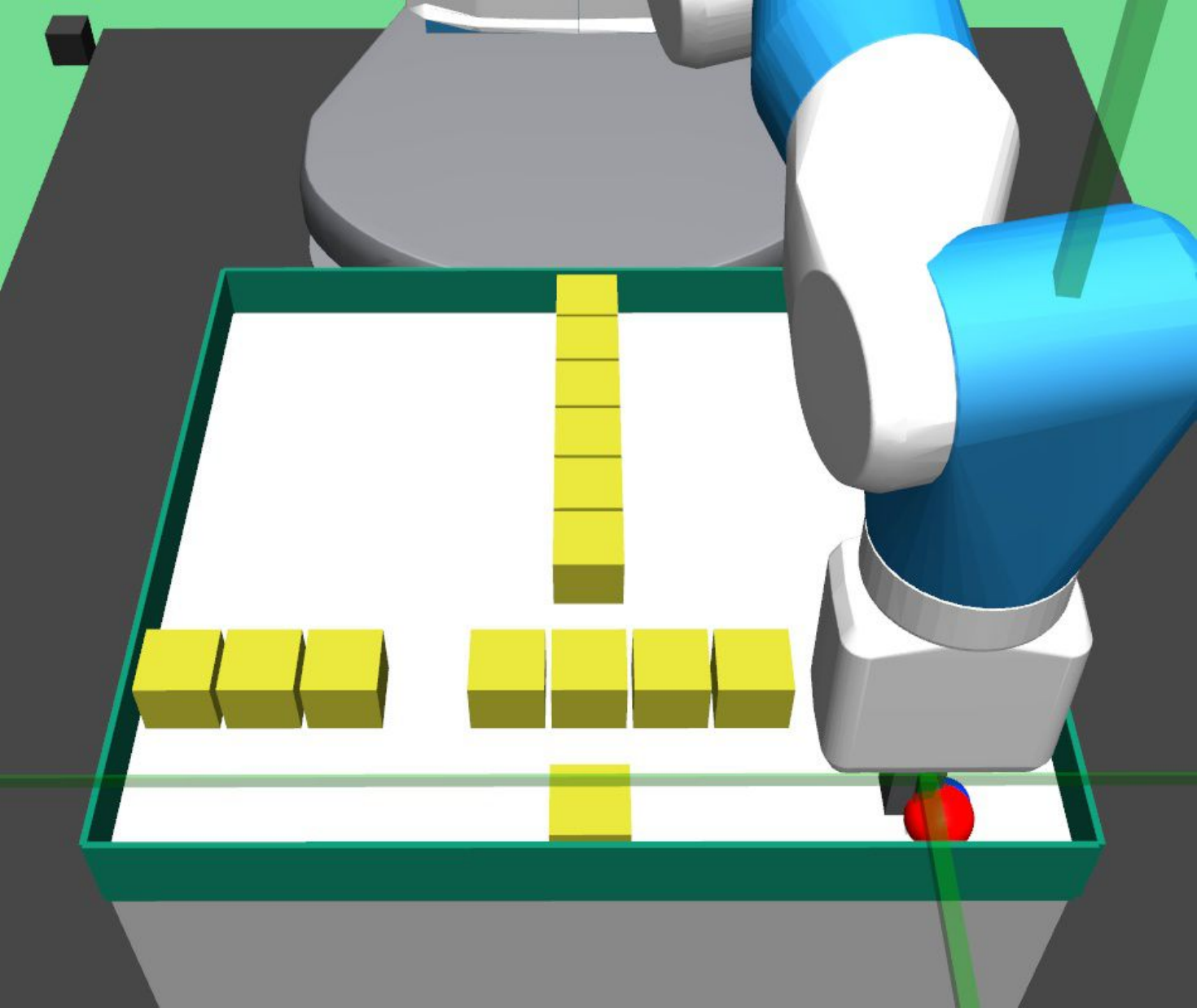}
\includegraphics[scale=0.11]{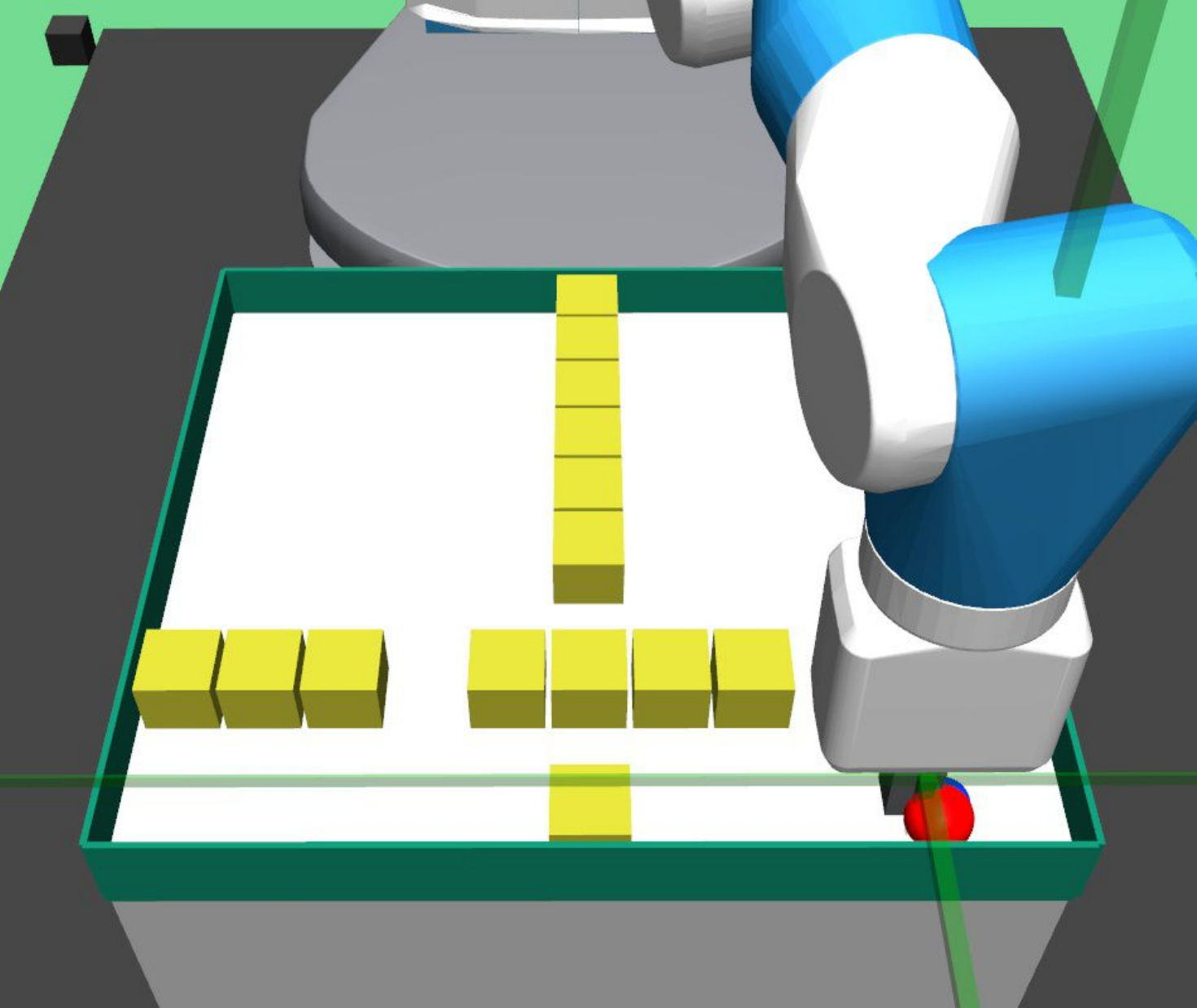}
\caption{\textbf{Maze navigation task visualization}: The visualization is a successful attempt at performing maze navigation task}
\label{fig:maze_viz_success_2_ablation}
\end{figure}

\begin{figure}[H]
\vspace{5pt}
\centering
\includegraphics[scale=0.11]{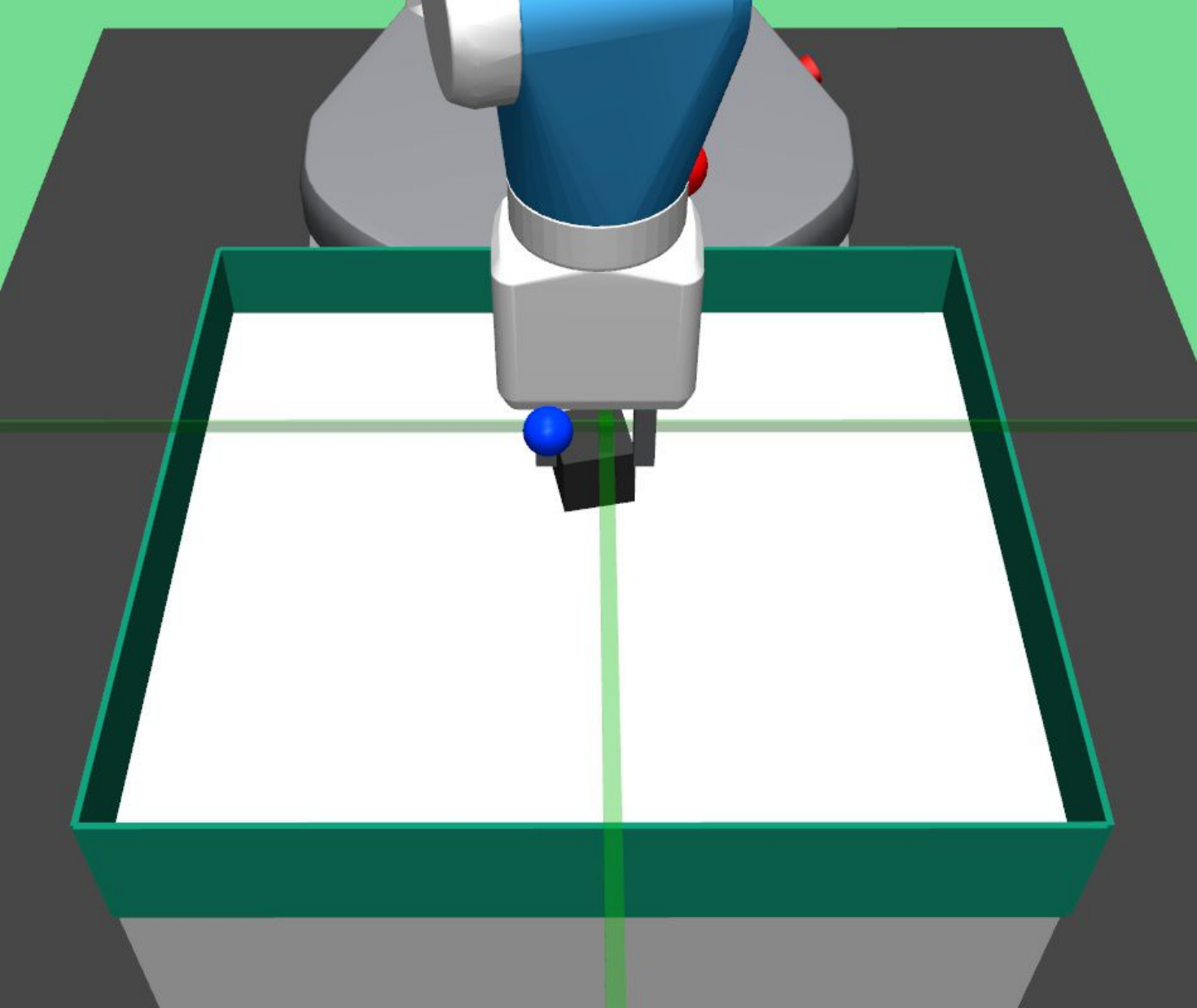}
\includegraphics[scale=0.11]{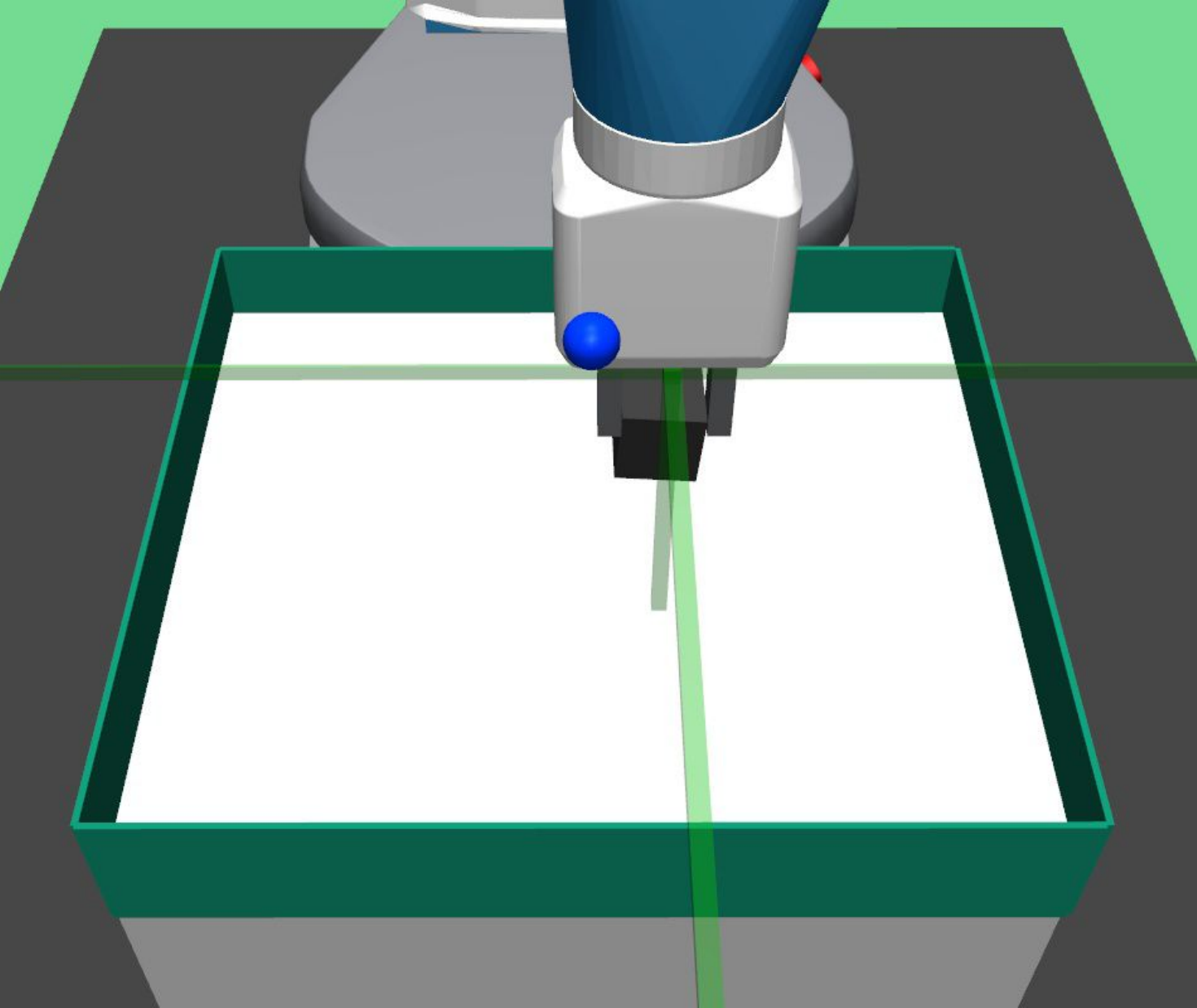}
\includegraphics[scale=0.11]{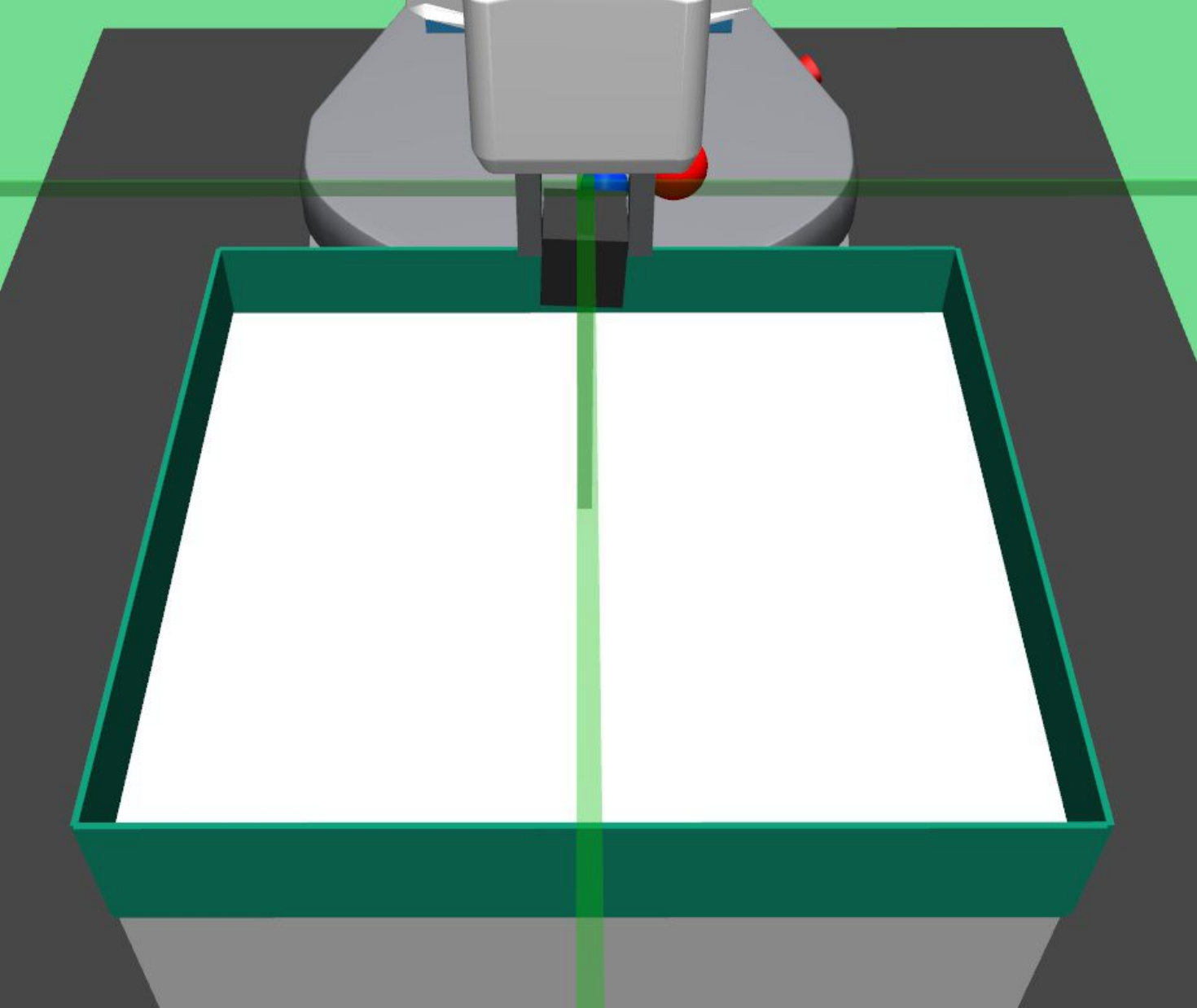}
\includegraphics[scale=0.11]{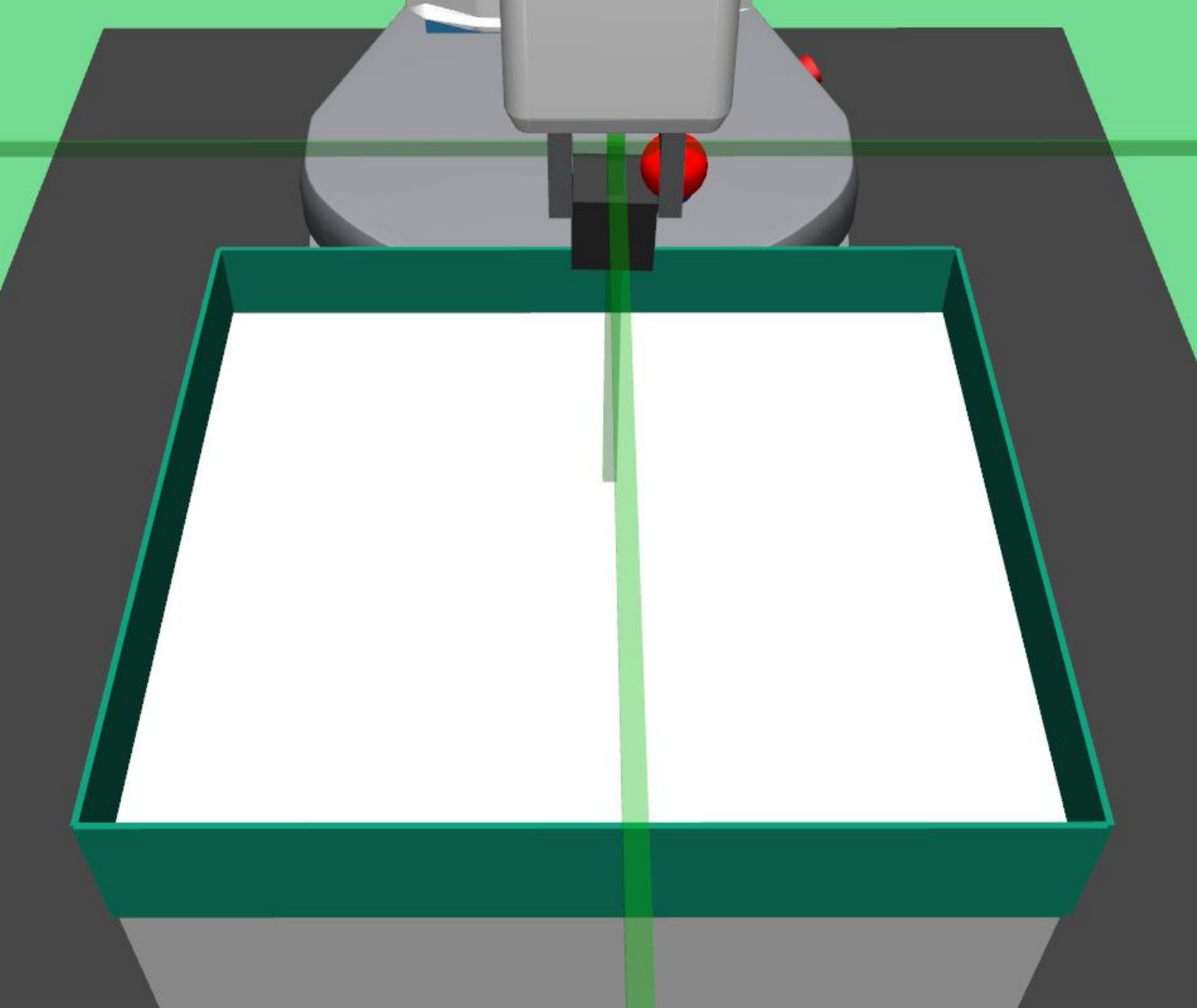}
\includegraphics[scale=0.11]{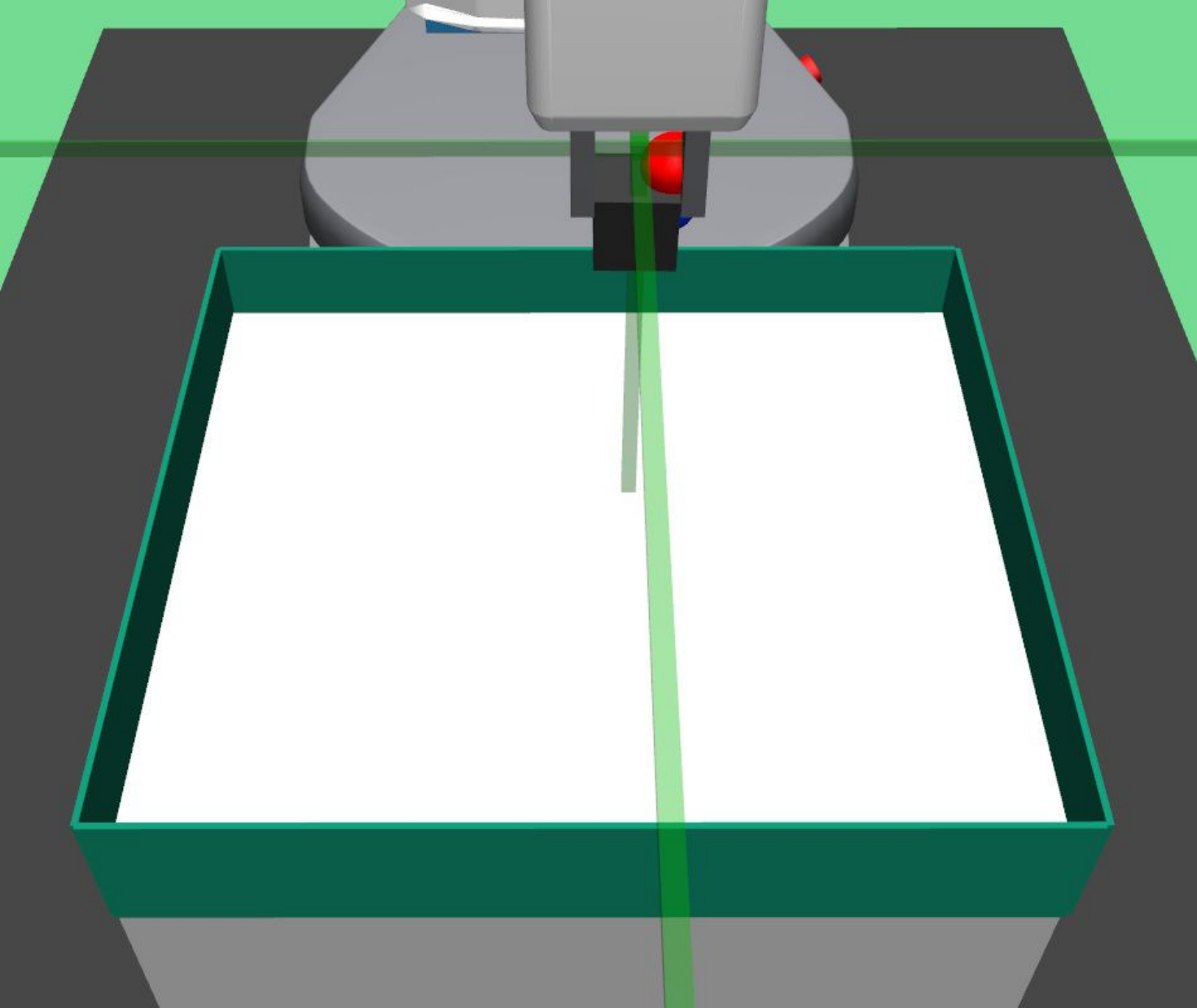}
\includegraphics[scale=0.11]{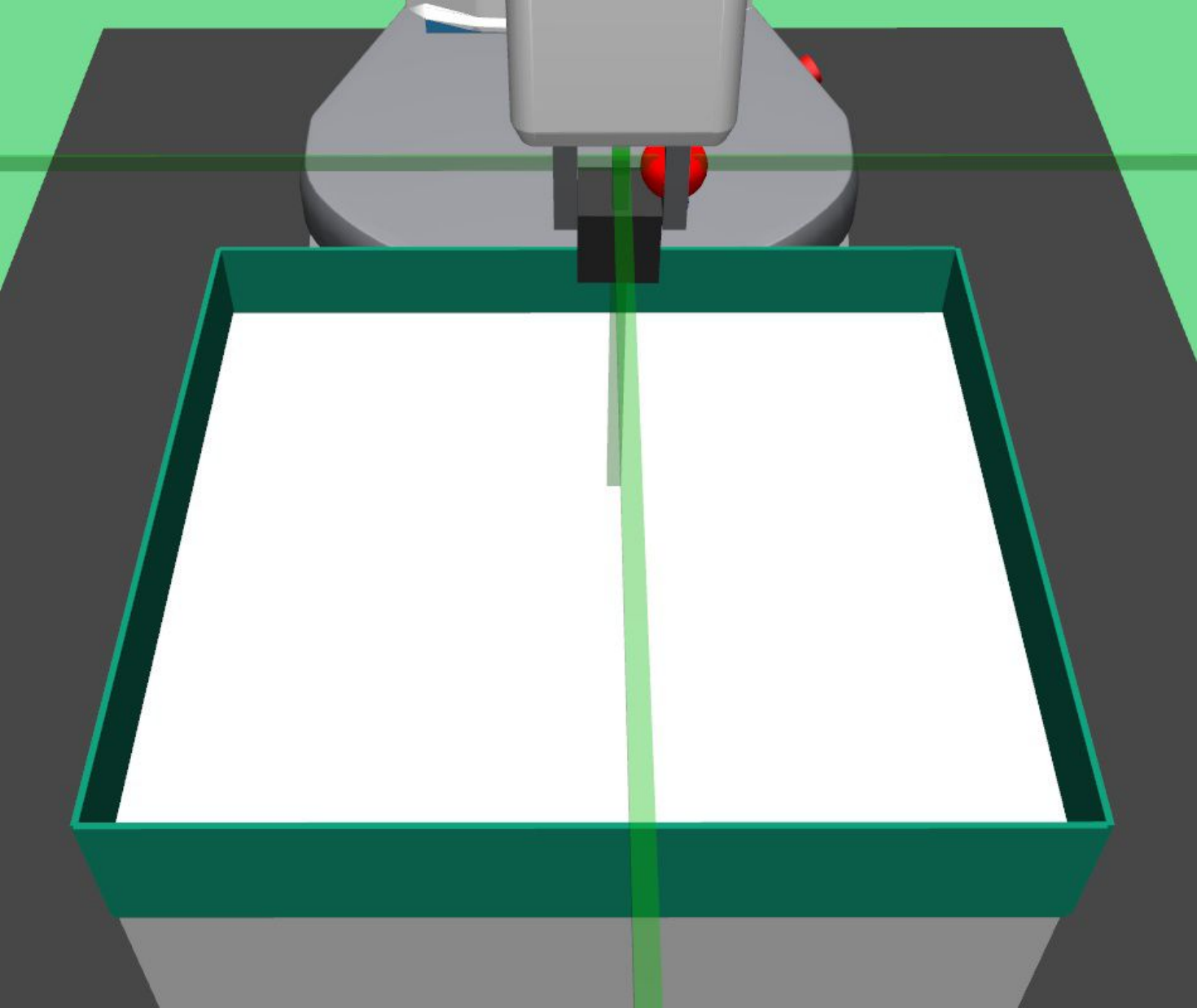}
\caption{\textbf{Pick and place task visualization}: This figure provides visualization of a successful attempt at performing pick and place task}
\label{fig:pick_viz_success_2_ablation}
\end{figure}

\begin{figure}[H]
\vspace{5pt}
\centering

\includegraphics[scale=0.08]{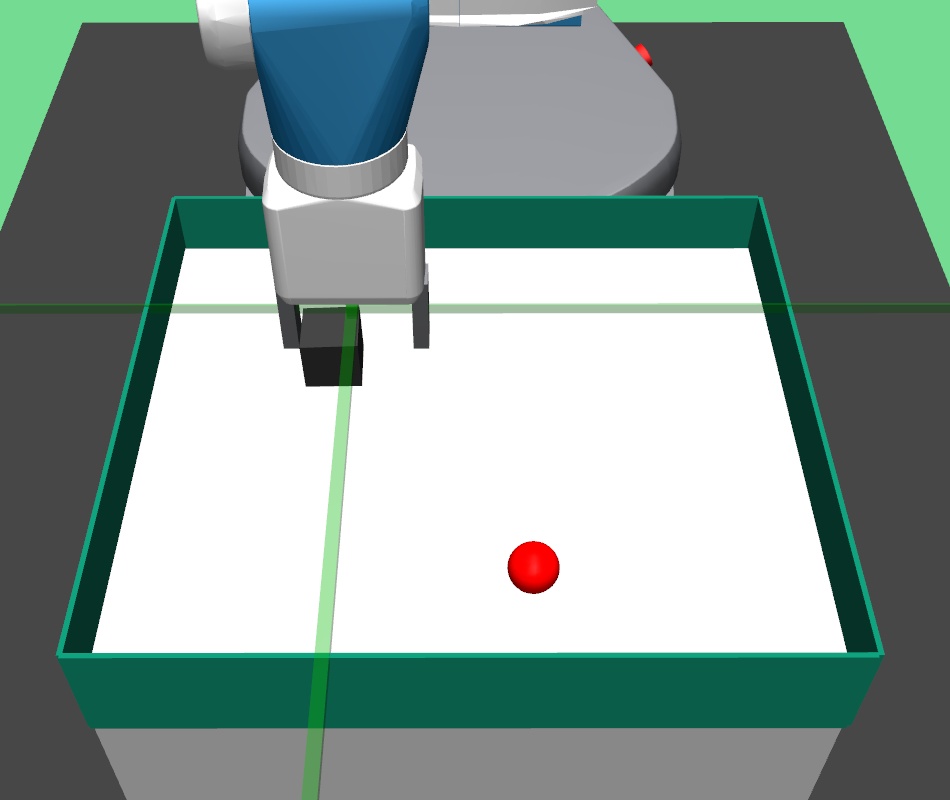}
\includegraphics[scale=0.08]{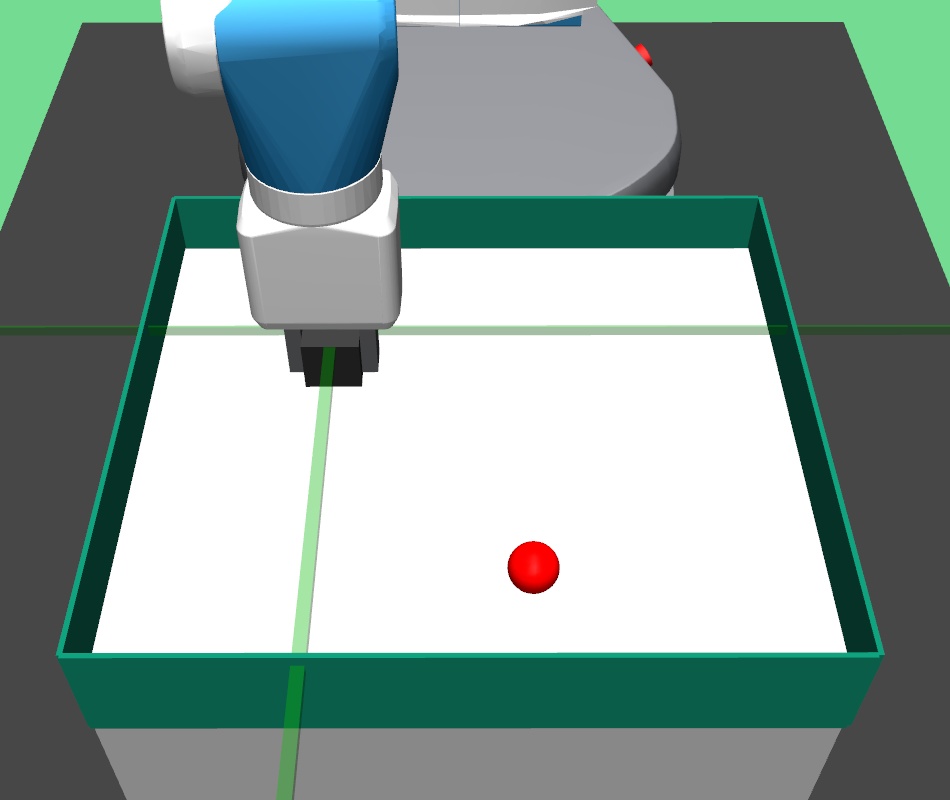}
\includegraphics[scale=0.08]{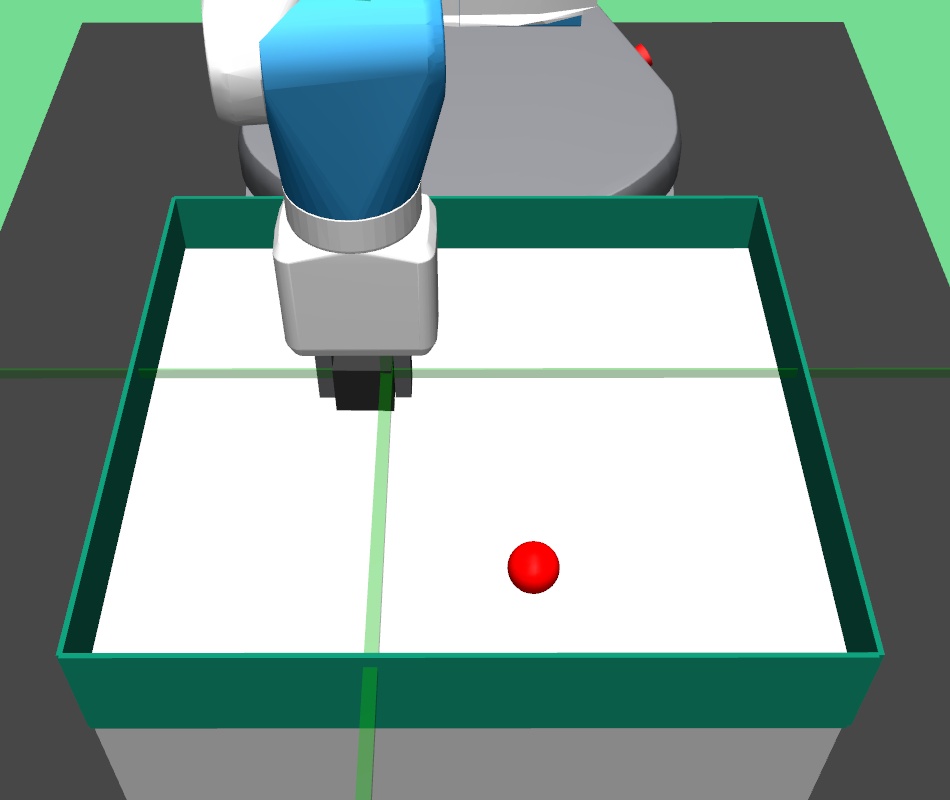}
\includegraphics[scale=0.08]{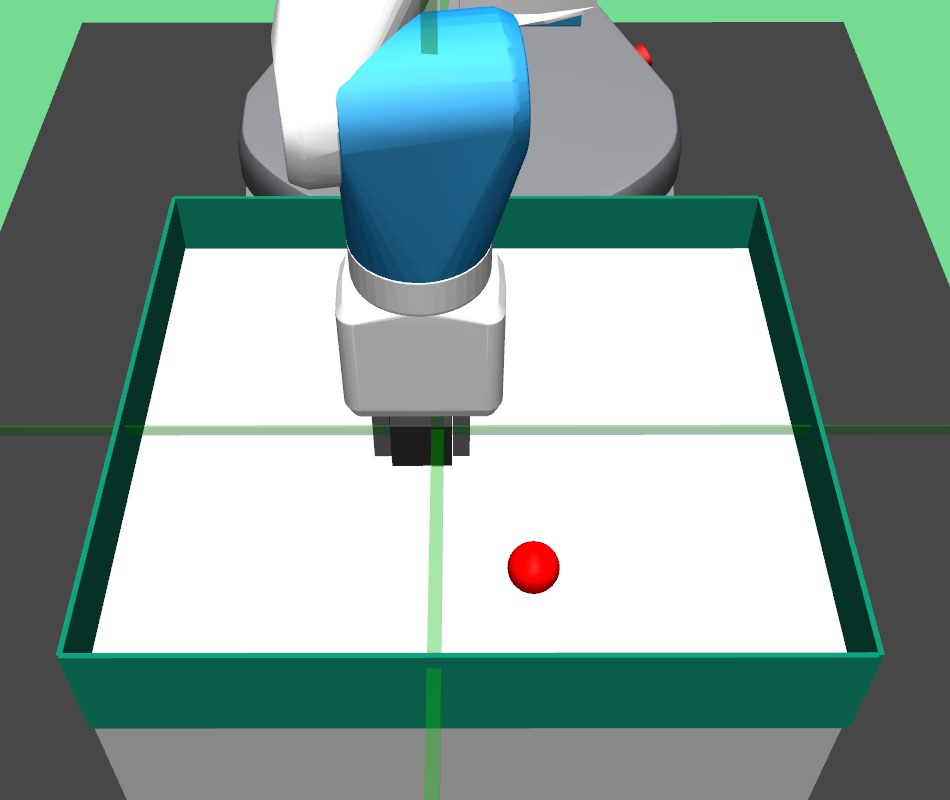}
\includegraphics[scale=0.08]{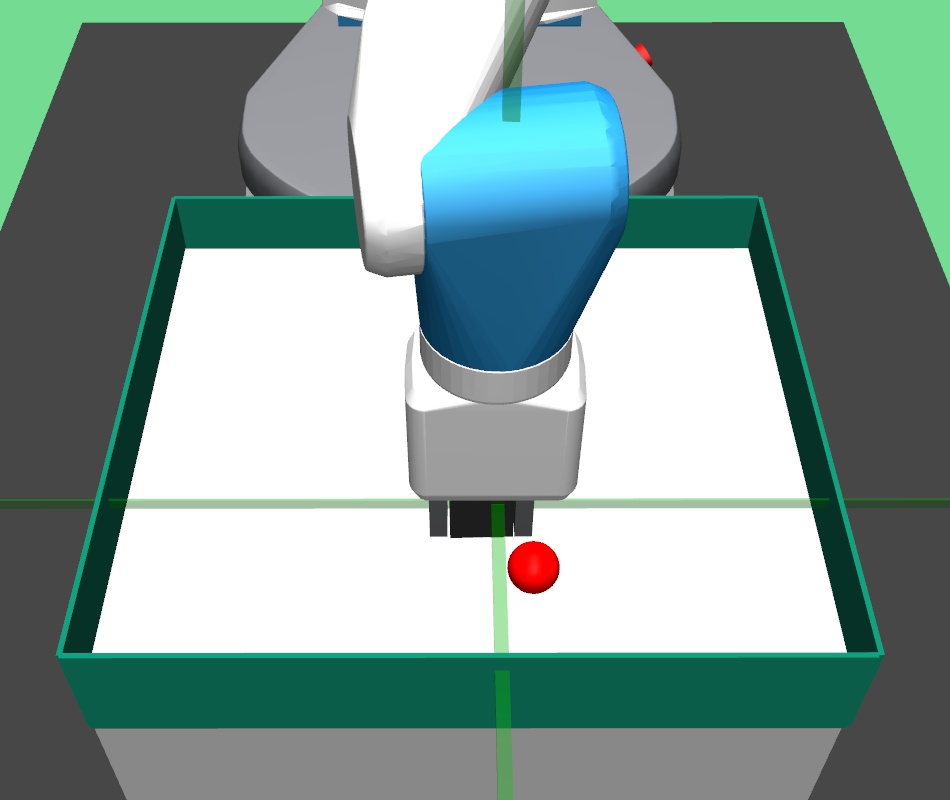}
\includegraphics[scale=0.08]{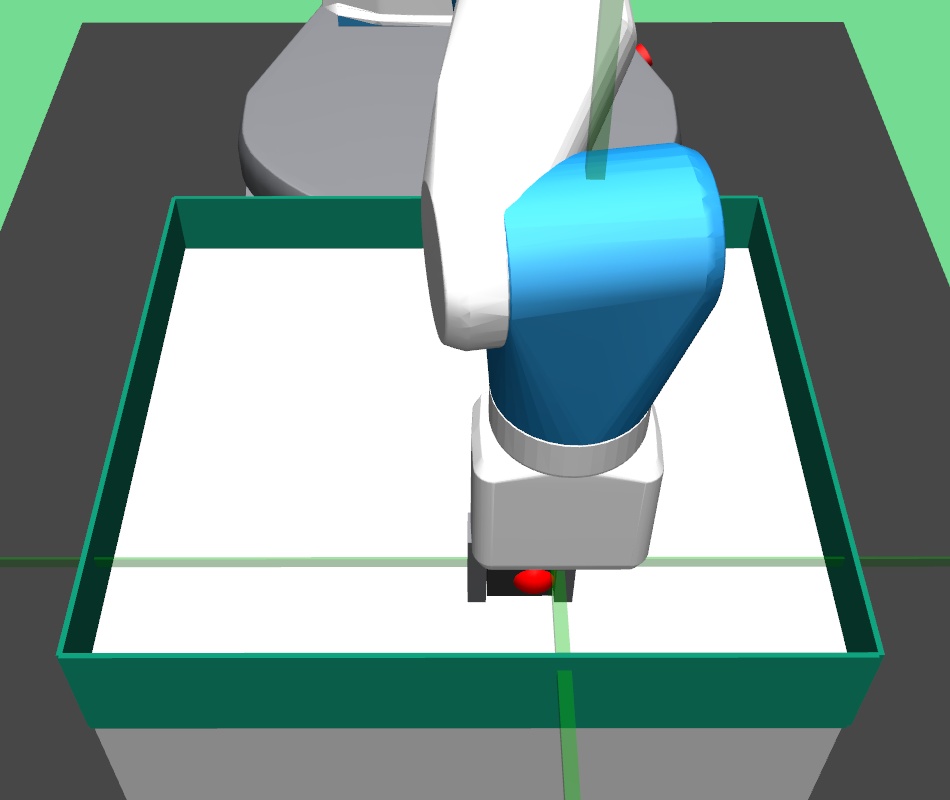}
\caption{\textbf{Push task visualization}: The visualization is a successful attempt at performing push task}
\label{fig:push_viz_success_2_ablation}
\end{figure}

\begin{figure}[H]
\vspace{5pt}
\centering

\includegraphics[scale=0.09]{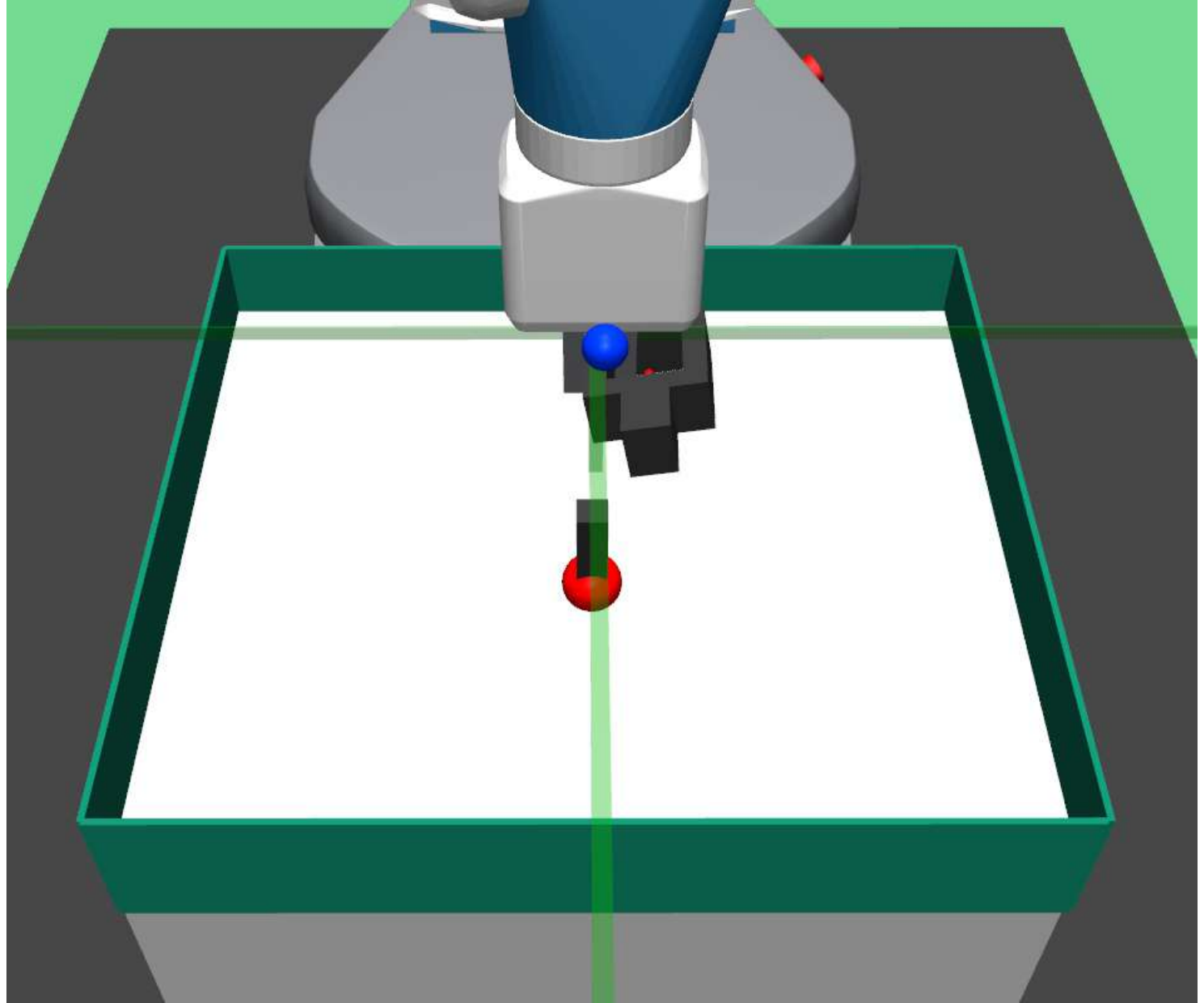}
\includegraphics[scale=0.09]{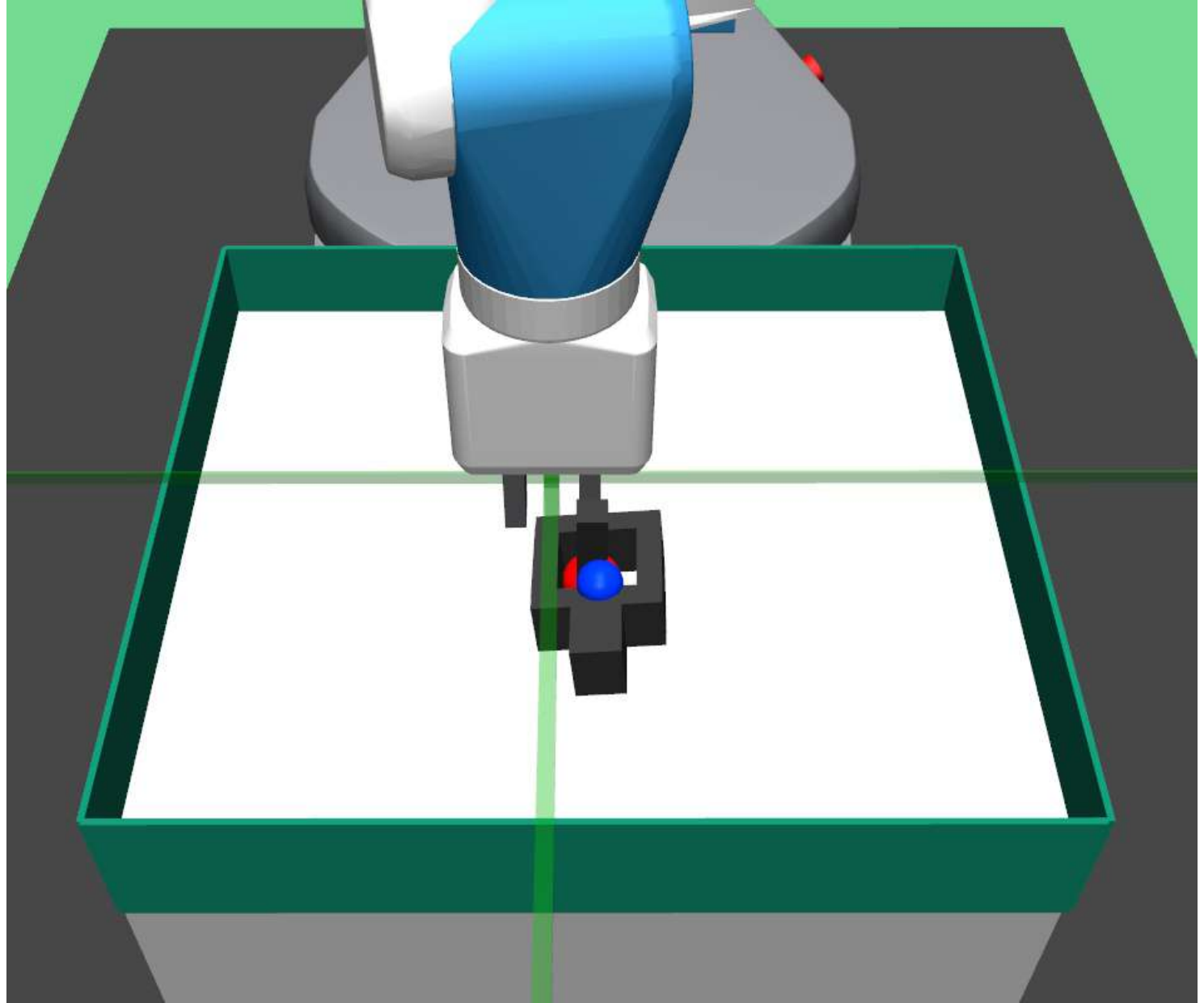}
\includegraphics[scale=0.09]{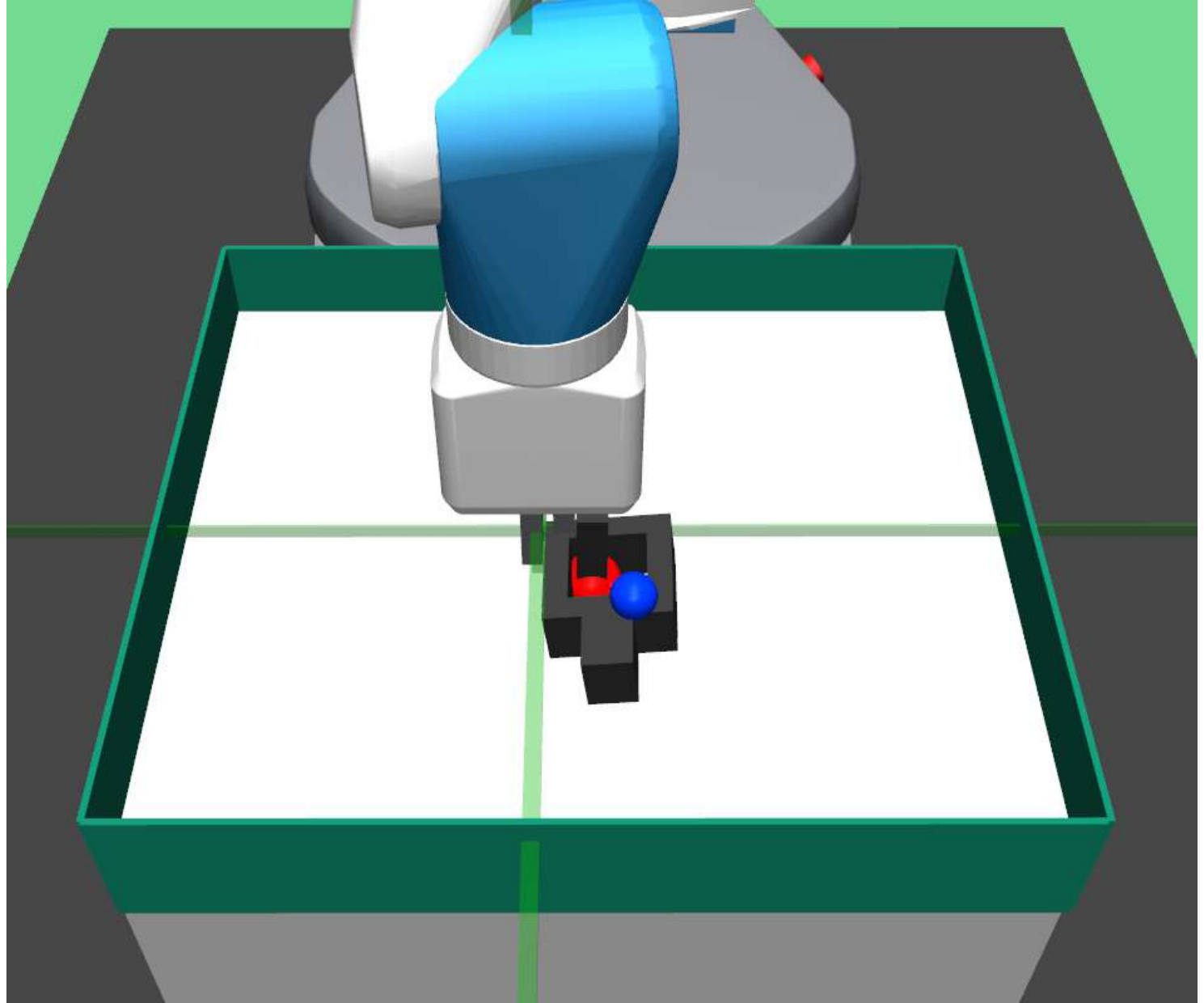}
\includegraphics[scale=0.09]{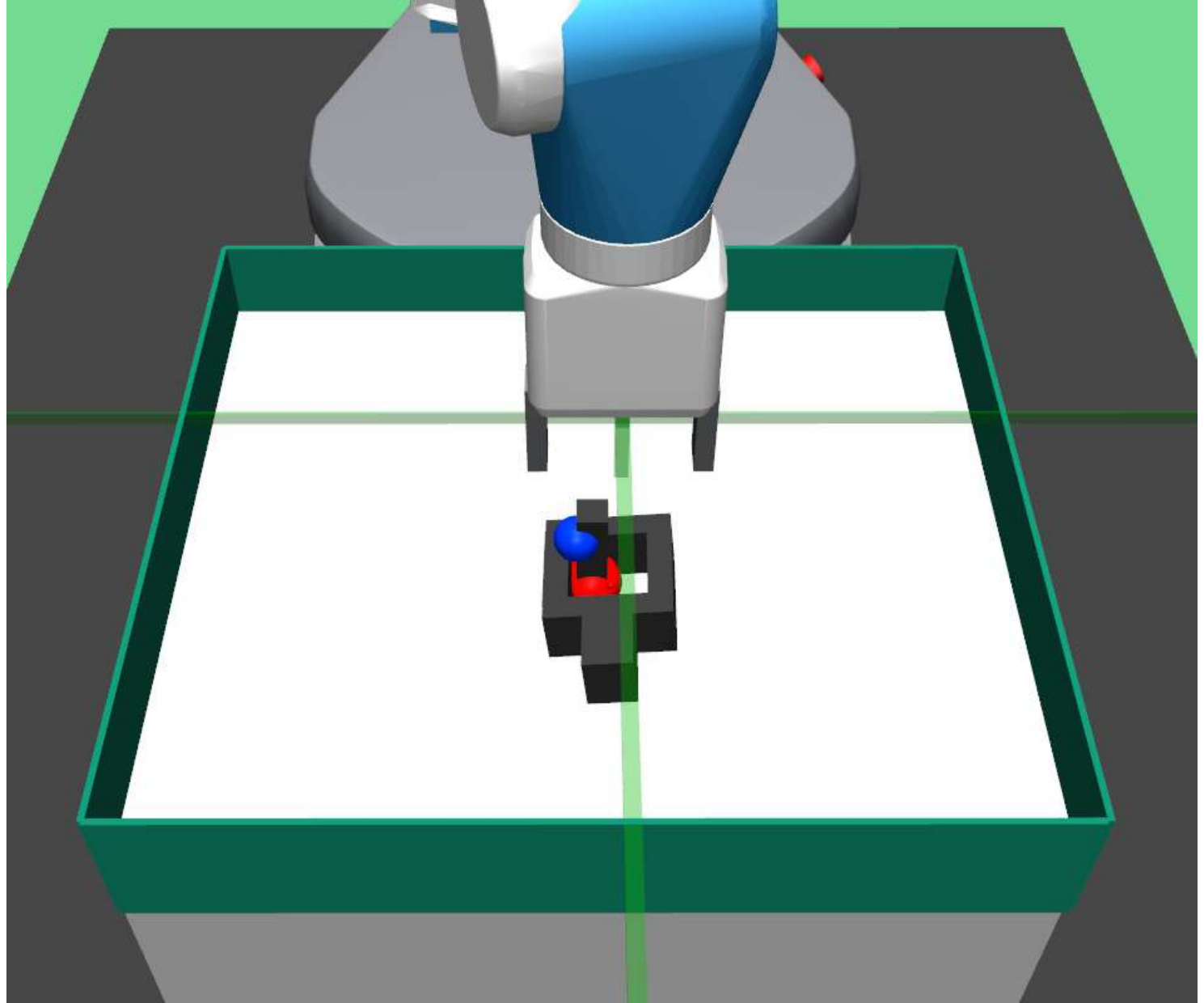}
\includegraphics[scale=0.09]{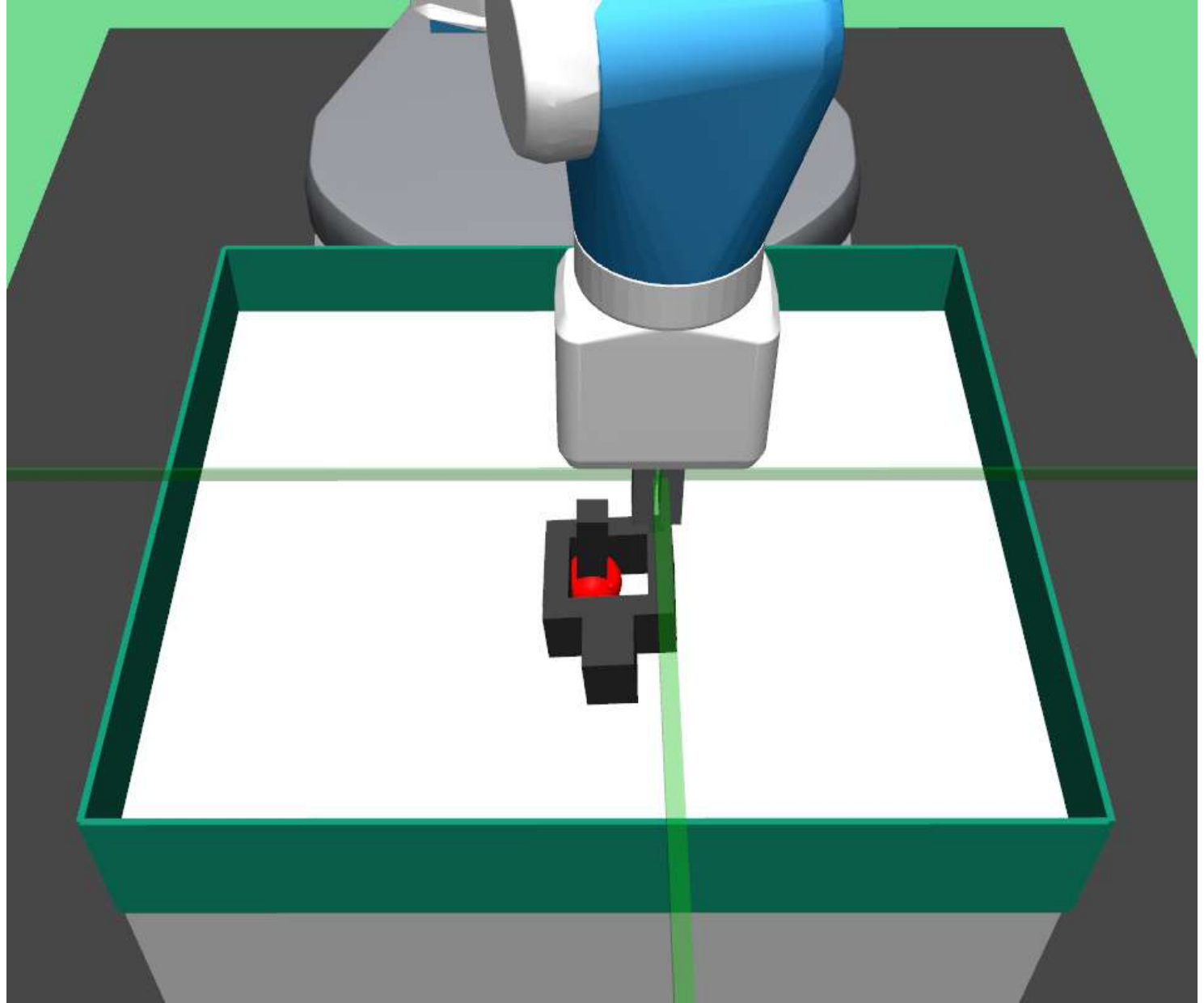}
\includegraphics[scale=0.09]{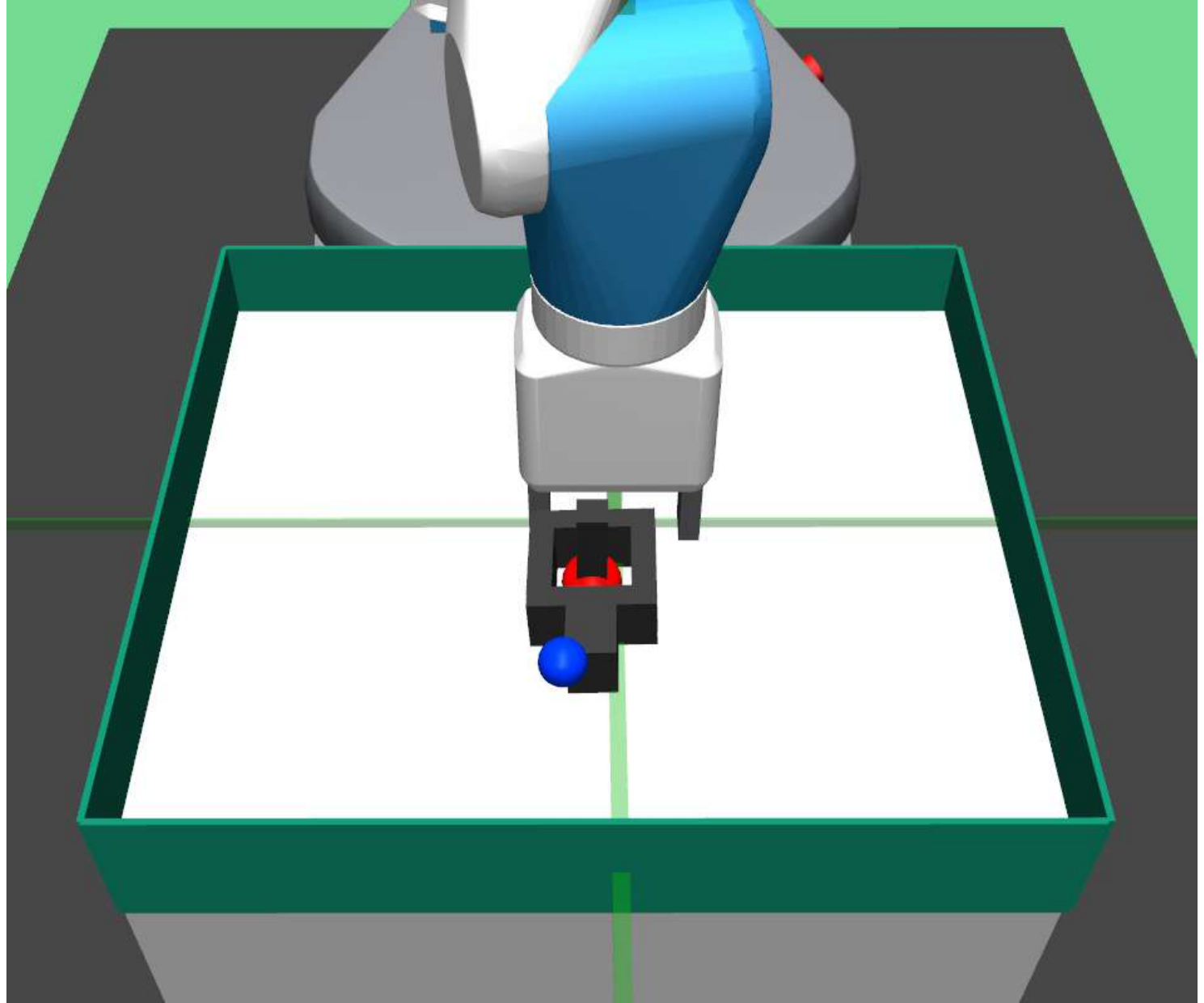}
\caption{\textbf{Hollow task visualization}: The visualization is a successful attempt at performing hollow task}
\label{fig:hollow_viz_success_2_ablation}
\end{figure}

\begin{figure}[H]
\vspace{5pt}
\centering

\includegraphics[scale=0.1]{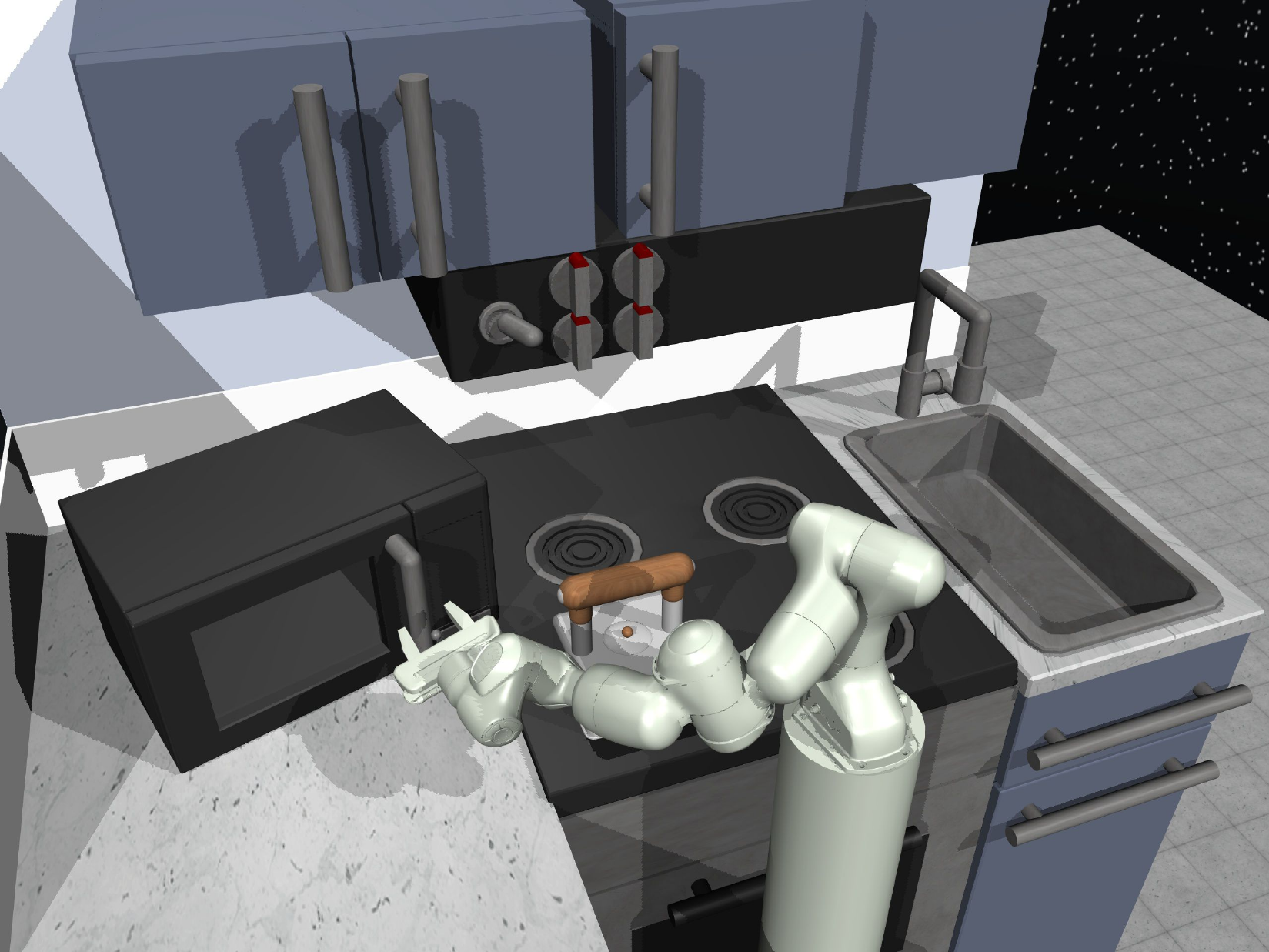}
\includegraphics[scale=0.1]{figures/kitchen_seed_6_viz_1.pdf}
\includegraphics[scale=0.1]{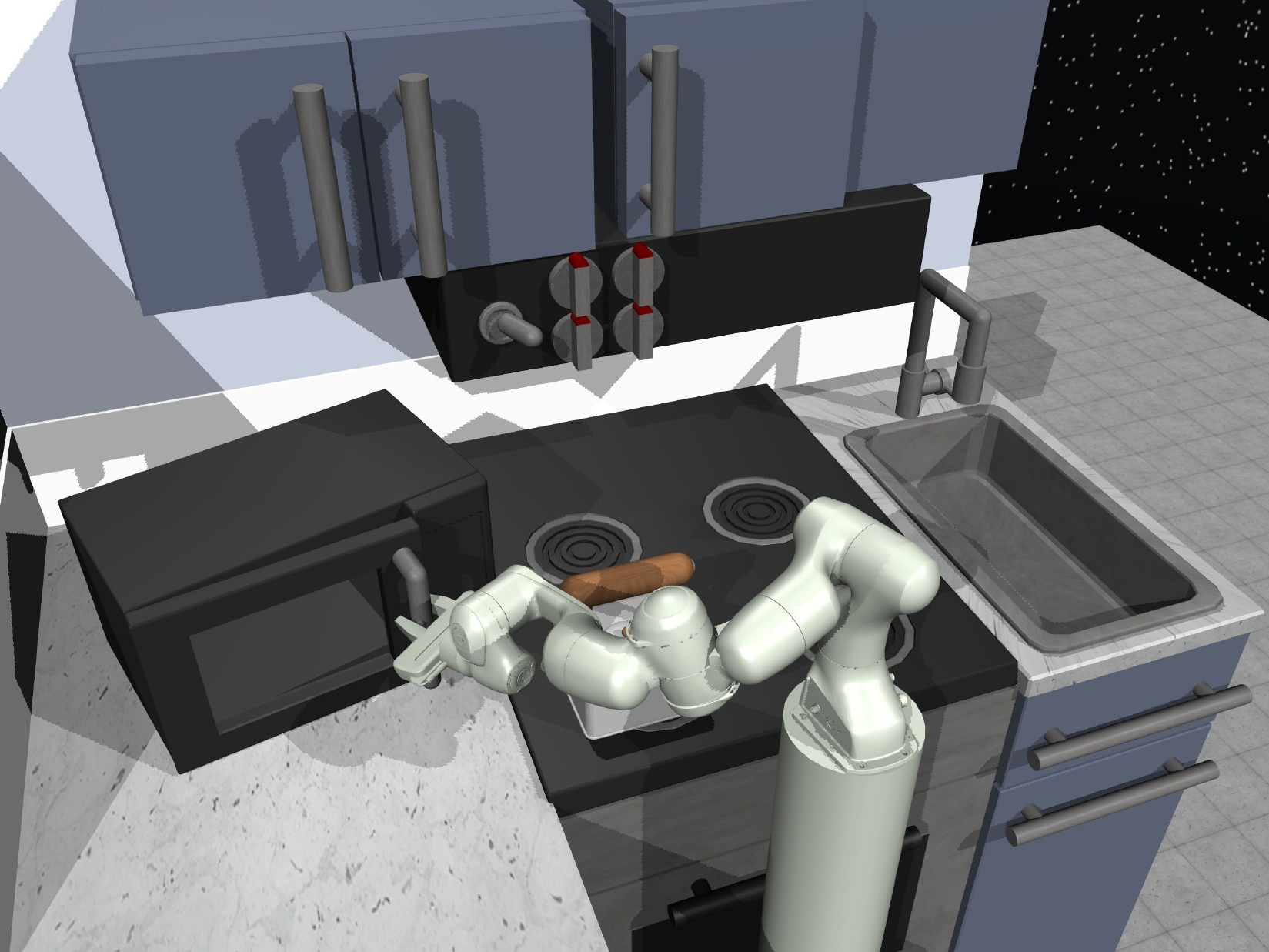}
\includegraphics[scale=0.1]{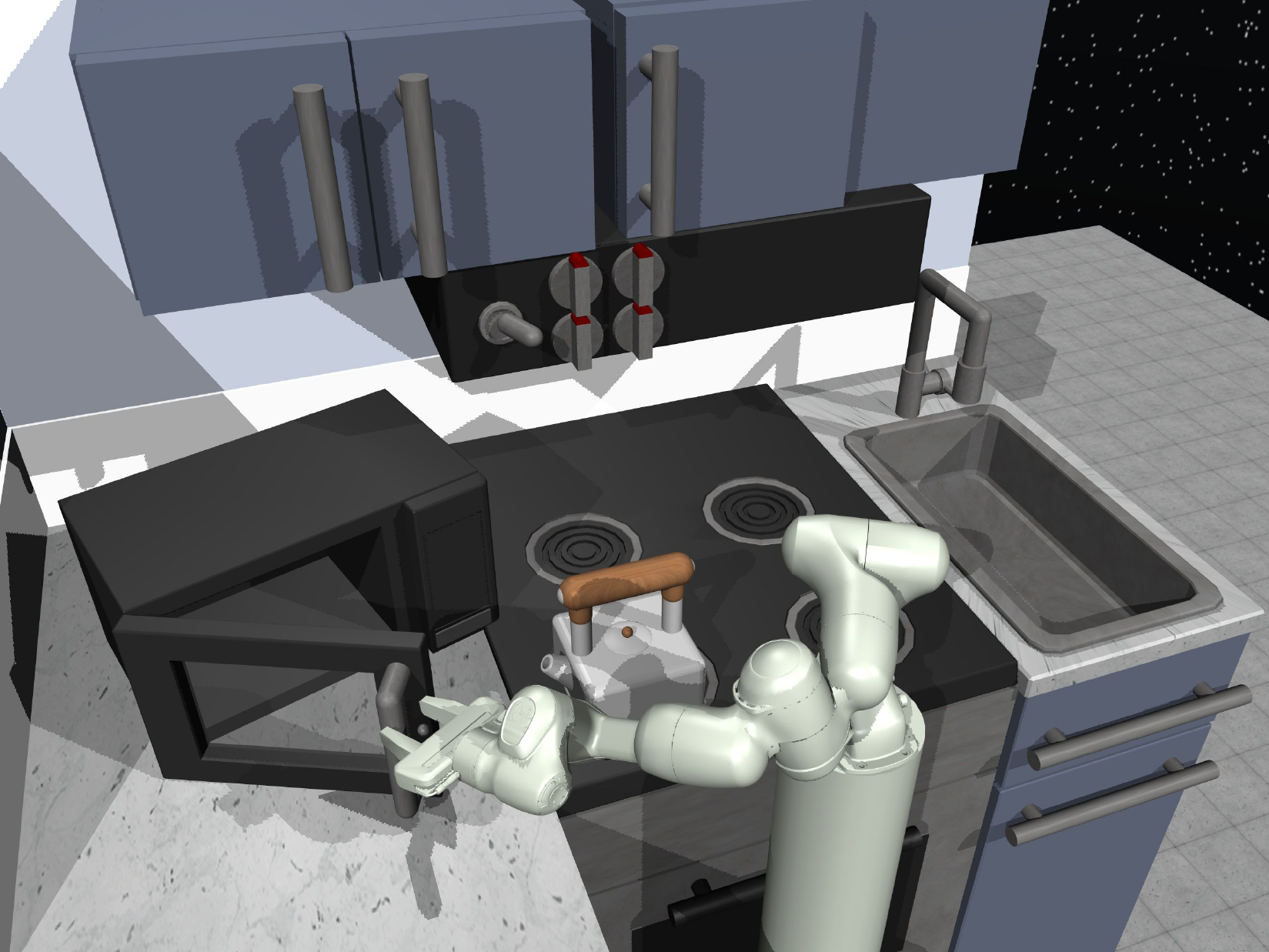}
\includegraphics[scale=0.1]{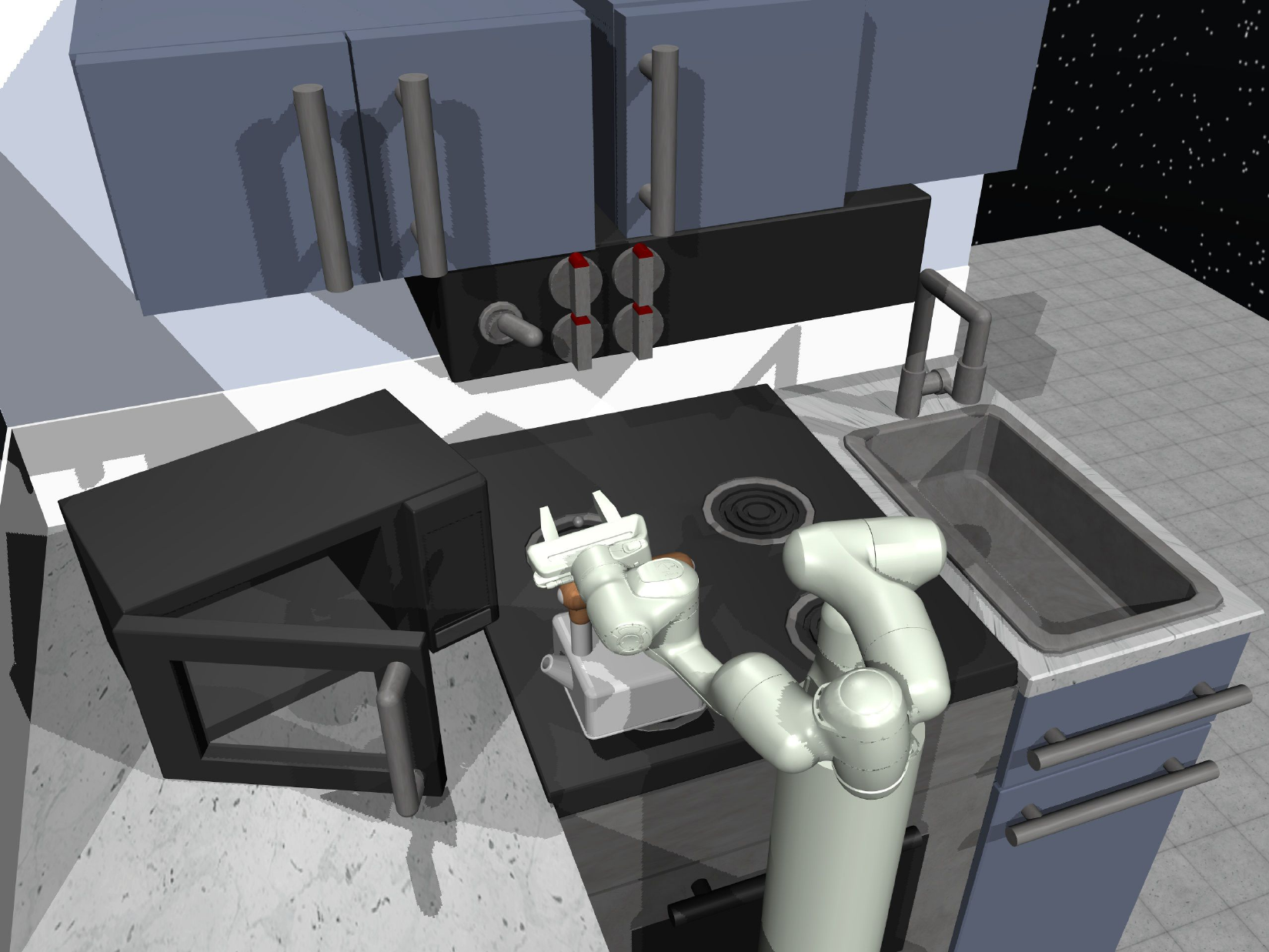}
\caption{\textbf{Kitchen task visualization}: The visualization is a successful attempt at performing kitchen task}
\label{fig:kitchen_viz_success_2_ablation}
\end{figure}


\end{document}